\pdfoutput=1

\documentclass[11pt]{article}

\usepackage[preprint]{acl}
\usepackage{multicol}
\usepackage{multirow}
\usepackage{times}
\usepackage{latexsym}
\usepackage{booktabs}
\usepackage{amssymb}
\usepackage{listings}
\usepackage{bbm}
\usepackage[T1]{fontenc}

\usepackage[utf8]{inputenc}
\usepackage{amsmath}
\usepackage{microtype}

\usepackage{inconsolata}

\usepackage{graphicx}
\usepackage{xcolor}
\usepackage{tcolorbox}
\tcbuselibrary{skins,breakable}
\usepackage{tabularx}
\usepackage{array}
\usepackage{enumitem}
\usepackage{makecell}
\usepackage{booktabs}
\usepackage{wrapfig}
\usepackage{pifont}
\usepackage{algorithm}
\usepackage{algpseudocode}

\newcommand{\cmark}{\textcolor[rgb]{0.0, 0.6, 0.0}{\ding{51}}} 
\newcommand{\xmark}{\textcolor[rgb]{0.7, 0.0, 0.0}{\ding{55}}} 
\newcommand{\gmark}{\textcolor[rgb]{1,0.647,0}{\ding{51}}}

\definecolor{rgbBlueBg}{HTML}{EAF4FF}
\definecolor{emreblue}{HTML}{4d7ea8}

\NewDocumentCommand{\jy}
{ mO{} }{\textcolor{blue}{\textsuperscript{\textit{Jiayu}}\textsf{\textbf{\small[#1]}}}}
\NewDocumentCommand{\qh}
{ mO{} }{\textcolor{purple}{\textsuperscript{\textit{Qihan}}\textsf{\textbf{\small[#1]}}}}
\NewDocumentCommand{\cheng}
{ mO{} }{\textcolor{orange}{\textsuperscript{\textit{cheng}}\textsf{\textbf{\small[#1]}}}}
\NewDocumentCommand{\rui}
{ mO{} }{\textcolor{red}{\textsuperscript{\textit{Rui}}\textsf{\textbf{\small[#1]}}}}
\NewDocumentCommand{\xc}
{ mO{} }{\textcolor{cyan}{\textsuperscript{\textit{Xiaocheng}}\textsf{\textbf{\small[#1]}}}}
\NewDocumentCommand{\emre}
{ mO{} }{\textcolor{emreblue}{\textsuperscript{\textit{Emre}}\textsf{\textbf{\small[#1]}}}}
\NewDocumentCommand{\TODO}
{ mO{} }{\textcolor{red}{\textsuperscript{\textit{TODO}}\textsf{\textbf{\small[#1]}}}}
\NewDocumentCommand{\heng}
{ mO{} }{\textcolor{red}{\textsuperscript{\textit{Heng}}\textsf{\textbf{\small[#1]}}}}

\title{\textbf{PlanBench-XL}: Evaluating Long-Horizon Planning of LLM Tool-Use Agents in Large-Scale Tool Ecosystems}

\tcbset{
  aibox/.style={
    width=\linewidth,
    top=7pt,
    bottom=2pt,
    colback=rgbBlueBg,
    colframe=black,
    colbacktitle=black,
    enhanced,
    center,
    attach boxed title to top left={
      yshift=-0.1in,
      xshift=0.15in
    },
    boxed title style={
      boxrule=0pt,
      colframe=white
    },
  }
}

\newcounter{takeaway}
\renewcommand{\thetakeaway}{\arabic{takeaway}}

\newenvironment{takeaway}[1][]%
{%
  \refstepcounter{takeaway}%
  \begin{tcolorbox}[
    aibox,
    title={Takeaway~\thetakeaway},
    #1
  ]%
}%
{%
  \end{tcolorbox}%
}

\newcommand{\BENCH}[1]{\textbf{PlanBench-XL}}
\newcommand{\bench}[1]{PlanBench-XL}

\newcommand{\logoh}{1.35em}

\newcommand{\linkblock}{%
\small
\href{https://github.com/JiayuJeff/PlanBench-XL}{%
\raisebox{-0.15\height}{\includegraphics[height=\logoh]{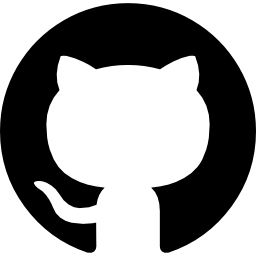}}%
\hspace{0.35em}{\textbf{Code}}%
}%
\quad
\href{https://huggingface.co/datasets/JiayuJeff/PlanBench-XL}{%
\raisebox{-0.2\height}{\includegraphics[height=\logoh]{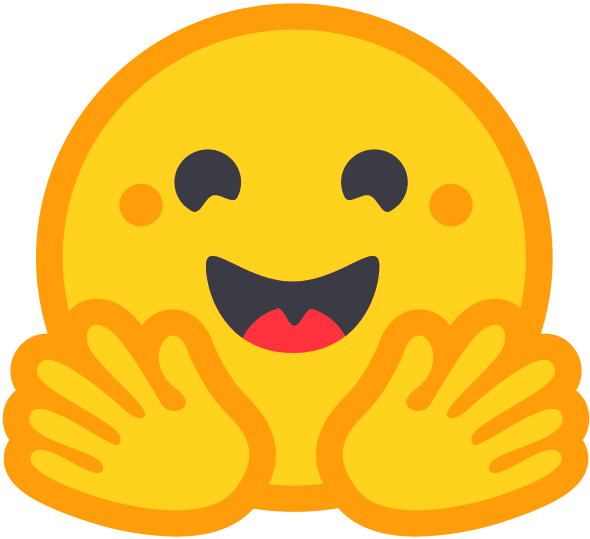}}%
\hspace{0.35em}{\textbf{Dataset}}%
}%
\quad
\href{https://planbench-xl.github.io/}{%
\raisebox{-0.2\height}{\includegraphics[height=\logoh]{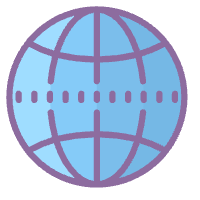}}%
\hspace{0.35em}{\textbf{Project Page}}%
}%
}

\author{
\textbf{Jiayu Liu\textsuperscript{*}},
\textbf{Qihan Lin\textsuperscript{*}},
\textbf{Cheng Qian\textsuperscript{}},
\textbf{Rui Wang\textsuperscript{}}, 
\textbf{Emre Can Acikgoz\textsuperscript{}}, \\
\textbf{Xiaocheng Yang\textsuperscript{}}, 
\textbf{Jiateng Liu\textsuperscript{}},
\textbf{Zhenhailong Wang}\textsuperscript{},  \\
\textbf{Xiusi Chen\textsuperscript{}},
\textbf{Heng Ji\textsuperscript{}},
\textbf{Dilek Hakkani-T\"ur\textsuperscript{}} \\
University of Illinois Urbana-Champaign \\
 \texttt{\{jiayul12,hengji,dilek\}@illinois.edu}
\\[1.5ex]
\linkblock
}

\begin{document}

\maketitle

\begin{abstract}


LLM agents increasingly operate in large tool ecosystems, where real-world tasks require discovering relevant tools, inferring implicit sub-goals, and adapting to dynamic environments over long horizons.
However, existing benchmarks rarely evaluate planning under retrieval-limited tool visibility.
To address this gap, we introduce \BENCH{}, an interactive benchmark of 327 retail tasks over 1,665 tools that tests whether agents can iteratively retrieve usable tools, invoke them to uncover intermediate evidence for subsequent calls toward the final goal.
\bench{} further features an optional blocking mechanism that simulates real-world unpredictability through missing, failing, or distracting tool functions, forcing agents to detect disrupted paths and adapt at runtime.
Experiments on ten leading LLMs show that massive-tool planning remains challenging: while \textit{GPT-5.4} achieves 51.90\% accuracy in block-free settings, it collapses to 11.36\% under the most severe blocking condition.
Further analysis shows that agents are especially vulnerable when failures lack explicit error signals or when recovery requires longer alternative tool-use paths.
These results establish \bench{} as a testbed for diagnosing agentic planning failures and highlight the need for robust adaptive planning in long-horizon tasks with large, imperfect tool environments.\footnote{*Equal Contribution.}

\end{abstract}
\section{Introduction}

Large language model (LLM) agents equipped with external tools have shown strong capabilities in solving complex real-world tasks~\citep{ToolRL, Code2Math}. 
In practice, however, tool environments are often large-scale, including enterprise MCP servers~\citep{MCP, MCP-Bench}, software ecosystems~\citep{APIBench} and web/API platforms~\citep{ToolBench}. 
Due to context-length limitations, agents often rely on tool retrieval~\citep{ToolGen, ToolRet} to access only a relevant subset of tool descriptions at each step, making tool use a retrieval-mediated process.
This partial tool visibility becomes especially challenging in long-horizon tasks~\citep{SELFGOAL, EscapeBench}, where agents must explore intermediate sub-goals, retrieve useful tools, and adapt their plans as the task unfolds.

\begin{table*}[t]
\small
\centering
\resizebox{0.97\linewidth}{!}{
    \begin{tabular}{lccccccc}
    \toprule
    \textbf{Benchmark} 
    & \textbf{\makecell{Tool-\\Use}} 
    & \textbf{\makecell{Tool\\Retrieval}}
    & \textbf{\makecell{Implicit\\Sub-goals}}
    & \textbf{\makecell{Bi-directional\\Exploration}}
    & \textbf{\makecell{Unreliable\\Tools}} 
    & \textbf{\makecell{Long-\\Horizon}}
    & \textbf{\makecell{Scalable\\Generation}} \\
    \midrule
    \textit{ToolBench}~\citep{ToolBench} & \cmark & \cmark & \xmark & \xmark & \xmark & \xmark & \gmark \\
    \textit{RestBench}~\citep{RestGPT} & \cmark & \xmark & \xmark & \xmark & \xmark & \xmark & \xmark \\
    \textit{APIBench}~\citep{APIBench} & \cmark & \cmark & \xmark & \xmark & \gmark & \xmark & \gmark \\
    \textit{ToolRet}~\citep{ToolRet} & \xmark & \cmark & \xmark & \xmark & \xmark & \xmark & \gmark \\
    \textit{API-Bank}~\citep{API-Bank} & \cmark & \cmark & \xmark & \xmark & \xmark & \xmark & \cmark \\
    \textit{EscapeBench}~\citep{EscapeBench} & \cmark & \xmark & \cmark & \gmark & \xmark & \cmark & \xmark \\
    \textit{MCP-Universe}~\citep{MCP-Universe} & \cmark & \xmark & \gmark & \xmark & \xmark & \xmark & \xmark \\
    \textit{MCPBench}~\citep{MCP-Bench} & \cmark & \cmark & \xmark & \gmark & \xmark & \gmark & \cmark \\
    \textit{ACEBench}~\citep{ACEBench} & \cmark & \xmark & \gmark & \xmark & \xmark & \xmark & \gmark \\
    \textit{BFCL v4}~\citep{BFCL-v4} & \cmark & \xmark & \xmark & \xmark & \cmark & \gmark & \xmark \\
    \textit{Tool Decathlon}~\citep{Tool-Decathlon} & \cmark & \xmark & \gmark & \xmark & \xmark & \cmark & \xmark \\
    \textit{LiveMCPBench}~\citep{LiveMCPBench} & \cmark & \cmark & \xmark & \gmark & \xmark & \gmark & \xmark \\
    \textit{ToolGym}~\citep{ToolGym} & \cmark & \cmark & \xmark & \xmark & \cmark & \cmark & \cmark \\
    \textit{AgentNoiseBench}~\citep{AgentNoiseBench} & \cmark & \xmark & \xmark & \xmark & \cmark & \xmark & \cmark \\
    \textit{OpaqueToolsBench}~\citep{OpaqueToolsBench} & \cmark & \xmark & \xmark & \xmark & \cmark & \xmark & \cmark \\
    \textit{WildAGTEval}~\citep{kim2026perfectapiscomprehensiveevaluation} & \cmark & \xmark & \xmark & \xmark & \cmark & \xmark & \cmark \\
    \midrule
    \textit{\BENCH{} (\textbf{\textit{Ours}})} & \cmark & \cmark & \cmark & \cmark & \cmark & \cmark & \cmark \\
    \bottomrule
    \end{tabular}
}
\caption{For each benchmark, the table reports whether each trait is fully (\cmark), partially (\gmark), or not (\xmark) addressed. Detailed explanations of each traits are provided in Appendix~\ref{app:comparison-traits}. }
\label{tab:comparison-table}
\vspace{-0.1in}
\end{table*}

Furthermore, retrieval over large tool sets is inherently unreliable~\citep{Tool-Retrieval-is-not-reliable-1}. 
Relevant tools may be missed~\citep{COLT}, while retrieved tools may be damaged~\citep{OpaqueToolsBench}, stale~\citep{ToolQA-D}, misleading~\citep{TRUSTDESC}, or unreliable~\citep{ToolSandbox}. 
As summarized in Table~\ref{tab:comparison-table}, existing benchmarks do not fully capture this setting, as many assume a fixed, visible toolset~\citep{AppWorld}, explicit intermediate goals~\citep{ToolHop}, clean tool descriptions~\citep{OpaqueToolsBench}, or one-shot retrieval~\citep{ToolBench}, and therefore abstract away the uncertainty introduced by retrieved tool.
Together, these limitations leave open a central question: \textbf{Can LLM agents solve long-horizon tasks in large tool ecosystems by iteratively exploring partial tool retrieval results and adapting when plausible tool-use paths fail?} 
Evaluating this requires a benchmark with:
\textbf{(1) Partial tool visibility}, where agents access only retrieved subsets of a large tool space and must iteratively discover tools for intermediate information; and
\textbf{(2) Unreliable and noisy tools}, where retrieved tools may be missing, failing, or misleading, requiring adaptation when plausible tool-use paths break.


To this end, we introduce \BENCH{}, an interactive and dynamic tool-use benchmark situated in the retail domain. 
\bench{} consists of 327 distinct evaluation instances built over 1,665 tools, with each instance designed as a multi-step retail workflow that requires approximately 25 turns on average.
Unlike settings where agents are given a fixed and fully visible toolset, \bench{} places agents in a retrieval-mediated environment where useful tools must be discovered during problem solving.
By design, tool discovery in \bench{} is structured around \textbf{bi-directional anticipation}: (1) \textit{Forward Anticipation}: agents may search forward from accumulated evidence, (2) \textit{Backward Anticipation}: search backward from desired outcomes, or bridge known and hypothesized states during planning.
This mirrors human problem solving, where people often reason from both the current state and the desired goal, forming intermediate sub-goals to close the gap~\citep{newell1972human}.
To model unreliable tool environments, PlanBench-XL includes documented noisy tools in both default and block settings, where some retrieved tools explicitly indicate that they may return irrelevant, outdated, or erroneous outputs, reflecting the noise among functionally similar tools in real-world ecosystems.
On top of this base noise, we design a optional retrieval-time blocker module that replaces path-critical tools with explicit or implicit failures, forcing agents to identify disrupted tool-use paths and adaptively re-plan as execution unfolds.

We evaluate ten leading open-source and proprietary LLMs on \bench{}. 
Our results show that long-horizon planning over massive tool ecosystems remains highly challenging. 
While most models remain below two-thirds accuracy in the default setting, performance drops sharply under retrieval-time blocking, with \textit{GPT-5.4} falling to around 30\% when only one feasible path remains and to slightly above 10\% when only the longest recovery path is preserved.
We further find that success requires not only broad exploration, but also precise exploitation of discovered information and reliable tool invocation. 
These findings suggest that current LLM agents still lack robust adaptive planning capabilities in large, partially observable, and unreliable tool environments.
We summarize our main contributions as follows:
\begin{itemize}[topsep=3pt, partopsep=1.5pt, leftmargin=*, itemsep=-4.5pt]
    \item \textbf{Scalable Benchmark Framework:} We introduce \bench{}, a scalable LLM-based task generator that automatically creates grounded queries with paired tools, creating diverse, complex tasks to evaluate agents’ planning abilities.
    
    \item \textbf{Dynamic Interaction Environment:} We design an interactive environment that simulates real-world unpredictability with three blocking events, forcing real-time adaptation and re-planning, and can serve as an RL playground for agent training.
    
    \item \textbf{Comprehensive Analysis:} We evaluate ten frontier LLMs and reveal that current agents struggle with massive-tool planning, especially under corrupted tool access and longer recovery paths.
\end{itemize}
Looking ahead, \BENCH{} lays the foundation for robust and adaptive LLM agents that actively explore large tool ecosystems, detect unreliable tool access, and re-plan under real-world uncertainty.

\section{Related Work}

\begin{figure*}[t]
  \centering
  \includegraphics[width=1\linewidth]{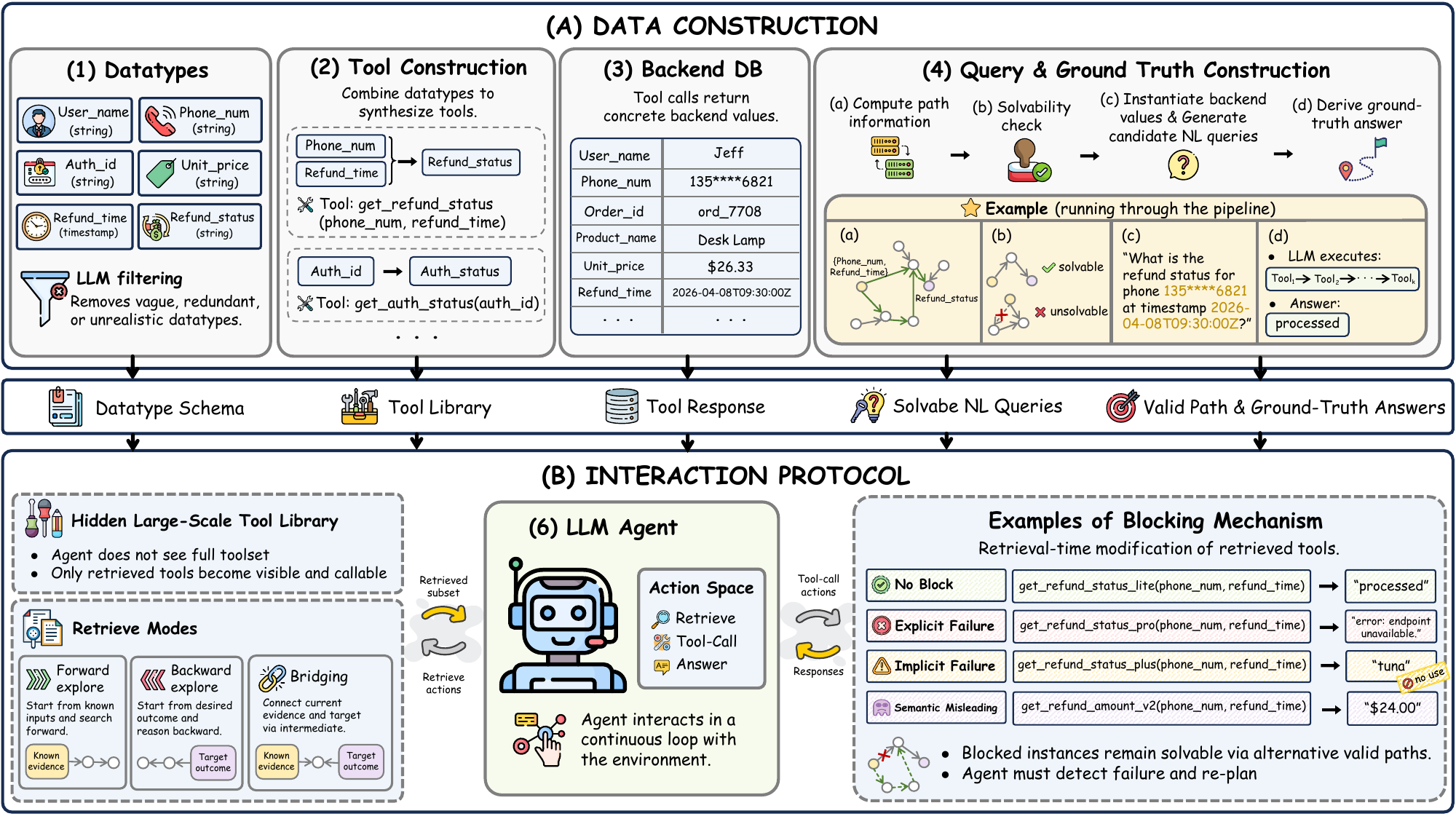}
  \vspace{-0.25in}
  \caption{Overview of \bench{}. \textit{\textbf{Data construction}} turns typed retail datatypes into executable tools and backend records, then derives solvable queries with ground-truth answers. \textit{\textbf{Runtime protocol}} evaluates agents as they retrieve and invoke tools via bidirectional exploration, while retrieval-time blockers force re-planning.}
  \label{fig:demo_figure}
  \vspace{-0.15in}
\end{figure*}

\paragraph{Evaluating Agentic Planning with Large-Scale Toolsets.}







Planning over large-scale tool ecosystems is central to complex agentic tasks, where agents must compose multi-step tool-use trajectories while handling unreliable tool access. 
Existing benchmarks have examined this challenge from several perspectives. 
Large-scale tool-use and retrieval benchmarks evaluate tool selection over broad tool collections~\citep{ToolBench,APIBench,API-Bank,ToolRet,MCP-Bench,LiveMCPBench}, but often focus on relatively explicit task goals. 
Long-horizon benchmarks cover stateful execution~\citep{ToolSandbox}, multi-hop chaining~\citep{ToolHop}, implicit-goal solving~\citep{EscapeBench}, and extended workflows~\citep{Tool-Decathlon}, but generally assume available tools rather than unreliable, retrieval-mediated access.
Robustness-oriented benchmarks introduce noisy users~\citep{AgentNoiseBench,UserBench}, imperfect tools~\citep{OpaqueToolsBench,ToolQA-D}, or tool-state failures~\citep{ToolGym,CostBench}, but rarely combine adaptation with active search over large tool spaces. 
Thus, our benchmark evaluates adaptive planning under partial tool observability and unreliable tool availability, requiring agents to explore bi-directionally and adapt when tool-use paths are disrupted.

\paragraph{Agent Designs for Large-Scale Tool Planning.}




Prior work has developed various agent designs for planning with large-scale tool ecosystems. 
Among them, some single-agent frameworks focus on modular tool invocation \citep{karpas2022mrkl, schick2023toolformer}, reasoning--acting control and long-horizon planning \citep{erdogan2025planact, ReAct, koh2025treesearch}, and scalable tool selection from large tool libraries \citep{AnyTool, ToolGen, AutoTool}. 
Multi-agent systems further reduce long-horizon cognitive load through communicative collaboration \citep{li2023camel, wu2024autogen, chen2024agentverse} and role- or workflow-based task decomposition \citep{hong2024metagpt, chen2025smurfs, ling2025elhplan}. 
However, even these designs primarily scaffold tool invocation, tool selection, and task decomposition, while leaving unresolved the core challenge posed by our benchmark: agents must actively explore implicit sub-goals and intermediate information, and then adaptively re-plan as observations emerge.

\section{\bench{}}
\label{sec:benchmark}

\BENCH{} is an interactive benchmark designed to evaluate LLM agents’ ability to explore and plan in massive tool-use environments.
Unlike conventional tool-use settings, \bench{} requires agents to infer implicit sub-goals and discover useful tool-use paths.
This setting reflects real-world agent applications, where models often lack full tool visibility and must explore tools to uncover useful intermediate information.
As demonstrated in Figure~\ref{fig:demo_figure}, \bench{} automatically constructs a broad ecosystem of task-grounded tools and exposes it through a retriever.
The retriever supports bi-directional exploration, enabling forward anticipation from available evidence and backward anticipation from the target outcome.
This design requires agents to explore the tool space while inferring intermediate information.
Moreover, \bench{} supports two evaluation modes: a \textbf{default mode} that evaluates tool use under ordinary retrieval noise, with useful tools mixed among distracting alternatives; and a \textbf{block mode}, which further disrupts selected useful paths while preserving solvability.
It therefore tests whether agents can recover and re-plan when a plausible path becomes unreliable.
Each task unfolds through an iterative interaction loop in which the agent observes the state, reasons about its next step, retrieves or invokes tools, and receives executable feedback from the environment. 
We describe the benchmark construction below and discuss its significance and generalization in Appendix~\ref{app:significance-generalization}.



\subsection{Environment Setup}

\paragraph{Tool Library.}
We introduce self-defined datatypes to provide a typed representation of domain-specific information and to support the systematic construction of tools.
For the retail domain, we first define a set of datatypes $\mathcal{D}$.
Each datatype $d \in \mathcal{D}$ represents a concrete type of distinct domain information, such as \texttt{person\_name} or \texttt{purchase\_status}.~The initial datatype inventory is proposed by a generation LLM $M_{\mathrm{gen}}$ and then automatically filtered by another LLM $M_{\mathrm{fil}}$ to remove vague, redundant, unreasonable or unrealistic datatypes. We then construct~candidate tools by considering all pairs of datatype sets $(\mathcal{D}_{\mathrm{in}}, \mathcal{D}_{\mathrm{out}})$ over $\mathcal{D}$, where $|\mathcal{D}_{\mathrm{in}}|{=}m$ and $|\mathcal{D}_{\mathrm{out}}|{=}n$ denote the given numbers of input and output datatypes, respectively.
For each such pair, $M_{\mathrm{gen}}$ proposes a candidate tool functionality whose input schema is given by $\mathcal{D}_{\mathrm{in}}$ and whose output schema is given by $\mathcal{D}_{\mathrm{out}}$.
For each~resulting tool $\tau$, we denote its input and output datatype sets by $\mathcal{D}_{\mathrm{in}}(\tau)$ and $\mathcal{D}_{\mathrm{out}}(\tau)$.
For example,~given
$\mathcal{D}_{\mathrm{in}}{=}\{\texttt{product\_name}, \texttt{store\_name}\}$
and
$\mathcal{D}_{\mathrm{out}}{=}\{\texttt{inventory\_status}\}$,
$M_{\mathrm{gen}}$ may propose a tool functionality that checks whether the specified product is available at the specified store and returns its inventory status. 
The final tool library $\mathcal{T}$ is obtained by applying $M_{\mathrm{fil}}$ to remove unreasonable, redundant, or undesired candidates. 
We further augment the functional tool library with paired noisy tools to simulate realistic ecosystems where functionally similar tools may include distractors~\citep{huang2024metatoolbenchmarklargelanguage}. These noisy tools are semantically similar to previously constructed tools, but explicitly disclose their unavailability or unreliability in the descriptions, allowing careful agents to reject them after inspection.
We provide executable tool construction details in Appendix~\ref{app:schema-construction-details}, filtering rules in Appendix~\ref{app:detailed-filtering-criteria}, and noisy tool construction details in Appendix~\ref{app:noisy_tool}.


\paragraph{Tool Response Construction.}
We further construct a backend database to support tool responses with concrete data instances. 
For each retail case, we prompt the generation LLM $M_{\mathrm{gen}}$ to instantiate values for the full datatype set in the retail domain, producing a complete structured record (to ensure the completeness of the backend). 
During execution, a tool $\tau$ matches its input arguments to the corresponding record and returns the values for its output datatype set $\mathcal{D}_{\mathrm{out}}(\tau)$.
We ensure that the instantiated values are non-trivial and cannot be inferred from common sense, making tool use necessary for deriving the requested output values.
Appendix~\ref{app:tools-necessity-details} details backend construction and why answers require tool-derived values.
For noisy tools, we assign fixed return values at construction time. These responses are shared across models and query instances and are designed to provide no useful evidence for solving the task (details provided in Appendix~\ref{app:noisy_tool}).

\paragraph{Queries.}
Figure~\ref{fig:demo_figure} gives an example of this pipeline.
We generate queries through a three-step pipeline.
First, we specify each task internally as $r=(\mathcal{D}_0, \mathcal{Y})$, where $\mathcal{D}_0 \subseteq \mathcal{D}$ denotes the initial input datatype set and $\mathcal{Y} \subseteq \mathcal{D}$ denotes the target datatype set.
Second, we compute the set $\Pi(r)$ of ground-truth tool-call sequences that transform $\mathcal{D}_0$ into $\mathcal{Y}$ via code, and discard any task with $\Pi(r)=\emptyset$.
We formalize this internal reachability structure as a state graph and use it for path enumeration and solvability checks (details are provided in Appendix~\ref{app:state-graph}).
Third, we instantiate each remaining task with concrete entities and attribute values by sampling values from the backend while ensuring that the resulting instance is solvable, and use $M_{\mathrm{gen}}$ to verbalize the instantiated task as a natural-language query $q$.
Given $q$ and step-by-step tool-call instructions, we then prompt $M_{\mathrm{gen}}$ to produce an answer $o^\star$ following one valid sequence $\pi \in \Pi(r)$, and use $o^\star$ as the ground truth.
Finally, we retain only instances whose shortest valid solution requires at least five distinct tool calls.
We ensure tool and datatype quality through partial manual inspection, with details in Appendix~\ref{app:human-annotation}, and provide concrete data cases in Appendix~\ref{app:data-case-study}.
Details on selecting $M_{\mathrm{gen}}$ and $M_{\mathrm{fil}}$, query filtering, and ground-truth construction are provided in Appendices~\ref{app:model-choice}, \ref{app:query-construction-details}, and \ref{app:groundtruth-details}, respectively.

\subsection{Agent--Environment Interaction}

Each query is executed as a multi-turn interaction between the agent and the environment.
At each step, the agent outputs exactly one of three actions:
retrieving candidate tools, calling a retrieved tool with structured arguments, or returning the final answer, denoted by
$a_t \in \{\texttt{retrieve}, \texttt{tool-call}, \texttt{answer}\}$.
The environment responds according to the chosen action:
\textbf{(1)} for \textbf{\texttt{retrieve}}, it returns candidate tools and an additional note if the requested tool does not exist;
\textbf{(2)} for \textbf{\texttt{tool-call}}, it executes the call on the constructed backend and returns the result;
\textbf{(3)} for \textbf{\texttt{answer}}, the trajectory terminates and the final answer is evaluated.
The trajectory also terminates when a predefined, agent-visible budget $T_{\max}$ is exhausted.
To prevent pure guessing or other shortcuts, we apply datatype checking to ensure that the final answer is derived from a valid \texttt{tool-call} result. 
Please refer to Appendix~\ref{app:runtime-interaction-details} for more details.

\paragraph{Agent State Definition.} At runtime, the environment maintains an explicit agent state. 
We define the state at step $t$ as $s_t = (q, \mathcal{U}_t, \mathcal{D}_t),$ where $q$ is the user query, $\mathcal{U}_t$ is the set of all tools discovered so far in the current query and available for the agent to call, and $\mathcal{D}_t \subseteq \mathcal{D}$ is the set of datatypes already obtained through successful tool calls. 
The initial state is determined by the datatypes used to construct the query, with initial evidence $\mathcal{D}_0$ and $\mathcal{U}_0=\varnothing$. 
Whenever a tool call succeeds and outputs values with datatype set $\mathcal{D}_{\mathrm{out}}(\tau)$, the environment updates $\mathcal{D}_{t+1}=\mathcal{D}_t \cup \mathcal{D}_{\mathrm{out}}(\tau)$.
The agent's objective is to obtain the target information specified in the query and return the correct final answer through tool use.
Note that the state $s_t$ is an environment-maintained latent state which is used to track progress across tool calls, but is not directly visible to the agent.

\paragraph{Tool Retriever.}

Tool-use in \bench{} is supported by a retriever that lets the agent explore the tool space through high-level natural-language queries rather than by inspecting the full tool library. 
At each step $t$, the agent may query the retriever in three complementary modes aligned with bi-directional anticipation:
\textbf{(1) Input-conditioned retrieval}, which supports \textbf{\textit{Forward Anticipation}} by asking what information or tool affordances can be reached from the evidence currently available;
\textbf{(2) Output-conditioned retrieval}, which supports \textbf{\textit{Backward Anticipation}} by asking what tools or prerequisite information may lead to a desired intermediate or final outcome;
and \textbf{(3) Input-output-conditioned retrieval}, which constrains the search by specifying both the available information and the desired result. 
The retriever grounds these queries to tool signatures and adds matched tools to the agent-callable set \(\mathcal{U}_t \subseteq \mathcal{T}\) for subsequent turns.
Implementation details of request matching and retrieval modes are provided in Appendix~\ref{app:retriever-details}.
To ensure that retrieval is reliable under natural-language queries, we conduct a single-step retrieval study and report the results in Appendix~\ref{app:retriever-robustness}.

\paragraph{Retrieval-Time Blocking.}

Our benchmark includes an optional blocking module used during retrieval.
Before evaluation, the environment identifies a set of tools $\mathcal{T}_{\mathrm{blk}} \subseteq \mathcal{T}$ that lie on valid solution paths and should be blocked, while keeping this information hidden from the agent.
During retrieval, if a tool $\tau \in \mathcal{T}_{\mathrm{blk}}$ is retrieved as a blocked candidate, the environment replaces it with semantically similar alternatives $\tau'$ under the standard interaction protocol.
Since each blocked tool is unavailable to the agent, any solution path that depends on at least one tool in $\mathcal{T}_{\mathrm{blk}}$ becomes infeasible from the agent's perspective.
By construction, each blocked instance preserves at least one feasible tool-call path, and therefore remains solvable.
The path-preserving selection procedure is described in Appendix~\ref{app:blocked-tool-selection}.
Appendix~\ref{app:blocker-details} details the blocking pipeline, including additional tool construction in Appendix~\ref{app:noisy-tool-construction}.
We consider three types of retrieval-time blocking perturbations:
    
    

\begin{itemize}[topsep=2pt, partopsep=3pt, leftmargin=*, itemsep=-3pt]
    \item \textbf{Explicit Failure Blocks:}
    the returned tool $\tau'$ produces an explicit error message, such as \texttt{error: endpoint unavailable}.

    \item \textbf{Implicit Failure Blocks:}
    the returned tool $\tau'$ produces an unhelpful response that silently violates its documented behavior.

    \item \textbf{Semantically Misleading Blocks:}
    the returned tool $\tau'$ has related but different functionality, making it appear usable as a substitute.
\end{itemize}



For example, if~\texttt{get\_refund\_status(order\_id)} is blocked, the retriever replaces it with either an tool returning explicit error, a tool returning an~implicit wrong value such as \texttt{refund\_status=tuna}, or a semantically misleading tool such as \texttt{get\_order\_status(order\_id)} (formal definitions of the blocking mechanism in Appendix~\ref{app:blocking-formalization}).
The mechanism ensures the following:
(1) \textit{Evaluation Fairness}: Within the same setting, all models face the same blocked tools, enabling fair comparison (details are shown in Appendix~\ref{app:evaluation-fairness}).
(2) \textit{Independence Across Instances}: Blocked tools are sampled independently for each task instance, even within the same run.
(3) \textit{Controlled Difficulty}: The benchmark varies the number of replaced baseline tools, the number of injected alternatives, and the type of blocking to control disruption severity.
This setting reflects realistic tool retrieval failures: tools may be explicitly broken, silently unreliable, or semantically similar but functionally wrong. 
\bench{} therefore tests whether agents can detect these failure signals, avoid misleading tools, and recover via alternative paths.
We illustrate a blocking example in Figure~\ref{fig:demo_figure} and compare the three blocking types in Table~\ref{tab:blocking-comparison-tab}.


\section{Experiment}

\begin{table}[t]
\begin{center}
\small
\begin{tabular}{lc}
\toprule
\textbf{Parameter} & \textbf{Value} \\
\midrule
Number of datatypes & 56 \\
Number of queries & 327 \\
Number of tools & 1,665 \\
Shortest path length ($L^*$) & 5 / 6 / 7 / 8 / 9 \\
Maximum turns ($T_{\max}$) & 100 \\
Per-retrieval return cap ($\Lambda_{\mathrm{ret}}^{\mathrm{cap}}$) & 30  \\
Global seed ($\sigma_0$) & 42 \\
\bottomrule
\end{tabular}
\end{center}
\vspace{-0.13in}
\caption{Environment details. Global seed $\sigma_0$ ensures reproducible stochasticity.}
\label{tab:default_env_settings}
\vspace{-0.25in}
\end{table}

\begin{table*}[t]
\begin{center}
\tabcolsep=0.015\linewidth
\resizebox{0.97\linewidth}{!}{
\begin{tabular}{lccccccc}
\toprule
\multicolumn{1}{c}{\multirow{2}{*}{\textbf{Model Name}}} 
& \multicolumn{3}{c}{\textbf{Task Completion}} 
& \multicolumn{2}{c}{\textbf{Exploration Behavior}} 
& \multicolumn{2}{c}{\textbf{Execution Quality}} \\
\cmidrule(lr){2-4} \cmidrule(lr){5-6} \cmidrule(lr){7-8}
& Accuracy (\%) $\uparrow$ 
& EGT Prec. (\%) $\uparrow$
& Avg. Turns
& Mean EDT 
& S/C Ratio 
& ITCR (\%) $\downarrow$
& UIRR (\%) $\downarrow$ \\
\midrule
\textit{Qwen3-8B} & \phantom{0}0.00 & 35.31 & 25.65 & \phantom{0}7.64 & \phantom{0}0.20 & \phantom{0}6.11 & \underline{\phantom{0}0.10} \\
\textit{Qwen3-14B} & \phantom{0}0.92 & 47.77 & 35.74 & 12.01 & \phantom{0}0.09 & \phantom{0}3.94 & \phantom{0}0.93 \\
\textit{Qwen3-32B} & \phantom{0}2.75 & 62.36 & 12.03 & 18.54 & \phantom{0}1.59 & 10.05 & \phantom{0}7.43 \\
\textit{Llama-3.1-8B-Instruct} & \phantom{0}0.00 & 41.33 & 21.62 & \phantom{0}9.89 & \phantom{0}1.49 & 18.03 & \phantom{0}5.25 \\
\textit{Llama-3.3-70B-Instruct} & 18.96 & 59.67 & 19.13 & 19.20 & \phantom{0}2.22 & 21.47 & \phantom{0}2.13 \\
\textit{DeepSeek-V4-Flash} & \underline{63.08} & 65.57 & 31.41 & \underline{25.34} & \phantom{0}2.80 & \phantom{0}8.27 & \phantom{0}3.29 \\
\textit{Gemini-3.1-Pro} & \textbf{77.06} & \textbf{91.47} & 19.55 & \textbf{27.41} & \phantom{0}1.59 & \textbf{\phantom{0}0.68} & \phantom{0}0.30 \\
\textit{Gemini-3.5-Flash} & 52.19 & \underline{85.29} & 57.87 & 25.16 & 10.44 & \underline{\phantom{0}2.94} & \textbf{\phantom{0}0.00} \\
\textit{GPT-5.4-Mini} & \phantom{0}3.07 & 71.25 & 10.81 & \phantom{0}9.22 & \phantom{0}1.97 & 51.71 & \phantom{0}4.42 \\
\textit{GPT-5.4} & 51.90 & 72.92 & 22.92 & 20.65 & \phantom{0}2.70 & \phantom{0}6.28 & \phantom{0}1.91 \\
\bottomrule
\end{tabular}
}
\end{center}
\caption{\bench{} main evaluation results in the \textit{default} setting under the grounded accuracy criterion.Scores in \textbf{bold} indicate the best performance among all models, and \underline{underlined} scores denote the second-best performance. For \textit{Mean EDT}, higher values generally indicate broader exploration but are not strictly better in all cases; we still highlight the top two values to aid readability.The metrics in (\%) are reported in percentage form.}
\label{tab:main_unblocked_results_grounded}
\end{table*}






\subsection{Settings}

\paragraph{Models.}

We evaluate both proprietary and open-source models to ensure comprehensive assessment of current frontier LLM capabilities.
Proprietary models include 
\textit{GPT}~\citep{GPT-5.4}, and \textit{Gemini}~\citep{Gemini-3.1-Pro}, while open-source models include \textit{Qwen3}~\citep{Qwen3}, \textit{Llama3}~\citep{Llama-3} and \textit{Deepseek}~\citep{DeepSeek-V4}. 
All models use a temperature of 0.0 for deterministic decoding and a max token length of 8192 to prevent truncation.

\paragraph{Metrics.}
 
We evaluate performance using metrics spanning three categories: task completion, exploration behavior and execution quality.
Together, these metrics provide a holistic view of model performance in massive-tool environments.
Full metric definitions and illustrative examples are provided in Appendix~\ref{app:full-metric-definition} and Appendix~\ref{app:metric-illustrative-examples}.

\noindent \textbf{(1) Accuracy (\%):} It measures the proportion of queries with a correct final answer.


\noindent \textbf{(2) Executed Ground-Truth Datatype Precision (EGT Prec.):} This metric measures the fraction of unique datatypes produced by executed tool calls that belong to the ground-truth datatype set, indicating how often execution stays relevant.

\noindent \textbf{(3) Average Turns (Avg. Turns):} This metric measures the average interaction turns per query.

\noindent \textbf{(4) Mean Explored Datatypes (Mean EDT):} This metric measures the average number of new datatypes uncovered through tool retrieval beyond each query’s initial input datatypes.

\noindent \textbf{(5) Search-to-Call Ratio (S/C Ratio):} This metric quantifies the tool-retrieval/tool-call turn ratio, capturing the exploration–exploitation balance.


\noindent \textbf{(6) Invalid Tool Call Rate (ITCR) (\%):} This metric measures the fraction of structurally or procedurally invalid calls, such as using unretrieved tools, mismatched arguments, or unavailable inputs.

\noindent \textbf{(7) Untrusted Input Rejection Rate (UIRR) (\%):} This metric measures the fraction of tool-call attempts rejected because at least one argument value is taken from a noisy tool response.



\paragraph{Prompts and Environment Settings.}
Table~\ref{tab:default_env_settings} summarizes the main benchmark and environment parameters, and Figure~\ref{fig:inference-prompt} shows the inference prompt.
The benchmark contains $56$ datatypes, $327$ evaluation queries, and $1{,}665$ tools.
Here, $L^*$ denotes the shortest valid solution length in tool calls ($5$--$9$), $T_{\max}=100$ denotes the interaction budget, $\Lambda_{\mathrm{ret}}^{\mathrm{cap}}=30$ denotes the maximum number of tools returned per retrieval, and $\sigma_0=42$ denotes the global random seed.



\subsection{Results}


\begin{takeaway}
Planning with massive toolsets remains highly challenging for most models.
\end{takeaway}

\noindent Table~\ref{tab:main_unblocked_results_grounded} shows a sharp divide between frontier models and the rest. 
\textit{Gemini-3.1-Pro} achieve the highest accuracy of 77.06\%, while maintaining the highest EGT Precision with about 20 turns. 
In contrast, most other models remain below 60\% accuracy, with \textit{Qwen3-8B} and \textit{Llama-3.1-8B-Instruct} achieving 0\%, revealing the difficulty of long-horizon, exploration-driven planning across large tool ecosystems.
The gaps are also large within model families: larger Qwen and Llama variants outperform their smaller counterparts, and full frontier models far exceed their lightweight versions such as \textit{Gemini-3.5-Flash} and \textit{GPT-5.4-Mini}. 
Together, these results suggest that both model family and scale matter for this task.
Robustness checks, including confidence intervals, are reported in Appendix~\ref{app:evaluation-robustness}.

\begin{takeaway}
Tool discovery requires broad, accurate, and bi-directional exploration.
\end{takeaway}


\paragraph{Exploration tendency strongly relates to task success.}
Broad tool retrieval can expose agents to more potential intermediate information, making exploration tendency strongly associated with task success.
We measure this exploration tendency using Mean EDT, the average number of new datatypes uncovered through retrieval beyond the query's initial datatypes. 
Across models, Mean EDT is strongly correlated with accuracy (Pearson coefficient~\citep{Pearson} $r=0.902$), suggesting that agents uncover more intermediate information are generally more likely to complete the task successfully. 
However, exploration tendency does not explain performance entirely. 
For example, \textit{Llama-3.3-70B-Instruct} achieves a Mean EDT score comparable to \textit{GPT-5.4} (19.20 vs. 20.65), yet the two models differ substantially in accuracy (18.96\% vs. 51.90\%).
This indicates that broad retrieval is an important driver of success, but it remains only part of effective long-horizon tool~use. 



\paragraph{Frequent retrieval does not guarantee effective exploration.}
A high S/C Ratio and a large number of interaction turns indicate that an agent spends substantial effort on retrieval, but such effort does not necessarily translate into broad discovery of useful intermediate information.
For example, \textit{Gemini-3.5-Flash} has the highest S/C Ratio among all models (10.44) and also uses the most turns on average (57.87).
However, \textit{Gemini-3.1-Pro} achieves a higher Mean EDT while using only about one third of the turns and one seventh of the S/C Ratio.
This contrast shows that frequent searching and long interactions are not sufficient for effective exploration.
An agent may search proactively, but if these searches repeatedly revisit unuseful or uninformative tools, they contribute little to the discovery of new task-relevant datatypes.



\paragraph{Effective tool discovery requires bi-directional anticipation.}
We define the forward/backward retrieval ratio (F/B ratio) as the total number of input-conditioned retrievals divided by the total number of output-conditioned retrievals across all query traces for a model. 
Across models, agents generally issue more input-conditioned than output-conditioned retrievals, indicating a common preference for forward anticipation. 
Lower-performing models, such as \textit{Llama-3.1-8B-Instruct} and \textit{Qwen3-14B}, rely heavily on input-conditioned retrieval, yielding F/B ratios of 16.56 and 14.18, respectively. 
This suggests that retrieving tools compatible with currently available inputs is often insufficient: these agents may identify executable next steps, but fail to reason backward about which intermediate datatypes must be discovered to reach the final goal.
This pattern is further supported by correlation analysis, where the relative frequency of output-conditioned retrieval is strongly correlated with accuracy (Pearson $r=0.800$). 
Together, these results suggest that effective tool discovery requires agents to combine forward exploration from current evidence with backward anticipation from the desired outcome.

\begin{takeaway}
Successful long-horizon tool use requires both effective exploration and accurate exploitation.
\end{takeaway}

\noindent Beyond effective exploration, successful agents must accurately exploit the information they uncover, which EGT Precision captures by measuring whether executed tool calls stay on task-relevant paths. 
EGT Precision is highly correlated with accuracy (Pearson coefficient $r=0.781$), indicating that models are much more likely to succeed when they execute relevant tool-use trajectories.
The strongest models further support this pattern: \textit{Gemini-3.1-Pro} achieve the highest accuracies (77.06\%) while also attaining the highest EGT Precision (91.47\%). 
These results suggest that effective long-horizon tool use requires not only sufficient exploration, but also precise execution over the explored tool space.

\begin{takeaway}
Basic tool-use reliability remains necessary.
\end{takeaway}

\noindent Accurate exploration and exploitation also depend on reliable tool-use. 
Accuracy is negatively correlated with ITCR (Pearson coefficient $r=-0.443$), showing that models that frequently make invalid tool calls are much less likely to complete the task successfully. 
For example, \textit{Llama-3.1-8B-Instruct} has the second highest ITCR and the lowest accuracy, while the strongest model, \textit{Gemini-3.1-Pro}, keep ITCR near zero (0.68\%). 
This suggests that effective long-horizon tool-use requires not only broad exploration and accurate exploitation, but also basic reliability in invoking tools with valid arguments.
The moderate correlation also indicates that invalid tool calls alone do not explain performance differences, and that failures also arise from ineffective exploration and exploitation.

\section{Analysis}
\label{sec:analysis}

In this section, we analyze factors that influence model performance from three complementary
perspectives. As described in Section~\ref{sec:benchmark}, our task and tool construction is 
structured, enabling controlled analysis along several dimensions:
\begin{itemize}[topsep=2pt, partopsep=3pt, leftmargin=*, itemsep=-3pt]
    \item \textbf{Block Types and Severity} (Takeaway~\ref{tak:analysis-block-type-severity}):
    How agents perform under different types and severities of tool-path blocking.

    \item \textbf{Inference-Time Augmentation} (Takeaway~\ref{tak:analysis-test-time-tricks}):
    Whether additional test-time computation improves adaptation under blocked tool paths.

    \item \textbf{Path Length Effects} (Takeaway~\ref{tak:analysis-path-length}):
    How minimal solution paths affect task accuracy.
\end{itemize}


\begin{figure*}[t]
  \centering
  \includegraphics[width=1\linewidth]{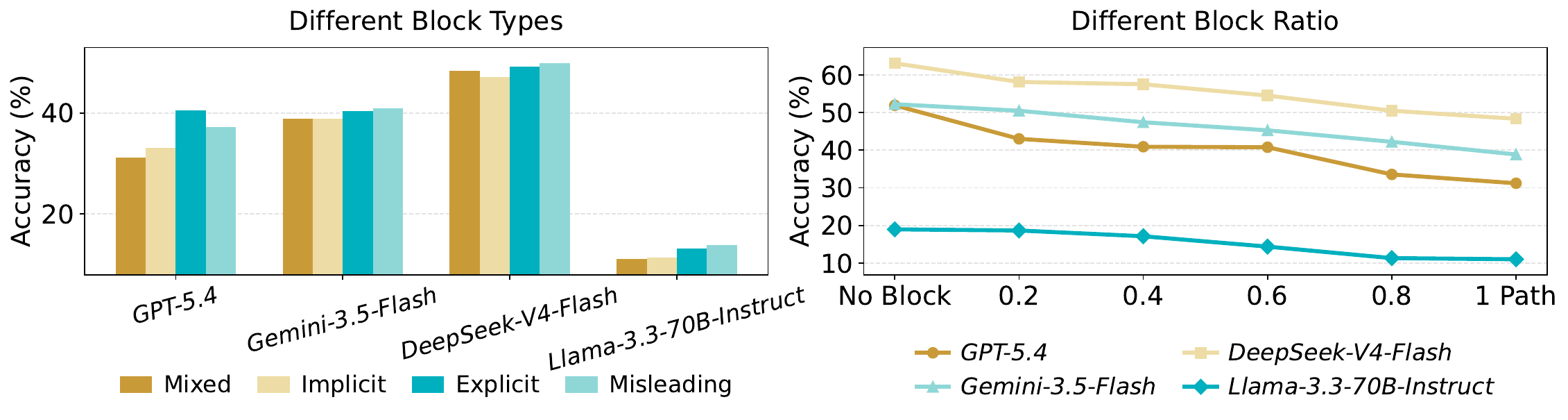}
  \vspace{-0.25in}
  \caption{Accuracy under retrieval-time blocking. \textit{\textbf{Left}}: performance across different block types when only one feasible solution path remains, including mixed blockers (\textit{Mixed}), implicit failures (\textit{Implicit}), explicit failures (\textit{Explicit}), and semantic distractions (\textit{Misleading}). \textit{\textbf{Right}}: performance as blocking becomes stronger, ordered from no-block to block ratios of 0.2, 0.4, 0.6, and 0.8, with “1 Path” leaving only a single feasible path.}
  \label{fig:noise_combined}
  \vspace{-0.1in}
\end{figure*}

\begin{figure}[t]
    \centering
    \includegraphics[width=1\linewidth]{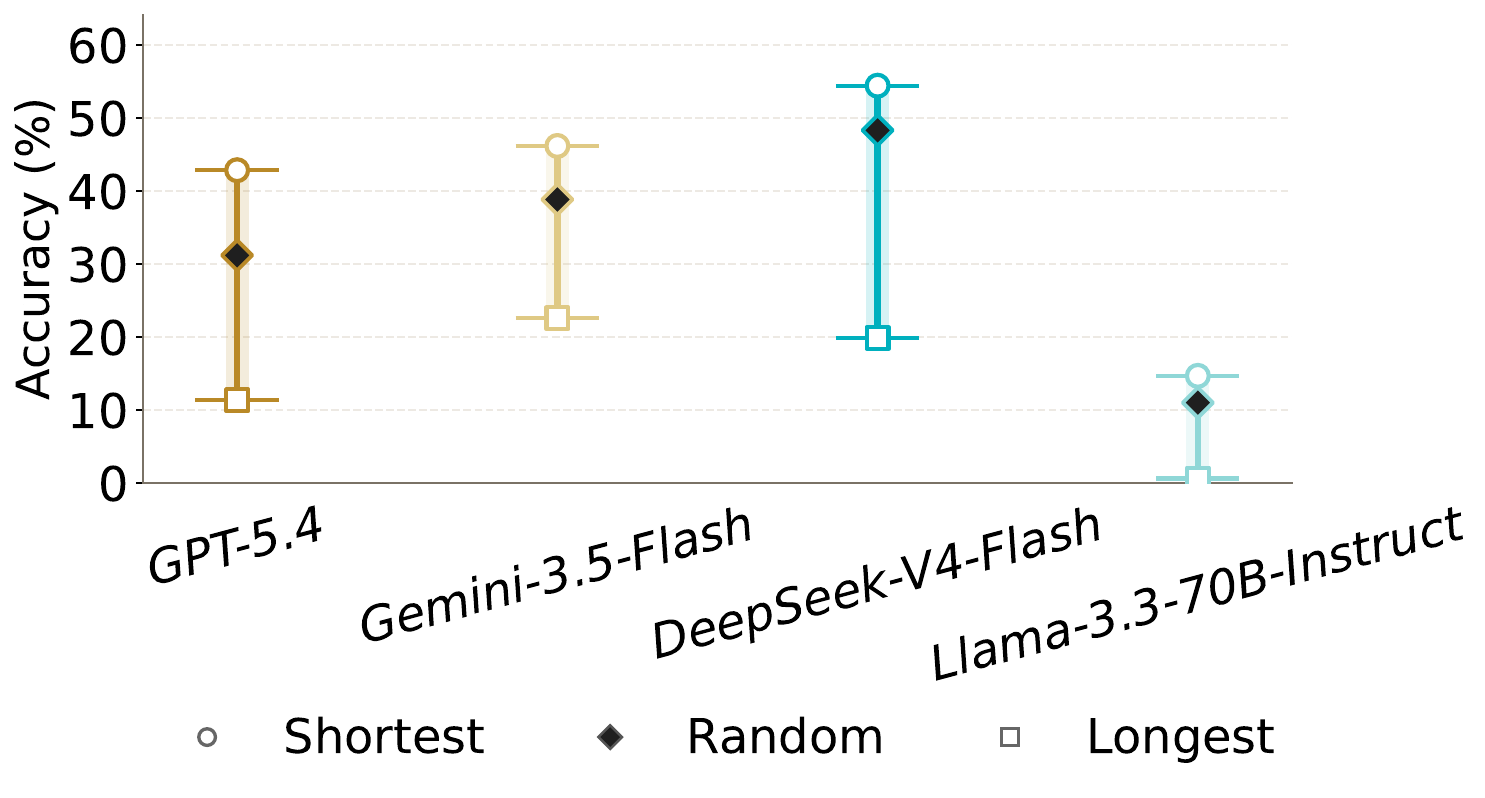}
    \vspace{-0.15in}
    \caption{Accuracy (\%) on \bench{} under fine-grained blocking. \textit{Shortest} and \textit{Longest} denote runs where blocking preserves the shortest and longest admissible solution paths, respectively; \textit{Random} denotes the basic blocked configuration. Circles, squares, and diamonds mark these three settings.}
    \label{fig:fine_grained_blocking}
    \vspace{-0.1in}
\end{figure}

\begin{figure*}[t]
    \centering
    \includegraphics[width=1\linewidth]{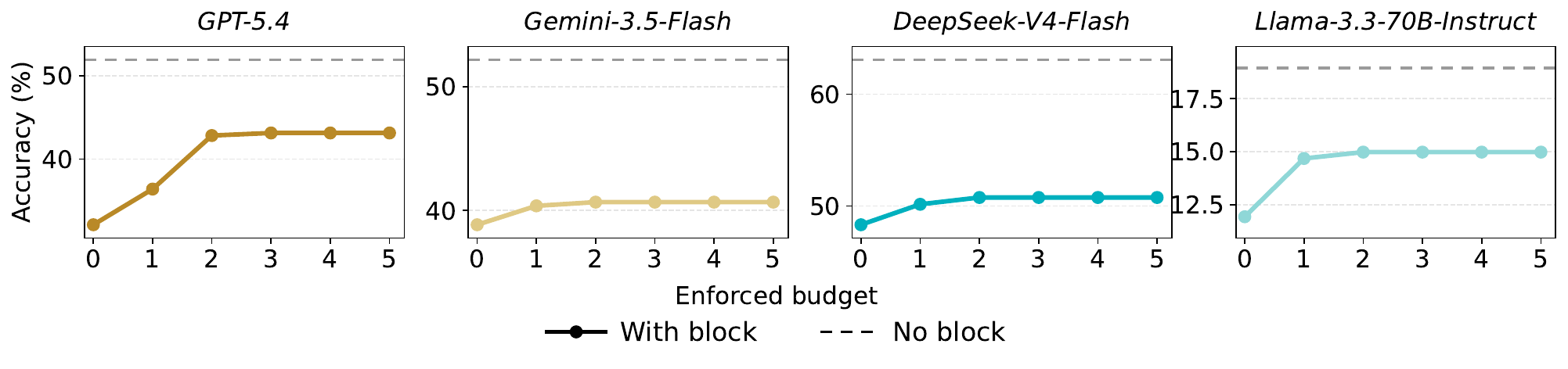}
    \vspace{-0.35in}
    \caption{Effect of enforced exploration under blocking. The enforced budget $B_{\mathrm{enf}}$ is the maximum number of continuation prompts added after incorrect termination, with $B_{\mathrm{enf}}=0$ denoting the standard block setting and $B_{\mathrm{enf}}=5$ the largest budget tested. Accuracy gains quickly saturate and remain far below the no-block accuracy shown by the dashed line.}
    \label{fig:enforced_budget_preview}
\end{figure*}


\begin{takeaway}[label={tak:analysis-block-type-severity}]
Agents are vulnerable under corrupted tool environments, especially when failures are silent or alternative paths become longer.
\end{takeaway}

\paragraph{Viable-path reduction sharply weakens performance.}
We vary the block ratio, defined as the proportion of originally feasible solution paths disabled by replacing selected path-critical tools with block alternatives. 
As shown in the right panel of Figure~\ref{fig:noise_combined}, as the block ratio increases, all four models perform worse, showing that agents become less reliable as the environment leaves fewer viable paths available. 
This degradation is especially pronounced for \textit{GPT-5.4}, whose accuracy drops by more than 20 percentage points, and the same trend is consistently observed across the other three models.
These results suggest that stronger environmental constraints systematically weaken adaptive planning by forcing agents to recover within a smaller solution space.



\paragraph{Silent tool failures are the most~harmful.}
To isolate the effect of each block type, we keep the block tools and block ratio fixed, and vary only the behavior of the replacement tools. 
The left panel of Figure~\ref{fig:noise_combined} further breaks down performance by block type. 
Among the single-type perturbations, excluding the mixed condition, implicit failures lead to the lowest accuracy for all selected models. 
This suggests that silent failures are especially disruptive: unlike explicit errors or semantically mismatched tools, they provide weak failure signals and are therefore harder for agents to recognize.
UIRR provides a quantitative view of this failure pattern.
On average, UIRR is highest under implicit failures ($11.99\%$), compared with explicit failures ($9.67\%$) and misleading tools ($9.89\%$).
This indicates that agents are more likely to reuse values returned by silently failed tools as inputs to later tool calls, allowing the failure to propagate along the trajectory.
As a result, agents are less likely to recognize that the current tool path has become unreliable and to recover through alternative planning.


\paragraph{Agents struggle to re-plan through longer recovery paths.}
To examine whether block performance depends on the structure of the remaining solution space, we compare two path-restricted settings: one where only the shortest valid solution path remains available, and one where only the longest valid solution path remains available. 
As shown in Figure~\ref{fig:fine_grained_blocking}, the gap is substantial: when only the longest path is kept, accuracy drops sharply across models, indicating that agents struggle to effectively adapt and re-plan when recovery requires following longer and less direct tool-use trajectories. 
This effect is especially pronounced for stronger models: for example, \textit{GPT-5.4} falls to only slightly above 10\% accuracy, compared with around 30\% under the standard block setting. 
This suggests that current agents remain limited in deep adaptive planning: once direct recovery routes are removed, they struggle to reconstruct longer tool-use chains through more distant intermediate steps.

\begin{figure*}[t]
    \centering
    \includegraphics[width=1\linewidth]{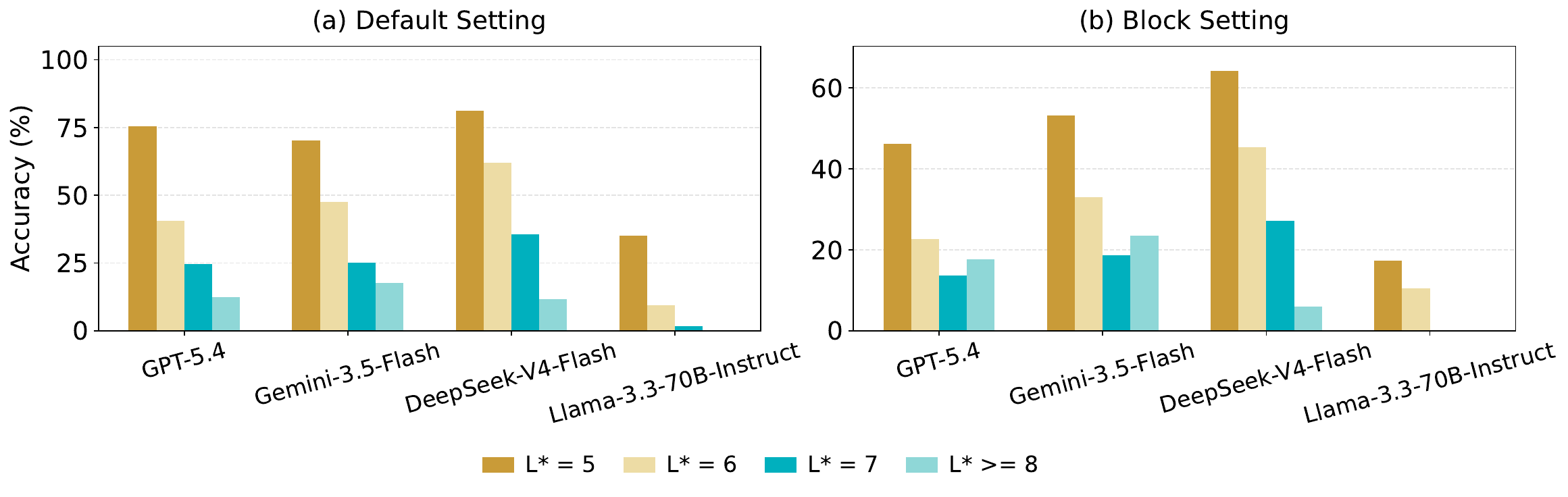}
    \caption{Accuracy (\%) by shortest path length $L^*$ in the default and block settings. Queries are grouped by $L^*$, with longer tasks aggregated into the $L^*\ge 8$ group. Accuracy generally decreases for longer-path groups, and the decline becomes sharper under blocking.}
    \label{fig:task_step}
\end{figure*}

\begin{takeaway}[label={tak:analysis-test-time-tricks}]
Simply scaling test-time compute yields only limited performance gains.
\label{sec:enforce-analysis}
\end{takeaway}

\noindent To examine whether additional test-time interaction can mitigate the performance drop under blocking, we run an enforced-exploration experiment.
Specifically, whenever an agent attempts to terminate with an incorrect answer, we insert an additional user message instructing the model to continue exploring rather than stop.
This intervention can be applied multiple times within the same trajectory, up to a predefined enforced budget, with the detailed prompt shown in Figure~\ref{fig:enforced-prompt}.
As shown in Figure~\ref{fig:enforced_budget_preview}, increasing the enforced budget provides only limited gains: most models improve by less than 5 percentage points.
Moreover, even with additional interaction opportunities, block-setting performance remains far below the corresponding no-block accuracy.
This persistent gap suggests that blocking exposes deeper limitations in adaptive re-planning under unexpected events, rather than a failure that can be resolved by extra test-time interaction alone.

\begin{takeaway}[label={tak:analysis-path-length}]
Longer shortest-path groups are associated with lower accuracy.
\end{takeaway}


\noindent We group queries by the shortest valid solution length $L^*$ and report accuracy within each group, which provides a useful view of how performance changes across tasks with different minimal tool-use horizons. 
As shown in Figure~\ref{fig:task_step}, accuracy generally decreases as $L^*$ increases.
This pattern suggests that longer minimal tool-use horizons are associated with greater difficulty, even before any blocking perturbation is applied.

\section{Error Analysis}
\label{sec:error-analysis}

\paragraph{Overview.}
The main results show that agents often fail even after substantial retrieval and tool interaction, suggesting that the key bottleneck lies inside the trajectory rather than only in final-answer generation.
We therefore analyze failures in three stages:
\textbf{(1) Section~\ref{sec:error-planning-failures}} first studies how planning trajectories break in the default setting: agents often make partial progress, drift away from valid solution paths, and then fail to recover.
We further show that this drift is frequently not caused by the absence of useful retrieved tools, but by the model's failure to select or return to tools that would move the trajectory forward.
\textbf{(2) Section~\ref{sec:error-blocking}} then uses retrieval-time blocking as a controlled stress test of this selection and recovery weakness.
When a required tool is replaced by a corrupted alternative, the agent must detect that the current branch is unreliable and re-plan from remaining feasible paths; failures under blocking therefore reveal whether models can reject misleading or executable-looking bad evidence.
\textbf{(3) Section~\ref{sec:error-modelwise}} compares model-specific termination behaviors after navigation has already failed, showing how different model families expose the same underlying trajectory breakdown through distinct ending patterns.

\subsection{Trajectory Drift and Tool-Selection Failures}
\label{sec:error-planning-failures}

To investigate planning failures in \bench{}, we annotate valid tool calls according to whether they advance a feasible solution path.
We define a tool call as a \textbf{\textit{progress call}} if it produces a new intermediate value needed by at least one valid tool-use path toward the final answer.
Otherwise, it is a \textbf{\textit{non-progress call}}.
This definition does not require the model to follow one fixed ground-truth path; it rewards any step that obtains useful intermediate evidence for completing the task.
Using this definition, we categorize failed trajectories as follows:
\begin{itemize}[topsep=2pt, partopsep=3pt, leftmargin=*, itemsep=-3pt]
\item \textbf{\textit{No Traction}}: the failed trajectory never makes a progress call, so the model obtains no useful intermediate evidence toward the final answer.
\item \textbf{\textit{Irrecoverable Drift}}: the model makes at least one progress call, then makes a non-progress call, and never makes progress again.
\item \textbf{\textit{Weak Recovery}}: the model makes progress again after a non-progress call, but still fails before reaching the correct answer.
\item \textbf{\textit{Format Error}}: the failure is caused by invalid or incompatible formatting (e.g., tool call or retrieval format) in the interaction process.
\end{itemize}
The first three categories describe where a failed trajectory breaks during solution-path execution, while \textit{Format Error} captures interface-level failures that we report separately from the main navigation analysis.

\begin{figure}[t]
\centering
\includegraphics[width=1\linewidth]{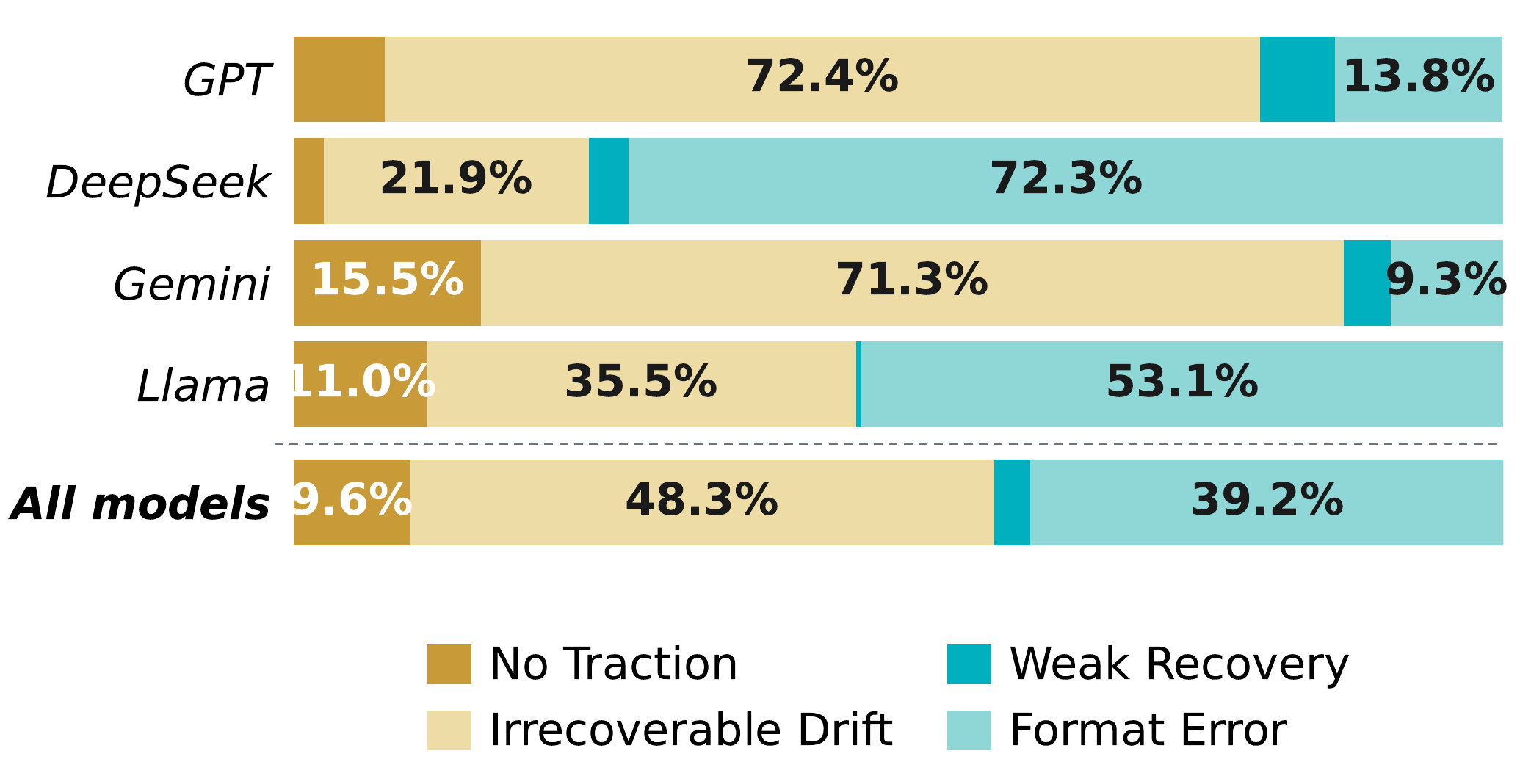}
\vspace{-0.2in}
\caption{
Trajectory Failure Category Distribution under the \textit{default} setting (\%). A tool call is counted as a progress call if it produces a new intermediate value required by at least one ground-truth tool-use path toward the final answer. We abbreviate \textit{GPT-5.4}, \textit{Gemini-3.5-Flash}, \textit{DeepSeek-V4-Flash}, and \textit{Llama-3.3-70B-Instruct} as \textit{GPT}, \textit{Gemini}, \textit{DeepSeek}, and \textit{Llama}.
}
\label{fig:mechanistic_navigation_breakdown}
\vspace{-0.12in}
\end{figure}

\begin{takeaway}
Most failures occur after models have already obtained partially useful evidence, but then drift away from all ground-truth solution paths and rarely recover.
\end{takeaway}

\paragraph{Failures usually happen after partial progress.}
Figure~\ref{fig:mechanistic_navigation_breakdown} shows that models do not simply fail from the start; instead, they often make partial progress before drifting away from valid solution paths.
In the default setting, \textit{Irrecoverable Drift} is the largest planning failure category for \textit{GPT-5.4} and \textit{Gemini-3.5-Flash}, accounting for $72.4\%$ and $71.3\%$ of their reported failure categories, respectively, and remains the dominant category when aggregated across all four analyzed models.
This indicates that models often begin following a useful solution direction, but later take an off-path tool call that stops further progress.

\paragraph{Recovering after drift is difficult.}
The same figure shows that \textit{Weak Recovery} accounts for only $3.0\%$ of the reported default failure categories when aggregated across models.
This low recovery rate suggests that the main difficulty is not simply failing to find a useful tool-use direction.
Many failed trajectories make partial progress at first, but then a single non-progress step can push the agent away from the productive solution direction.
After such drift occurs, agents rarely repair the trajectory sufficiently to complete the task, suggesting that current agents still lack a stable adaptive re-planning mechanism for detecting and self-correcting unproductive directions.

\begin{takeaway}
Drift is often a tool-selection failure, not simply a retrieval failure.
\end{takeaway}

\paragraph{Useful alternatives are often already in the retrieved history.}
A natural explanation for such drift is that the agent may fail to retrieve any tool that can advance the current solution path.
Before failed-run \textit{non-progress calls}, the model had already retrieved at least one valid tool that could support progress toward the solution in $78.0\%$ of default cases and $71.1\%$ of block cases.
Thus, many wrong calls occur even though the model has already seen a solution-relevant alternative.
The bottleneck is therefore not only tool discovery, but also deciding which discovered tool should be used next.

\paragraph{Models over-select recently retrieved tools.}
The failed \textit{non-progress calls} also show a strong recency pattern in tool selection.
In both settings, most \textit{non-progress calls} use tools from recent retrieval windows, accounting for $74.1\%$ in the default setting and $63.6\%$ in the block setting.
However, when a clean tool that could make progress is available, it is often not recent: $44.7\%$ of default cases and $43.2\%$ of block cases involve such tools that were retrieved more than two retrieval windows earlier.
This contrast suggests that models tend to act on recently retrieved tools even when older tools would provide more useful progress toward the final answer.

\paragraph{Even when tools that could advance the task re-appear, agents still do not reliably recover.}
This failure cannot be reduced to memory loss alone.
After drift, a tool whose execution would be counted as a \textit{progress call} re-appears in the recent retrieval context in $42.5\%$ of default cases and $53.4\%$ of block cases.
In these cases, the issue is not that the useful tool only appeared in earlier retrieval history: it becomes available again after the trajectory has already drifted.
Yet the model still often fails to select it.
Recovery therefore requires more than retrieving additional tools or extending the interaction budget.
Agents need to retain useful candidates seen earlier and re-rank both previously seen and newly retrieved tools according to whether executing them would move the trajectory toward the final target.

\paragraph{Selection failure explains why broad exploration is not sufficient.}
This diagnosis also clarifies the gap between exploration and success observed in the main results.
Broad retrieval can expose useful intermediate information, but it does not guarantee that the model will exploit the right tool at the right time.
A model may retrieve many tools, or repeatedly search after drifting, while still failing to choose the progress-capable tool already available in its history.
The core navigation bottleneck is therefore the transition from discovery to useful execution.

\begin{figure}[t]
    \centering
    \includegraphics[width=1\linewidth]{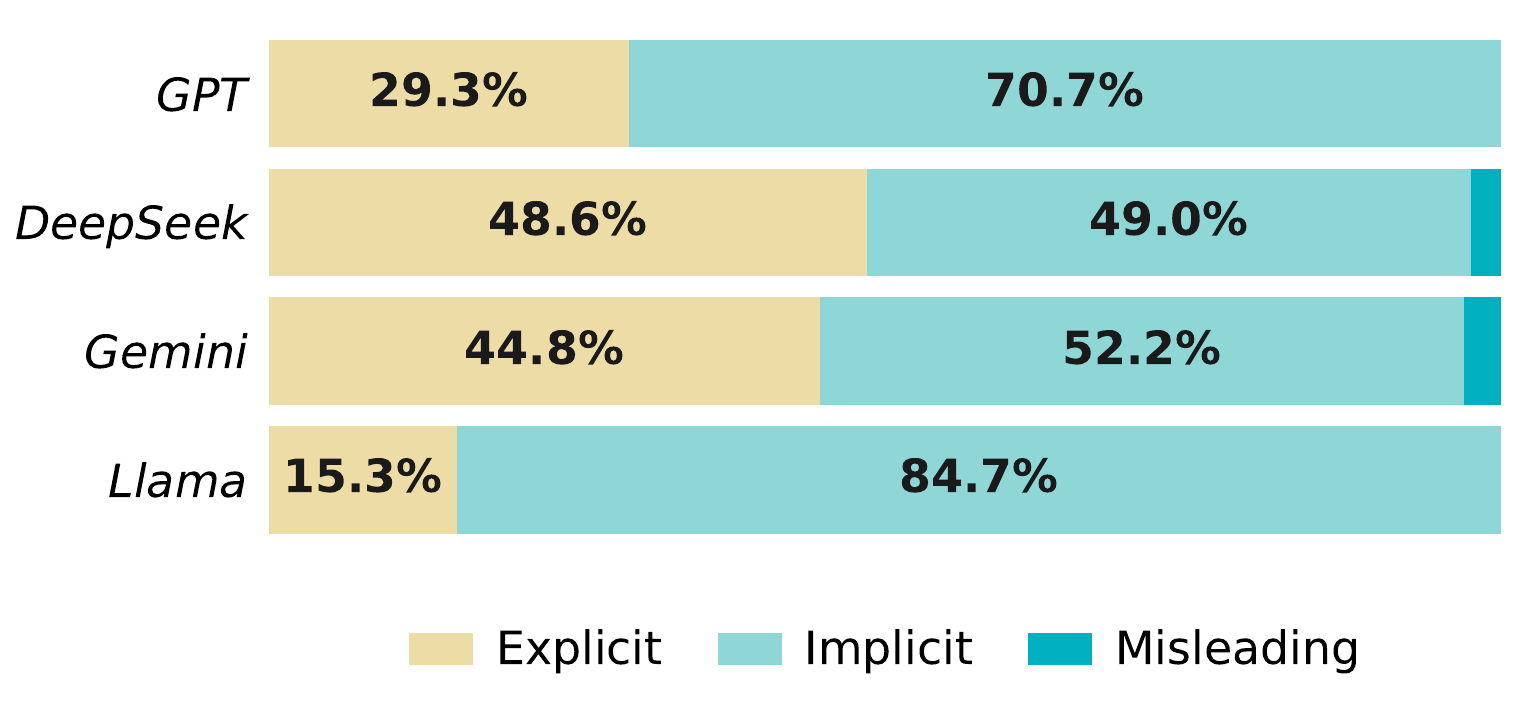}
    \vspace{-0.35in}
    \caption{Distribution of invoked block alternatives under \textit{mixed} blocking. For each model, the horizontal bar normalizes all invoked block alternatives to $100\%$. Colors indicate whether the invoked block alternative is an explicit failure, implicit failure, or semantic misleading tool. We abbreviate \textit{GPT-5.4}, \textit{Gemini-3.5-Flash}, \textit{DeepSeek-V4-Flash}, and \textit{Llama-3.3-70B-Instruct} as \textit{GPT}, \textit{Gemini}, \textit{DeepSeek}, and \textit{Llama}.}
    \label{fig:noisy_tool_type_mix}
\end{figure}

\subsection{Blocked-Alternative Misuse}
\label{sec:error-blocking}


The previous error analysis shows that many failures arise even when useful tools have already been retrieved, because models do not reliably select or recall tools that would make useful progress. Retrieval-time blocking turns this weakness into a controlled stress test. In Takeaway~\ref{tak:analysis-block-type-severity}, we have already shown that block types differ in severity, with silent failures causing especially large performance drops. Here, we move from aggregate performance to fine-grained tool-use behavior: when a required tool is replaced by a block alternative, success depends on whether the agent can detect the disrupted branch, avoid propagating corrupted observations, and re-plan through the remaining feasible paths.

\begin{figure}[t]
    \centering
    \includegraphics[width=1\linewidth]{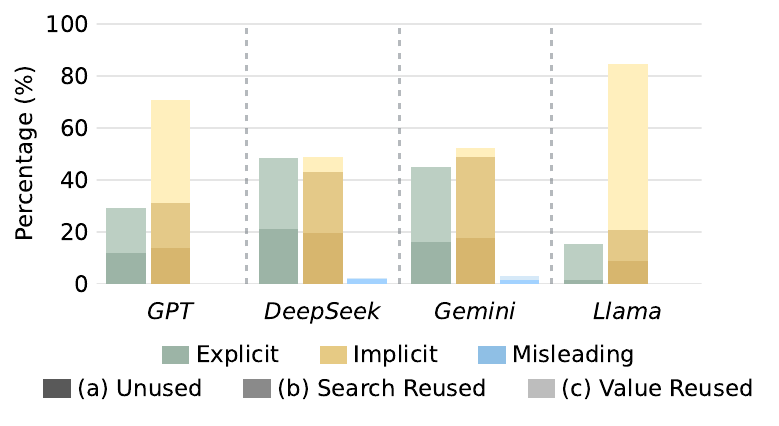}
    \vspace{-0.35in}
    \caption{Follow-up behavior after agents invoke block alternatives under \textit{mixed} blocking. Each colomn corresponds to one model and one block type. The row length shows the ratio of all invoked block alternatives for that model, and colored segments denote \textit{Unused}, \textit{Search Reused}, and \textit{Value Reused}. \textit{GPT} denotes \textit{GPT-5.4}, \textit{Gemini} denotes \textit{Gemini-3.5-Flash}, \textit{DeepSeek} denotes \textit{DeepSeek-V4-Flash}, and \textit{Llama} denotes \textit{Llama-3.3-70B-Instruct}.}
    \label{fig:noisy_tool_followup_behavior}
    \vspace{-0.2in}
\end{figure}

Figures~\ref{fig:noisy_tool_type_mix} and~\ref{fig:noisy_tool_followup_behavior} analyze agents' behavior after they invoke block alternatives returned under retrieval-time blocking.
Here, block alternatives refer to the additional tools inserted in place of blocked executable tools.
Figure~\ref{fig:noisy_tool_type_mix} shows which type of block alternative each model invokes.
Figure~\ref{fig:noisy_tool_followup_behavior} further shows how agents use the output of each invoked block alternative in later steps.
We categorize follow-up behavior into three cases:
\begin{itemize}[topsep=2pt, partopsep=3pt, leftmargin=*, itemsep=-3pt]
\item \textbf{\textit{Unused}}: the agent uses neither the tool's output information for later retrieval nor its returned value in a later tool call.
\item \textbf{\textit{Search Reused}}: the agent uses the tool's output information to retrieve further tools, but does not use its concrete returned value as an argument in a later tool call.
\item \textbf{\textit{Value Reused}}: the agent uses the tool's concrete returned value as an argument in a later tool call.
\end{itemize}
This distinction separates two ways in which a block alternative can affect later planning.
A model may continue from the block alternative's declared output information without trusting the returned value, or it may directly propagate the returned value into later tool calls.

\begin{takeaway}
Models can reject superficial distractors but struggle with executable-looking failures, especially silent ones with superficially plausible outputs.
\end{takeaway}

\paragraph{Semantically misleading tools are usually not the main issue.}
Across models, semantic misleading tools account for at most $3\%$ of invoked block alternatives, and are never selected by \textit{GPT-5.4} or \textit{Llama-3.3-70B-Instruct}.
This suggests that current models can often use tool names and descriptions to reject alternatives whose functions are only superficially related to the current need.
The more difficult cases are block alternatives that look compatible with the current schema or return concrete values that appear usable.

\paragraph{Explicit failures stop direct value propagation but not follow-up search from the failed tool.}
After invoking explicit-failure block alternatives, agents never use the returned error message as an argument in later tool calls.
This shows that explicit error signals are effective at preventing direct propagation of obviously invalid values.
However, explicit failures do not always stop the agent from searching for follow-up tools from the failed call.
Agents may still use the failed tool's declared output information to retrieve downstream tools, even though the concrete execution did not produce a valid value.
Thus, explicit errors make the failed branch visible, but they also expose a remaining challenge: the model must still abandon that branch and re-plan through another feasible path.

\paragraph{Implicit failures turn drift into value contamination.}
Implicit failures are more damaging because they return superficially plausible concrete values instead of explicit error messages.
For \textit{GPT-5.4} and \textit{Llama-3.3-70B-Instruct}, \textit{Value Reused} accounts for $55.9\%$ and $75.5\%$ of implicit-failure follow-ups, respectively; aggregated across models, implicit failures lead to \textit{Value Reused} in $42.2\%$ of cases, compared with $0\%$ for explicit failures.
This means that silent failures do not merely stop progress.
They inject invalid values that the agent treats as grounded evidence for later tool calls.
Once these values enter the trajectory, subsequent calls may look procedurally valid while moving the agent further away from any valid solution.


\subsection{Model-Specific Failure Patterns}
\label{sec:error-modelwise}

The previous findings describe where the trajectory breaks and why recovery is difficult.
We next ask how models behave after their trajectory is no longer grounded in a valid solution path.
We treat these final behaviors as termination policies rather than primary failure mechanisms, because the same trajectory-level failure can end in different surface outputs.
We define three dominant ending types:
\begin{itemize}[topsep=2pt, partopsep=3pt, leftmargin=*, itemsep=-3pt]
\item \textbf{\textit{Surrender}}: the final response explicitly states that the model cannot determine, verify, or provide the requested answer.
\item \textbf{\textit{Wrong tool value}}: the final response gives an incorrect answer that is not supported by any tool output relevant to the target query. The value may be fabricated or guessed, or it may be copied from another tool call whose output does not answer the requested query.
\item \textbf{\textit{Search exhaustion}}: the trajectory ends without a grounded answer after repeated retrieval/search actions fail to produce a decisive progress-making tool call.
\end{itemize}

\begin{takeaway}
After navigation failure, models expose different termination policies: GPT surrenders, DeepSeek and Llama commit wrong tool values, and Gemini keeps searching.
\end{takeaway}

\paragraph{GPT often surrenders despite the existence of valid solution paths.}
Among default non-interaction failures, \textit{GPT-5.4} ends $77.3\%$ with an explicit surrender statement, despite the prompt explicitly stating that each task is solvable, as shown in Figure~\ref{fig:inference-prompt}. Under the block setting, this rate further rises to $80.6\%$. These terminations therefore do not indicate genuinely unsolvable tasks: each blocked instance preserves at least one feasible solution path by construction. Rather, they suggest that \textit{GPT-5.4} often stops once its current branch no longer appears useful, instead of abandoning that branch and recovering through another valid path. Although this behavior is conservative in that the model usually avoids hallucinating an answer, it still reflects a failure of adaptive recovery in a solvable environment.

\paragraph{DeepSeek and Llama more often commit hallucinated values.}
\textit{DeepSeek-V4-Flash} commits wrong tool-returned values in $58.8\%$ of default failures and $65.9\%$ of block failures.
\textit{Llama-3.3-70B-Instruct} shows this pattern even more strongly, committing wrong tool values in $81.7\%$ of default failures and $71.7\%$ of block failures.
Compared with explicit refusal or surrender, this represents a more severe failure mode: the model still produces a final answer even when the trajectory has drifted away from any valid solution path. 
In such cases, it may rely on unsupported evidence, fabricate a value, or reuse an intermediate result from an irrelevant tool call as if it answered the target query.

\paragraph{Gemini keeps searching without converting search into progress.}
\textit{Gemini-3.5-Flash} ends $90.8\%$ of default failures and $85.1\%$ of block failures through search exhaustion.
Its S/C ratio averaged on the failed instances is also much larger than the other analyzed models, reaching $29.1$ in the default setting and $18.3$ under blocking.
This suggests that \textit{Gemini-3.5-Flash} mainly fails not by premature commitment, but by searching without making decisive progress.

\paragraph{These model-specific failure fingerprints are stable across settings.}
The trajectory-level analysis shows that these model-specific patterns are largely stable across settings.
Across the tested blocking settings, the dominant failure pattern remains unchanged for \textit{DeepSeek-V4-Flash}, \textit{Gemini-3.5-Flash}, and \textit{Llama-3.3-70B-Instruct}, and also holds for \textit{GPT-5.4} in most settings.
This indicates that blocking, path length, feedback, and other interventions mostly change how often each model fails, rather than replacing its characteristic failure policy.
The resulting picture is therefore hierarchical: models first fail through weak solution-path execution and recovery, corrupted tools then amplify this weakness through uncontained failure branches, and each model finally exposes the broken trajectory through its own termination bias.

\section{Conclusion}






In this work, we introduced \BENCH{}, an interactive benchmark for evaluating long-horizon adaptive planning in large-scale, retrieval-mediated tool ecosystems.
Our results show that current LLM agents remain brittle in massive-tool environments: even frontier models drop sharply when relevant tools are blocked, silently corrupted, or require longer recovery paths, while longer test-time interaction brings only limited gains.
These failures suggest that robust planning requires agents to recognize unreliable feedback, preserve intermediate evidence, and re-plan under partial and imperfect tool observability.
Looking forward, \bench{} provides a testbed for developing adaptive agents that explore large tool ecosystems, recover from unreliable tools, and operate under real-world uncertainty, with future directions in Appendix~\ref{app:potential-improvement-methods}.

\section*{Limitations}

While \bench{} provides a principled and scalable framework for evaluating adaptive planning under massive, retrieval-mediated tool access, it has several limitations.
First, \bench{} is currently instantiated in the retail domain. Although its queries cover diverse multi-step workflows, a single domain may not fully capture the breadth of real-world tool-use scenarios. Since \bench{} is built with a scalable generation pipeline over typed datatypes, executable tools, and backend databases, it can be extended to additional domains in future work.
Second, our retrieval-time blockers simulate representative tool-access failures, including explicit errors, counterfactual outputs, and irrelevant tools. However, real-world tool ecosystems may involve more complex and dynamic failures. This reflects a realism--controllability trade-off: our design enables systematic robustness analysis, while abstracting away some open-ended complexity.
Third, \bench{} uses a self-defined retriever to expose tools through controlled natural-language retrieval. While this enables reproducible evaluation of exploration and re-planning, it may not fully capture real-world retrieval systems, where tool discovery can be affected by noisy documentation, imperfect ranking, changing tool catalogs, and deployment-specific retrievers.
We discuss these trade-offs further in Appendix~\ref{app:justifications}.

\section*{Ethics Statements}


\paragraph{Offensive Content.}
Our benchmark focuses on the retail domain, where offensive content is relatively unlikely to arise. 
In addition, all data were carefully sampled and validated to ensure that the dataset does not contain offensive material. 
Therefore, we believe the benchmark presents minimal risk of negative societal impact.

\paragraph{Licenses.} 
Our code will be released under the MIT license to allow unrestricted research use. 
The \bench{} will be distributed under a Creative Commons (CC) license, providing free access for the academic community. 
Our use of existing models and tools is strictly consistent with their original licenses and intended research purposes. 
We take full responsibility for any potential rights violations or licensing issues, and all resources comply with their respective terms of use while supporting research purposes.

\paragraph{Model Usage.}
All open-source models were hosted and executed locally using the vLLM library~\cite{vllm}, while all closed-source models were accessed through their respective APIs. 

\paragraph{Data Annotations.}

All data annotation was performed by the paper's co-authors, who are qualified researchers with relevant expertise, ensuring that the process was conducted responsibly and in accordance with ethical standards.


\clearpage

\appendix

\begin{table*}[t]
\centering
\small
\setlength{\tabcolsep}{3.5pt}
\renewcommand{\arraystretch}{1.18}

\begin{tabularx}{\textwidth}{@{}
>{\raggedright\arraybackslash}p{0.14\textwidth}
>{\centering\arraybackslash}p{0.10\textwidth}
>{\centering\arraybackslash}p{0.10\textwidth}
>{\centering\arraybackslash}p{0.13\textwidth}
>{\raggedright\arraybackslash}X
@{}}
\toprule
\textbf{Blocking Type} &
\textbf{Detection Stage} &
\textbf{Signal Explicitness} &
\textbf{Alignment} &
\textbf{Agent-Side Interpretation} \\
\midrule

\textbf{Semantic Misleading} &
After retrieval &
Medium &
Aligned &
The retrieved tool is superficially similar to the blocked tool, but its description or schema reveals that it supports a different function. If invoked, the response is consistent with the tool's own description. \\

\textbf{Explicit Failure} &
After tool call &
High &
Misaligned &
The retrieved tool appears usable before invocation, but calling it returns an explicit error or failure message, directly signaling that the tool cannot support the intended path. \\

\textbf{Implicit Failure} &
After tool call &
Low &
Misaligned &
The retrieved tool appears usable and returns a value without an explicit error, but the response violates the tool's described behavior or task-world consistency, making the blockage harder to detect. \\

\bottomrule
\end{tabularx}

\caption{\textbf{Taxonomy of blocking types}. We characterize each blocker by the earliest stage at which an agent can detect it (\textit{Detection Stage}), the explicitness of the observable failure signal (\textit{Signal Explicitness}), whether the retrieved tool description aligns with the tool's actual behavior (\textit{Alignment}), and how the agent should interpret the blocking (\textit{Agent-Side Interpretation}).
}
\label{tab:blocking-comparison-tab}
\end{table*}

\section{Comparison Traits}
\label{app:comparison-traits}


In this appendix, we describe the comparison traits used in Table~\ref{tab:comparison-table}. 
Rather than treating these traits as isolated checklist items, we use them to characterize increasingly realistic conditions for tool-using agents~\citep{yao2024taubench}: operating with external tools~\citep{ToolOrchestra,xRouter}, discovering relevant tools from large tool spaces~\citep{APIBench,ToolBench, ToolRet}, exploring from both known evidence and desired outcomes~\citep{NAACL,EscapeBench}, remaining robust when retrieved tools are imperfect~\citep{ToolGym} and effectively re-plan~\citep{shen2023taskbench, CostBench,AdaPlanBench}.

\paragraph{Tool-Use.}
Tool-use is a basic requirement for evaluating agents that interact with external environments~\citep{ToolRL}.
In real applications, agents must often obtain information or perform operations through APIs~\citep{APIBench}, databases~\citep{LiveAPIBench}, or other executable interfaces~\citep{SWEAgent} rather than relying only on parametric knowledge.
We therefore distinguish benchmarks where tool invocation is central to task completion from those where tools are only optional.

\paragraph{Tool Retrieval.}
Real-world tool ecosystems can contain hundreds or thousands of APIs~\citep{ToolLLM, LiveAPIBench, liu-etal-2024-gprooft}, making it impractical to expose the full tool set in the prompt. 
Agents must instead retrieve or discover relevant tools during problem solving. 
This trait captures whether a benchmark evaluates tool use under such retrieval-mediated access, rather than assuming that the correct tools are already visible to the agent.

\paragraph{Implicit Sub-goals.}
Complex tasks rarely specify all intermediate steps or sub-goals explicitly~\citep{EscapeBench}. 
In realistic workflows, an agent may need to infer what intermediate information is missing, what sub-goal should be solved next, and which tools can bridge the gap between the current state and the final objective~\citep{PlanAndSolve}. 
This trait captures whether a benchmark requires agents to discover and pursue such latent intermediate goals, instead of following fully specified instructions.

\paragraph{Bi-directional Exploration.}
In real-world environments, effective exploration may need to proceed in two directions: forward from currently available evidence, and backward from the desired target outcome~\citep{BackwardPlanningLLM}. 
For example, an agent may ask what can be obtained from a known order ID, or instead ask what tools could produce a required refund status. 
This trait captures whether a benchmark encourages agents to combine forward planning from task goals with backward planning from available tool affordances~\citep{CreativityBench}, rather than limiting them to fixed tool lists, one-shot retrieval, or purely forward expansion from explicit instructions.

\paragraph{Unreliable Tools.}
Retrieved tools in real systems may be unavailable, stale, misleading, broken, or only partially relevant~\citep{CostBench, UserBench}. 
Robust agents therefore need to evaluate tool feedback, recognize failure modes, and recover from unreliable tool access. 
This trait captures whether a benchmark exposes agents to imperfect tools as part of the task environment, rather than assuming clean and fully reliable tool descriptions and executions.

\paragraph{Long-Horizon.}
Many practical tool-use tasks require multiple dependent steps, where the result of one tool call determines what should be retrieved or executed next~\citep{ReAct}.
Long-horizon settings test whether agents can maintain state, plan across several interactions, and avoid compounding errors~\citep{AgentBench, WebArena, NESTful}.
Following the criterion used in~\citet{Tool-Decathlon}, we regard tasks involving around 25 turns or more as long-horizon.
In \bench{}, each task is constructed to require at least five dependent tool invocations, ensuring that solving it cannot be reduced to single-step API selection.
Empirically, when retrieval and execution are both counted as interactions, agents require around 25 turns on average to complete a task, matching this long-horizon convention.
This trait captures whether a benchmark systematically evaluates such extended multi-step tool-use trajectories, rather than mostly single-step API selection or isolated function calling.


\paragraph{Scalable Generation.}
Realistic agent evaluation requires broad coverage over tasks, tools, and environments~\citep{RandomWorld, ToolLLM, LiveAPIBench,guo2025mathematical}. 
Manual benchmark construction alone often limits diversity and makes it difficult to stress-test agents under many configurations. 
This trait captures whether a benchmark includes a scalable construction pipeline for generating tasks, tools, environments, or perturbations, enabling broader and more systematic evaluation.

\section{Experiment Details}
\label{app:experiment-details}

\subsection{Construction Model Choice}
\label{app:model-choice}

We use $M_{\mathrm{gen}}=GPT-5.2$ and $M_{\mathrm{fil}}=GPT-5.2$ in the data construction process.

\subsection{Tool Construction Details}
\label{app:tool-construction-details}

\subsubsection{Tool Schema Construction Details}
\label{app:schema-construction-details}

Following the notation in the main text, each tool $\tau \in \mathcal{T}$ has an input datatype set $\mathcal{D}*{\mathrm{in}}(\tau)$ and an output datatype set $\mathcal{D}*{\mathrm{out}}(\tau)$, with $|\mathcal{D}*{\mathrm{in}}(\tau)|=m$ and $|\mathcal{D}*{\mathrm{out}}(\tau)|=n$.
In the released retail benchmark, we instantiate this construction with $m\in{1,2,3,4,5}$ and $n=1$.
Although the combinatorial candidate space grows rapidly with larger $m$, LLM-based filtering keeps the final released tool library sparse.
We use $m\leq 5$ because this range better matches common real-world tool schemas.
Many real-world tools need more than one input, such as an user name together with a timestamp, product item, or order status constraint.
Tools with more than five required inputs are relatively uncommon in the retail setting.
We therefore include multi-input tools to better match practical tool-use scenarios, but avoid creating tools with unrealistically large input signatures.
Benchmark difficulty instead comes from composing tools over long paths, retrieving useful tools, and re-planning when earlier choices fail.

\paragraph{Tool Name Construction.}
Tool names in our benchmark are generated automatically by code according to a fixed naming rule.
To better match real-world tool ecosystems, where the same information can be named differently across tools, we maintain $5$--$10$ aliases for each datatype.
For example, the same datatype may appear as ``user name'', ``customer name'', or ``name on the account'' in different tool names.
This alias design also help prevent agents from recovering the underlying tool graph by simply matching canonical datatype names across tool names.
When constructing a tool name, we select one alias for each input and output datatype.
Names follow the template \texttt{Get\_<Output>\_From\_<Input>[\_<Variant>]}, where \texttt{<Output>} is constructed from the selected alias of the output datatype, \texttt{<Input>} is constructed from the selected aliases of the required input datatypes, and the optional \texttt{<Variant>} suffix, such as ``\texttt{V2}'' or ``\texttt{Pro}'', provides additional version information when needed. 
The suffix does not always indicate a functional difference.
It is used to simulate the existence of multiple tool versions in real-world tool ecosystems.
This naming convention keeps the tool inventory human-readable and machine-parsable.

\paragraph{Tool Description Construction.}
Tool descriptions in our benchmark are generated using $M_{\mathrm{gen}}$ to produce natural-language explanations of tool functionality. 
Each description explains what information the tool retrieves from the given input and clarifies the contextual meaning of the returned output datatypes in high-level natural language. 
The descriptions provide natural-language context for the tool names and help distinguish tools whose input-output signatures are similar.

\subsubsection{Noisy Tool Construction}
\label{app:noisy_tool}

Real-world retrieval over large tool ecosystems is rarely fully clean: retrieved candidates may be semantically related to the current intent, but still differ in reliability, freshness, authority, or executability.
To better reflect this setting, we augment the tool ecosystem with noisy tools that appear relevant to the same retrieval intent as valid tools, but whose outputs are unreliable or otherwise unhelpful for solving the task.
To achieve this, We generate five noisy variants for each executable tool, resulting in $925$ noisy tools for $185$ executable tools.
Each noisy tool is paired with one executable tool and keeps the same required inputs and declared output datatype.
Its name follows the same alias-based naming rule, usually with a small suffix or version change.
Its description is generated by $M_{\mathrm{gen}}$ and stays close to the paired tool in task intent, but it also explicitly describes why the tool is unavailable or unreliable.
Thus, noisy tools are plausible retrieval candidates, but they are not meant to be indistinguishable from the corresponding executable tools.

We construct five noise categories:
\begin{itemize}
\item \textbf{Deprecated.} The tool looks like a normal endpoint, but returns an unsupported-endpoint error when invoked.
\item \textbf{Condition-limited.} The tool has a valid-looking signature, but is only available under special record conditions that are not satisfied in the benchmark instance.
\item \textbf{Stale.} The tool returns the right type of field, but the value comes from an outdated cache or lagging replica.
\item \textbf{Unreliable.} The tool declares the intended output field, but returns a value from a related but wrong backend field.
\item \textbf{Non-authoritative value.} The tool returns a preview, default, estimated, or otherwise non-final value instead of the final backend truth.
\end{itemize}

For each executable tool, we use $M_{\mathrm{gen}}$ to synthesize one noisy counterpart for each category.
The execution behavior is changed according to the target noise category.
This construction tests whether agents can use tool descriptions to reject superficially relevant but explicitly unreliable tools.

\subsubsection{Tool Filtering Details}
\label{app:detailed-filtering-criteria}


We use $M_{\mathrm{fil}}$=\textit{GPT-5.2} to screen all candidate tools, and only retain a candidate tool if it meets all of the following criteria:
\begin{itemize}
    \item \textbf{Deterministic dependency.} The declared input datatype set should be sufficient for determining the output datatype set. We exclude tools with redundant inputs when a strict subset already determines the same output datatype set. For example, if \texttt{Get\_Phone\_From\_User\_ID} already maps a user ID to the corresponding phone number, then \texttt{Get\_Phone\_From\_User\_ID\_and\_Email} should be filtered out, because the email input is redundant for determining the same output datatype.
    \item \textbf{Tool-grounded information access.} A tool must correspond to retrieving or deriving information from the external database, rather than encoding a pure reasoning shortcut that the model could perform without interacting with the environment. For example, \texttt{Get\_TotalPrice\_From\_Qty\_and\_UnitPrice} should be filtered out, because the total price can be directly computed as quantity multiplied by unit price without~calling the~tool.
    \item \textbf{Domain realism.} The input-output relation must be plausible under the retail domain semantics encoded in the database and datatype inventory. For example, \texttt{Get\_Order\_ID\_From\_Product\_Size} should be filtered out, because a product size alone would not normally be enough to determine a specific order in a retail setting.
    \item \textbf{Non-trivial information gain.} The output datatype set must contribute genuinely new state information, rather than merely renaming, formatting, or echoing already available values. For example, \texttt{Get\_Formatted\_Phone\_From\_Phone} should be filtered out, because converting a phone number such as \texttt{135****6821} into a formatted variant such as \texttt{(+1)135****6821} does not introduce new task-relevant information.
\end{itemize}

\subsubsection{Why Tool Calls Are Necessary}
\label{app:tools-necessity-details}


Tool-use is necessary in \bench{} because every tool call reveals backend values that are inaccessible before execution.
At the single-tool level, tools are constructed either as direct search operations or as action-like operations exposed through a search-style interface. 
A \textit{\textbf{direct search tool}} uses valid input identifiers to look up requested fields in the hidden backend database. Since these values are not stated in the user query and cannot be inferred, the agent must invoke the tool to obtain them.
The same principle applies to \textit{\textbf{action-like tools}}. 
Such a tool represents an operation and returns the values produced by that operation, such as an \texttt{operation ID}, \texttt{confirmation token}, or \texttt{created-record identifier}.
These values are only available after the tool is invoked and may serve as a required typed input for later tools. 
Because the concrete information needed to proceed is only exposed through tool execution, the agent cannot bypass any single tool call by reasoning about what should happen.

\subsection{Query Construction Details}
\label{app:query-construction-details}


We first enumerate all solvable tasks from the path catalogs. 
A task is solvable if the target datatype set can be reached from the declared initial datatype set through at least one valid tool sequence. 
We then filter these solvable tasks by solution length and input usage, and construct the tasks into queries.

The filtering rules are:
\begin{itemize}
    \item \textbf{Removing short tasks.} We remove tasks whose shortest valid solution path is shorter than 5 steps, since these tasks can be solved with only a short lookup chain.
    \item \textbf{Keeping useful multi-input tasks.} We keep both one-input and multi-input tasks. For a multi-input task, the target datatype set should not be reachable from only part of the declared input datatype set. This avoids cases where a task is labeled as multi-input but one of the inputs is never actually needed.
\end{itemize}



\subsection{State Graph}
\label{app:state-graph}

For a query instance indexed by $i$, let $r_i=(\mathcal{D}_{i,0}, \mathcal{Y}_i)$ denote its internal task specification, where $\mathcal{D}_{i,0}\subseteq\mathcal{D}$ is the initial datatype set and $\mathcal{Y}_i\subseteq\mathcal{D}$ is the target datatype set.
We define the state graph for query $i$ as $\mathcal{G}_i=(\mathcal{V}_i,\mathcal{E}_i)$.
Each node $s \in \mathcal{V}_i$ is an internal agent state, represented by the set of datatypes that have been obtained so far:
\begin{equation}
s \subseteq \mathcal{D}.
\end{equation}
The initial state is
\begin{equation}
s_i^0 = \mathcal{D}_{i,0}.
\end{equation}
Each directed edge corresponds to one valid tool invocation. 
For a state $s$ and a tool $\tau$, the edge induced by $\tau$ is defined as
\begin{equation}
\begin{aligned}
(s,\tau,s') \in \mathcal{E}_i
\quad \Longleftrightarrow \quad
& \mathcal{D}_{\mathrm{in}}(\tau) \subseteq s, \\
& \mathcal{D}_{\mathrm{out}}(\tau) \setminus s \neq \emptyset, \\
& s' = s \cup \mathcal{D}_{\mathrm{out}}(\tau).
\end{aligned}
\end{equation}
Thus, a tool can be invoked only when all of its input datatypes are already available, and executing it expands the current state by adding its output datatype set.
Under this representation, a tool-use trajectory is a path in $\mathcal{G}_i$:
\begin{equation}
\pi = (\tau_1,\ldots,\tau_K),
\end{equation}
which induces a sequence of states
\begin{equation}
s_i^0 \xrightarrow{\tau_1} s_i^1 \xrightarrow{\tau_2} \cdots \xrightarrow{\tau_K} s_i^K.
\end{equation}
The task $r_i$ is solvable when there exists a path $\pi$ such that
\begin{equation}
\mathcal{Y}_i \subseteq s_i^K.
\end{equation}
We use this state-graph representation for reachability checks, path enumeration, and blocking analysis.

\subsection{Ground-truth Details}
\label{app:groundtruth-details}

\subsubsection{Process-level Tool Call Ground-truth}
\label{app:process-level-groundtruth}

We construct process-level ground truth by enumerating valid paths in $\mathcal{G}_i$. 
For each query instance $i$, we first compute the inclusion-minimal tool sets that can reach 
$\mathcal{Y}_i$ from $\mathcal{D}_{i,0}$.
The computation is performed by a backward search from the target datatype set. 
Starting with the unresolved goal set $G=\mathcal{Y}_i$, the search repeatedly selects an unresolved datatype $g \in G$ and enumerates tools whose output datatype set contains $g$.
When a tool $\tau$ is selected, the search adds $\tau$ to the current tool set, removes the goals covered by its output datatype set, and adds its required input datatypes as new goals:
\begin{equation}
G' = 
(G \setminus \mathcal{D}_{\mathrm{out}}(\tau)) 
\cup 
(\mathcal{D}_{\mathrm{in}}(\tau) \setminus \mathcal{D}_{i,0}).
\end{equation}
The recursion terminates when all unresolved goals are already covered by the query-provided datatypes, i.e.,
\begin{equation}
G \setminus \mathcal{D}_{i,0} = \emptyset.
\end{equation}

During this backward search, we maintain an antichain of tool sets under set inclusion. 
A candidate tool set is discarded if it strictly contains an existing solution set, and any existing supersets are removed when a smaller solution set is found. 
The resulting collection is denoted as
\begin{equation}
\mathcal{M}_i = \{R_{i,1},\ldots,R_{i,J_i}\},
\end{equation}
where each $R_{i,j} \subseteq \mathcal{T}$ is an inclusion-minimal tool set sufficient to reach $\mathcal{Y}_i$ from $\mathcal{D}_{i,0}$.

\begin{algorithm}[t]
\caption{Computing inclusion-minimal tool sets}
\label{alg:minimal-tool-sets}
\begin{algorithmic}[1]
\Require Initial datatype set $\mathcal{D}_{i,0}$, target datatype set $\mathcal{Y}_i$, tools $\mathcal{T}$
\Ensure Inclusion-minimal tool sets $\mathcal{M}_i$
\Function{Solve}{$G$}
    \State $G \gets G \setminus \mathcal{D}_{i,0}$
    \If{$G = \emptyset$}
        \State \Return $\{\emptyset\}$
    \EndIf
    \State Select one datatype $g \in G$
    \State $\mathcal{A} \gets \emptyset$
    \ForAll{$\tau \in \mathcal{T}: g \in \mathcal{D}_{\mathrm{out}}(\tau)$}
        \State $G' \gets (G \setminus \mathcal{D}_{\mathrm{out}}(\tau)) \cup (\mathcal{D}_{\mathrm{in}}(\tau) \setminus \mathcal{D}_{i,0})$
        \ForAll{$R \in \Call{Solve}{G'}$}
            \State $R' \gets R \cup \{\tau\}$
            \State $\mathcal{A} \gets \mathrm{Add}(\mathcal{A}, R')$
        \EndFor
    \EndFor
    \State \Return $\mathcal{A}$
\EndFunction
\State \Return $\Call{Solve}{\mathcal{Y}_i}$
\end{algorithmic}
\end{algorithm}

For each inclusion-minimal tool set $R_{i,j} \in \mathcal{M}_i$, we enumerate all legal execution orders on $\mathcal{G}_i$. 
An ordered sequence $\pi=(\tau_1,\ldots,\tau_K)$ is legal for $R_{i,j}$ if it is a permutation of the tools in $R_{i,j}$ and every tool is executable when it is called. 
Equivalently, if the induced states are
\begin{equation}
s_i^0 \xrightarrow{\tau_1} s_i^1 \xrightarrow{\tau_2} \cdots \xrightarrow{\tau_K} s_i^K,
\end{equation}
then for every step $k$,
\begin{equation}
\mathcal{D}_{\mathrm{in}}(\tau_k) \subseteq s_i^{k-1},
\qquad
s_i^k = s_i^{k-1} \cup \mathcal{D}_{\mathrm{out}}(\tau_k).
\end{equation}
The sequence is retained only if it reaches the target datatype set:
\begin{equation}
\mathcal{Y}_i \subseteq s_i^K.
\end{equation}

Let $\mathrm{Legal}_i(R_{i,j})$ denote the retained legal execution orders of $R_{i,j}$ that reach $\mathcal{Y}_i$ in $\mathcal{G}_i$. 
The process-level ground truth for query $i$ is the union of these sequences:
\begin{equation}
\Pi_i = \bigcup_{R_{i,j}\in\mathcal{M}_i} \mathrm{Legal}_i(R_{i,j}).
\end{equation}
Each $\pi \in \Pi_i$ records the ordered tool names and its number of steps. 
This set serves as the reference for process-level evaluation and forms the path catalog used in later analyses.

\subsubsection{Final Answer}
\label{app:final-answer-groundtruth}



\paragraph{Final Answer Construction.}
We construct the gold final answer by executing one valid ground-truth path for each query. 
For query $i$, we select a path $\pi_i^\star \in \Pi_i$. 
Given the query input values and $\pi_i^\star$, a designated generation model $M_{\mathrm{gen}}$ walks through the sequence and invokes the tools in order. 
The values returned for the target datatype set $\mathcal{Y}_i$ are used to construct the gold final answer $o_i^\star$. 
In our experiments, we set $M_{\mathrm{gen}}$ to \textit{GPT-5.2}.

\paragraph{Answer Evaluation Protocol.}
A model prediction is judged by normalized containment matching~\citep{NAACL}. 
Let $\hat{o}_i$ denote the model's final answer for query $i$. 
We normalize both $o_i^\star$ and $\hat{o}_i$ by lowercasing the strings, removing lightweight markup and quotation characters, and collapsing repeated whitespace. 
The prediction is marked correct when the normalized gold answer appears as a sub-string of the normalized prediction. 
This criterion avoids relying on semantic equivalence while allowing harmless formatting around an otherwise correct value.

\subsection{Retriever Details}
\label{app:retriever-details}




\paragraph{Request Matching Mechanism.} The retriever maps an agent's natural-language request to canonical datatypes and then returns tools that satisfy the requested typed constraints. 
Each datatype contains a canonical name, a description, and a small set of aliases. 
At retrieval time, the query phrase and datatype aliases are encoded with a lightweight hashing encoder based on word tokens, token bigrams, and character trigrams. 
The retriever selects the top-1 datatype by cosine-style sparse-vector similarity. 
This alias-aware matching supports surface-form variation in agent requests, such as ``phone'', ``user phone'', or ``phone number for the user account''.
To reduce the context burden on the agent, we cap each retrieval result at $\Lambda_{\mathrm{ret}}^{\mathrm{cap}}=30$ tools. This cap does not exclude tools required for solving a task. In our benchmark, any single retrieval request matches at most $K_{\max}^{\mathrm{ret}}=14$ executable tools, where executable tools refer to the non-noisy tools that can produce valid intermediate or final outputs.
For each retrieval request, the retriever first identifies all executable tools whose stored typed interfaces match the resolved type-level retrieval query, and returns these tools before adding any distractors. It then fills the remaining slots with noisy variants paired with the matched executable tools, stopping when the returned list reaches $\Lambda_{\mathrm{ret}}^{\mathrm{cap}}=30$ or when no paired noisy variants remain. To avoid concentrating distractors around only a few executable tools, noisy variants are added as evenly as possible across the matched executable tools.

\paragraph{Retriever modes.}
The retriever supports the same three lookup modes described in the main text. 
Below, we provide the implementation details of how each mode maps the agent's request to tool candidates.
\begin{itemize}
    \item \textbf{Input-conditioned retrieval.} The agent specifies an input datatype set. The retriever maps the request to canonical input datatypes and returns tools $\tau$ whose input datatype set $\mathcal{D}_{\mathrm{in}}(\tau)$ matches the requested input constraint. Input datatypes are matched as an unordered set.
    \item \textbf{Output-conditioned retrieval.} The agent specifies an output datatype set. The retriever maps the request to canonical output datatypes and returns tools $\tau$ whose output datatype set $\mathcal{D}_{\mathrm{out}}(\tau)$ matches the requested output constraint. Output datatypes are also matched as an unordered set.
    \item \textbf{Input-output-conditioned retrieval.} The agent specifies both an input datatype set and an output datatype set. The retriever maps all requested datatypes to their canonical forms and returns tools $\tau$ whose typed interface satisfies both constraints.
\end{itemize}

After the datatype mapping step, tool retrieval is performed by exact matching over the typed tool interfaces. 
For each retrieval request, the environment returns between $K_{\min}^{\mathrm{ret}}=1$ and $K_{\max}^{\mathrm{ret}}=14$ candidate tools, depending on how many tools satisfy the resolved typed constraint.
Given a resolved output datatype set, input datatype set, or both, the retriever returns tools whose stored interface matches the requested constraint.
If no single tool satisfies the requested typed constraint, the environment returns a retrieval message indicating that no direct one-step tool is available and that the agent may need to search through intermediate datatypes.

\subsection{Runtime Protocol Details}
\label{app:runtime-interaction-details}

\paragraph{Datatype Checking.}
The runtime enforces datatype constraints during both tool execution and final-answer evaluation. 
Before a tool is executed, all datatypes required by the tool's input signature must already be present in the internal state. 
This prevents the agent from calling a tool before the necessary intermediate values have been obtained. 
When the agent submits its final answer, the evaluation environment also verifies whether the internal state contains all target datatypes in $\mathcal{Y}_i$.
An answer is marked incorrect if the current internal state does not contain $\mathcal{Y}_i$, even when the answer string matches the gold value.
This prevents the agent from bypassing the tool trajectory and guessing or copying a plausible final value without reaching the required typed state.

\paragraph{Tool Availability.}
The agent may only call tools that have been returned by the retriever during the current interaction. 
The callable set is accumulated across retrieval rounds, so a tool retrieved in any previous round remains callable in later rounds. 


\paragraph{Tool Output Resolution.}
For a valid call to tool $\tau$, the environment resolves the values associated with the output datatype set $\mathcal{D}_{\mathrm{out}}(\tau)$.
During backend construction, we ensure that each valid tool call determines a unique output-value tuple for $\mathcal{D}_{\mathrm{out}}(\tau)$.
This uniqueness follows from non-duplicated key fields and typed tool interfaces that correspond to well-defined backend relations.
At execution time, if exactly one matching output-value tuple is found, that tuple is returned.
If no matching output-value tuple is found, the runtime reports that the requested output datatypes cannot be obtained from the provided arguments through the called tool.

\paragraph{Termination.}
The environment finalizes a query under the following conditions:
\textbf{(1) Step budget reached:} Each model response counts as one step, including retrieval requests, tool calls, final answers, and invalidly formatted actions.
\textbf{(2) Final answer submitted:} When the agent submits a final answer, the runtime checks both the answer string and the internal typed state. 

\subsection{Metric Details}
\label{app:metric-details}

\subsubsection{Metrics Formulation}
\label{app:full-metric-definition}

Let $\mathcal{Q}$ be the evaluated query set. 
For each query $i \in \mathcal{Q}$, let $c_i \in \{0,1\}$ indicate whether the final answer is correct, let $T_i$ be the number of interaction turns, let $S_i$ and $C_i$ be the numbers of retrieval and tool-call turns, let $N_i^{\mathrm{inv}}$ be the number of structurally invalid tool calls, and let $N_i^{\mathrm{untr}}$ be the number of tool calls rejected because at least one argument value comes from a noisy-tool response.
Let $\mathcal{D}_{i,0}$ denote the initial datatype set, $\mathcal{D}_i^{\mathrm{exec}}$ the set of datatypes produced by successful executed tool calls, and $\mathcal{D}_i^{\mathrm{gt}}$ the process-level ground-truth datatype set derived from all valid paths of the task.

\paragraph{Task Completion Metrics.}
Accuracy measures whether the agent produces the correct final answer after following the runtime protocol. 
A prediction is counted as correct only when the submitted answer matches the gold answer under normalized containment matching and the internal typed state contains $\mathcal{Y}_i$.
Formally, we compute
\begin{equation}
\begin{aligned}
\mathrm{Acc} = \frac{1}{|\mathcal{Q}|}\sum_{i \in \mathcal{Q}} c_i.
\end{aligned}
\end{equation}

We also report execution-ground-truth precision, denoted as EGT Precision, to measure how much of the agent's executed datatype trajectory overlaps with the process-level ground truth. 
For each query, it computes the fraction of successfully executed datatypes that appear in the ground-truth datatype set:
\begin{equation}
\begin{aligned}
\mathrm{EGT\mbox{-}Prec} 
= \frac{1}{|\mathcal{Q}'|}\sum_{i \in \mathcal{Q}'}
\frac{|\mathcal{D}_i^{\mathrm{exec}} \cap \mathcal{D}_i^{\mathrm{gt}}|}
{|\mathcal{D}_i^{\mathrm{exec}}|},
\end{aligned}
\end{equation}
where $\mathcal{Q}'=\{i \in \mathcal{Q}: |\mathcal{D}_i^{\mathrm{exec}}|>0\}$ excludes queries with no executed datatypes.

Finally, we report the average number of interaction turns:
\begin{equation}
\begin{aligned}
\mathrm{AvgTurns} = \frac{1}{|\mathcal{Q}|}\sum_{i \in \mathcal{Q}} T_i.
\end{aligned}
\end{equation}
Each model response counts as one turn, including retrieval requests, tool calls, final answers, and invalidly formatted actions.

\paragraph{Exploration Behavior Metrics.}
We measure how broadly the agent explores the datatype space using the mean explored datatype count. 
For each query, the explored datatype closure $\mathcal{D}_i^{\mathrm{exp}}$ is initialized from the input datatypes $\mathcal{D}_{i,0}$ and expanded during the interaction by successful executions and datatype outputs implied by retrieved tools. 
The per-query explored datatype count is
\begin{equation}
\begin{aligned}
\mathrm{EDT}_i = |\mathcal{D}_i^{\mathrm{exp}} \setminus \mathcal{D}_{i,0}|,
\end{aligned}
\end{equation}
and the reported mean is
\begin{equation}
\begin{aligned}
\mathrm{MeanEDT} = \frac{1}{|\mathcal{Q}|}\sum_{i \in \mathcal{Q}} \mathrm{EDT}_i.
\end{aligned}
\end{equation}

We also report the search-to-call ratio, which compares how often the agent retrieves tools against how often it executes tools:
\begin{equation}
\begin{aligned}
\mathrm{S/C\ Ratio} = \frac{\sum_{i \in \mathcal{Q}} S_i}{\sum_{i \in \mathcal{Q}} C_i}.
\end{aligned}
\end{equation}
A larger value indicates more retrieval activity relative to tool execution.

\paragraph{Execution Quality Metrics.}
We use the invalid tool-call ratio to measure how often the agent attempts tool calls that violate the runtime protocol for structural reasons.
A tool call is counted as invalid when it fails parsing or violates tool-call constraints, such as calling an unretrieved tool, providing mismatched argument keys, or calling a tool before its required input datatypes are available.
The metric is computed as
\begin{equation}
\begin{aligned}
\mathrm{ITCR} = \frac{\sum_{i \in \mathcal{Q}} N_i^{\mathrm{inv}}}{\sum_{i \in \mathcal{Q}} C_i}.
\end{aligned}
\end{equation}

We also report the untrusted input rejection rate, which measures how often the agent uses a value returned by a noisy tool as an argument to another tool call.
Such calls are rejected because noisy-tool outputs are not treated as trusted runtime values.
The metric is computed as
\begin{equation}
\begin{aligned}
\mathrm{UIRR} = \frac{\sum_{i \in \mathcal{Q}} N_i^{\mathrm{untr}}}{\sum_{i \in \mathcal{Q}} C_i}.
\end{aligned}
\end{equation}

\subsubsection{Illustrative Examples for Metrics}
\label{app:metric-illustrative-examples}

\begin{figure*}[t]
\centering
\includegraphics[width=1\textwidth]{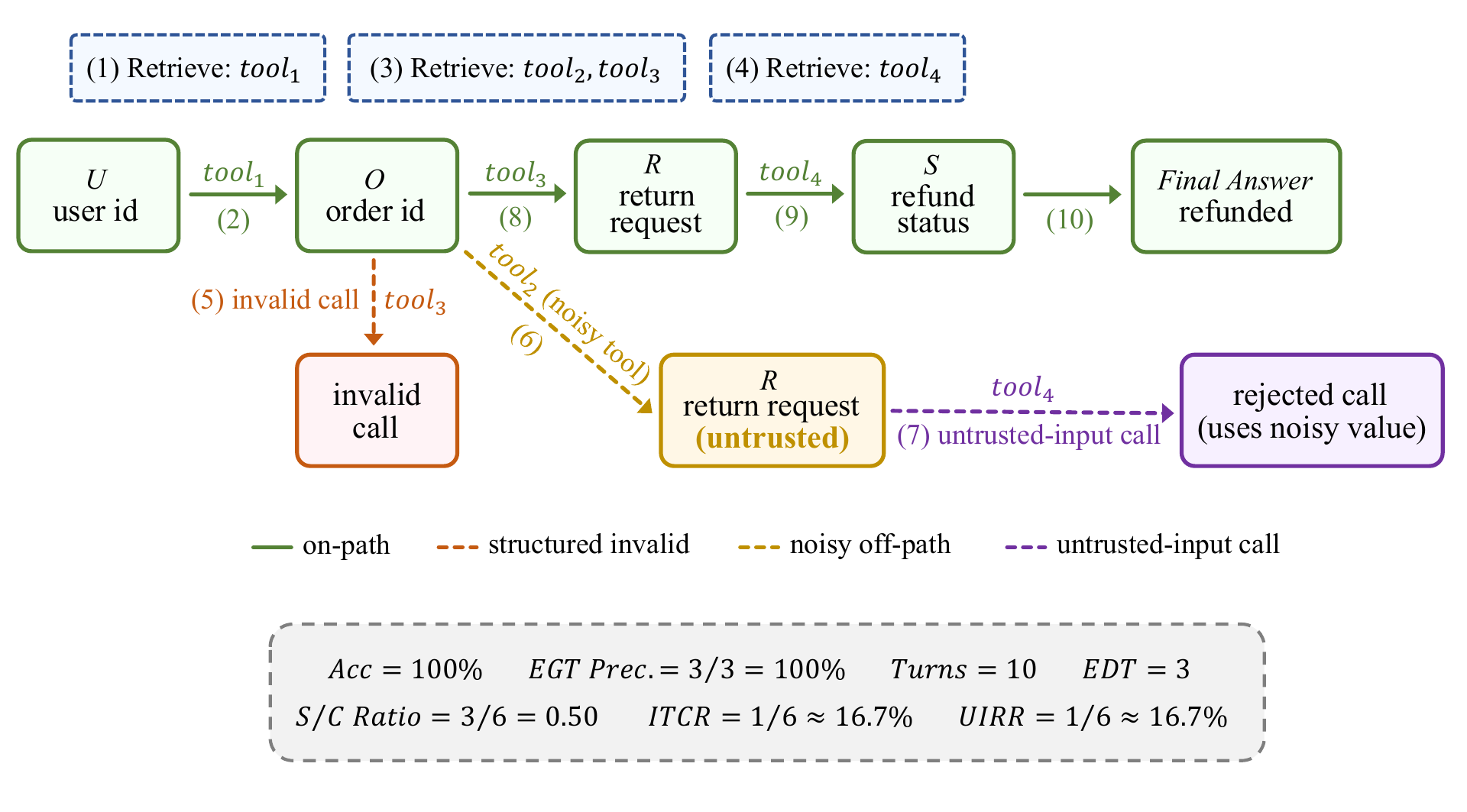}
\vspace{-0.08in}
\caption{
Illustrative trajectory for computing the seven evaluation metrics.
The query starts from $U$ and targets $S$ along the ground-truth path $U \rightarrow O \rightarrow R \rightarrow S$.
The agent makes one invalid call, one noisy call, and one later untrusted-input-rejected call.
This trajectory yields $100\%$ Accuracy, $100\%$ EGT Precision, $10$ turns, $3$ explored datatypes, an S/C Ratio of $0.50$, an ITCR of $16.7\%$, and a UIRR of $16.7\%$.
}
\label{fig:metric-illustrative-example}
\vspace{-0.12in}
\end{figure*}

We illustrate the seven metrics using the trajectory in Figure~\ref{fig:metric-illustrative-example}. 
The example is synthetic, but follows the same evaluation logic as our implementation. 
Let the initial datatype be $U$ and the target datatype be $S$, where $U$ denotes a user ID and $S$ denotes a refund status. 
The gold answer is \texttt{refunded}. 
The ground-truth path is
\begin{equation}
U \rightarrow O \rightarrow R \rightarrow S,
\end{equation}
where $O$ is an order ID and $R$ is a return request ID. 
Thus, the ground-truth datatype set is
\begin{equation}
\mathcal{D}^{\mathrm{gt}}_q = \{U,O,R,S\}.
\end{equation}

The trajectory contains ten actions in total. 
There are three retrieval turns, corresponding to steps (1), (3), and (4); six tool-call turns, corresponding to steps (2), (5), (6), (7), (8), and (9); and one final-answer turn at step (10).
Among the tool calls, step (5) is a structurally invalid call and produces no datatype.
Step (6) invokes a noisy tool and returns an \emph{untrusted} value for datatype $R$.
Step (7) then attempts to use that noisy return value as an argument in a later tool call, which is rejected and therefore counted by UIRR rather than ITCR.
The trusted successful tool calls are steps (2), (8), and (9), which produce $O$, $R$, and $S$, respectively.
Finally, the agent outputs \texttt{refunded}.

\paragraph{Accuracy.}
Accuracy uses the grounded criterion: the answer must match the gold answer and the target datatype must be reached. 
Here, the final answer is \texttt{refunded}, and $S$ is reached. 
Therefore,
\begin{equation}
\mathrm{Acc}(q)=1.
\end{equation}
For a one-query evaluation set, this corresponds to $100\%$ accuracy.

\paragraph{Executed Ground-Truth Datatype Precision.}
EGT Precision measures how many executed datatypes lie on the ground-truth path.
Only \emph{trusted} successful tool outputs are added to the executed datatype set.
Thus, the noisy return from step (6) is not counted here, while the trusted successful tool calls in steps (2), (8), and (9) produce
\begin{equation}
\mathcal{D}^{\mathrm{exec}}_q = \{O,R,S\}.
\end{equation}
All of them are in $\mathcal{D}^{\mathrm{gt}}_q$.
Therefore,
\begin{equation}
\mathrm{EGTPrec}(q)
\;=\; \frac{|\{O,R,S\}|}{|\{O,R,S\}|}
\;=\; \frac{3}{3}
\;=\; 100\%.
\end{equation}
This shows that an agent can still encounter invalid and noisy behavior while remaining perfectly on-path in terms of trusted executed datatypes.

\paragraph{Average Turns.}
The trajectory contains ten recorded turns in total: three retrieval turns, six tool-call turns, and one final-answer turn.
Thus,
\begin{equation}
\mathrm{Turns}(q)=10.
\end{equation}
For a one-query evaluation set, the Avg. Turns value is $10$.

\paragraph{Mean Explored Datatypes.}
EDT counts newly discovered datatypes beyond the initial input datatype.
Across the three retrieval turns, the retrieved tool set makes datatypes $O$, $R$, and $S$ reachable from the initial datatype $U$ under the closure rule used in evaluation.
Therefore, the explored datatype set is
\begin{equation}
\mathcal{D}^{\mathrm{explore}}_q = \{U,O,R,S\}.
\end{equation}
Since $U$ is the initial datatype, the discovered datatypes are
\begin{equation}
\mathcal{D}^{\mathrm{new}}_q = \{O,R,S\}.
\end{equation}
Therefore,
\begin{equation}
\mathrm{EDT}(q)=3.
\end{equation}
This metric captures how many new datatypes become reachable through retrieval, regardless of whether all later calls succeed.

\paragraph{Search-to-Call Ratio.}
The agent performs three retrieval turns and six tool-call turns.
Both the invalid call in step (5) and the untrusted-input-rejected call in step (7) are still counted as call turns.
Therefore,
\begin{equation}
\mathrm{S/C}(q)=\frac{3}{6}=0.50.
\end{equation}
This ratio captures the balance between tool search and tool execution.

\paragraph{Invalid Tool Call Rate.}
ITCR measures the fraction of attempted tool calls that are invalid for structural or protocol reasons. 
The agent makes six tool-call attempts in total, and exactly one of them, step (5), is counted by ITCR.
The rejected call in step (7) is \emph{not} counted by ITCR because the current implementation records it separately as an untrusted-input rejection.
Thus,
\begin{equation}
\mathrm{ITCR}(q)=\frac{1}{6}\approx 16.7\%.
\end{equation}

\paragraph{Untrusted Input Rejection Rate.}
UIRR measures the fraction of attempted tool calls that are rejected because one or more arguments come from untrusted inputs. 
Here, exactly one tool-call attempt, step (7), is rejected for that reason, out of six total call attempts.
Therefore,
\begin{equation}
\mathrm{UIRR}(q)=\frac{1}{6}\approx 16.7\%.
\end{equation}

In the full evaluation, Accuracy, EGT Precision, Avg. Turns, and Mean EDT are averaged over queries, while S/C Ratio, ITCR, and UIRR are computed from global totals across all evaluated trajectories.

\subsection{Blocking Details}
\label{app:blocker-details}


Blocking in \bench{} is implemented by replacing selected tools in the retrieval results. 
When the agent retrieves a tool that has been selected for blocking, the environment removes that tool from the returned candidate list and returns its corresponding additional tool instead. 
The agent therefore observes a normal retrieval result, but the tool it would otherwise rely on has been replaced. 
This behavior is determined by two components: how the additional tools are constructed and which tools are selected for blocking in each task. 
We describe these two components below in Appendix~\ref{app:noisy-tool-construction} and Appendix~\ref{app:blocked-tool-selection}, respectively.

\subsubsection{Blocking Formalization}
\label{app:blocking-formalization}


We formalize retrieval-time blocking as a perturbation applied to the tool candidates returned by the retriever. 
Let $\mathcal{T}$ denote the original tool set. 
For each task instance $i$, let $\mathcal{T}^{(i)}_{\mathrm{blk}} \subseteq \mathcal{T}$ be the set of baseline tools selected for blocking. 
This set is hidden from the agent.

Let $\mathcal{B}$ denote the set of blocking perturbation types. 
For a blocked tool $\tau \in \mathcal{T}^{(i)}_{\mathrm{blk}}$ and a perturbation type $b \in \mathcal{B}$, the environment constructs an additional tool
\begin{equation}
\phi_b(\tau) \in \mathcal{T}^{+},
\end{equation}
where $\mathcal{T}^{+}$ is the set of additional tools. 
The function $\phi_b$ maps a blocked tool to the corresponding replacement tool under perturbation type $b$.

At interaction step $t$, suppose the unperturbed retriever returns a ranked candidate list
\begin{equation}
L_t = [\tau_1,\ldots,\tau_K].
\end{equation}
The blocking environment transforms $L_t$ into the observed retrieval list
\begin{equation}
\widetilde{L}_t
=
\bigoplus_{k=1}^{K} \rho_i(\tau_k),
\end{equation}
where $\oplus$ denotes list concatenation and
\begin{equation}
\rho_i(\tau_k)
=
\begin{cases}
[\phi_b(\tau_k)]_{b\in\mathcal{B}}, 
& \tau_k \in \mathcal{T}^{(i)}_{\mathrm{blk}},\\
[\tau_k], 
& \tau_k \notin \mathcal{T}^{(i)}_{\mathrm{blk}}.
\end{cases}
\end{equation}
Thus, if a retrieved baseline tool is selected for blocking, the original tool is removed from the candidate list and replaced by its corresponding additional tools. 
If a retrieved tool is not selected for blocking, it remains unchanged.

A retrieval-time blocking event occurs at step $t$ if the original retrieval result contains at least one selected blocked tool:
\begin{equation}
E^{(i)}_{\mathrm{blk}}(t)
=
\mathbb{I}
\left[
L_t \cap \mathcal{T}^{(i)}_{\mathrm{blk}}
\neq \emptyset
\right].
\end{equation}
The number of blocked baseline tools encountered in the retrieval result is
\begin{equation}
N^{(i)}_{\mathrm{blk}}(t)
=
\left|
L_t \cap \mathcal{T}^{(i)}_{\mathrm{blk}}
\right|.
\end{equation}
The agent observes only $\widetilde{L}_t$, rather than the original retrieval list $L_t$ or the hidden blocked set $\mathcal{T}^{(i)}_{\mathrm{blk}}$.

\subsubsection{Additional Tool Construction}
\label{app:noisy-tool-construction}


For each original tool $\tau \in \mathcal{T}$, we construct three corresponding additional tools, one for each blocking perturbation.
The additional tools are generated by the same generation model used in query construction, $M_{\mathrm{gen}}=\textit{GPT-5.2}$, conditioned on the original tool name, description, input datatype set 
$\mathcal{D}_{\mathrm{in}}(\tau)$, and output datatype set 
$\mathcal{D}_{\mathrm{out}}(\tau)$.
The goal is to preserve the retrieval-facing form of the original tool while modifying the behavior or functionality according to the blocking type.

For explicit failure and implicit failure blocks, the additional tools are designed to be indistinguishable from the original tool based on their surface information. 
Their names follow the same template \texttt{Get\_<Output>\_From\_<Input>[\_<Variant>]} and may differ only in the optional suffix. 
Their descriptions are written in the same style as the original description, with similar length, wording structure, and functional meaning. 
Thus, from the agent's perspective, these tools appear to provide the same input-output mapping as the original tool. 
The difference is only revealed after execution: the explicit failure tool returns an error, while the implicit failure tool returns an impossible or counterfactual value.

For semantic misleading blocks, the additional tool also follows the same naming format and remains semantically related to the original tool. 
However, it is constructed to support a different function. 
Its description explicitly states this actual function, while keeping a style and length similar to the original description. 
Therefore, unlike the explicit and implicit failure tools, a semantic misleading tool can be distinguished from the original tool by carefully reading its description.

\subsubsection{Selection of Blocked Tools}
\label{app:blocked-tool-selection}

We next describe how the hidden blocked tool set $\mathcal{T}^{(i)}_{\mathrm{blk}}$ is selected for each task instance $i$ in the main block setting.
This selection is performed before the interaction starts and is hidden from the agent.
The goal is to perturb useful tools while preserving at least one feasible solution path, so that the task remains solvable but requires recovery through alternative tool-use paths.

\paragraph{Path-Based Blocking Objective.}
Let $\Pi_i$ denote the catalog of valid solution paths for task instance $i$.
For each path $\pi=(\tau_1,\ldots,\tau_K)\in\Pi_i$, let $\mathrm{Tools}(\pi)=\{\tau_1,\ldots,\tau_K\}$ denote the unordered set of tools appearing on that path.
Thus, the blocker selects which paths to affect, without considering the order in which those paths may be explored.
Let
\begin{equation}
\mathcal{P}_i = \bigcup_{\pi \in \Pi_i} \mathrm{Tools}(\pi)
\end{equation}
be the set of unique tools appearing in the path catalog.
For a candidate blocked tool set $C \subseteq \mathcal{P}_i$, a path is blocked if it contains at least one tool in $C$:
\begin{equation}
\mathrm{Blk}_i(C)
=
\{\pi \in \Pi_i : \mathrm{Tools}(\pi) \cap C \neq \emptyset\}.
\end{equation}
The number of remaining feasible paths is
\begin{equation}
n_i(C)
=
|\Pi_i| - |\mathrm{Blk}_i(C)|.
\end{equation}

\paragraph{Feasibility Constraints.}
The blocker is constrained to preserve at least one feasible solution path.
A candidate blocked tool set $C$ is feasible if
\begin{equation}
n_i(C)\geq n_{\min},
\end{equation}
\begin{equation}
|n_i(C)-n^\star|\leq\delta.
\end{equation}
Our main block setting uses a target remaining-path count of $n^\star=1$, a tolerance of $\delta=1$, and a minimum remaining-path count of $n_{\min}=1$.
Since $n_i(C)$ is an integer, these constraints require $n_i(C)\in\{1,2\}$.

\paragraph{Finite Candidate Enumeration.}
Blocked tool candidate selection is performed over a finite enumerated pool rather than over all subsets of $\mathcal{P}_i$.
Let $\widetilde{\mathcal{C}}_i$ denote this finite candidate pool.
The blocker first includes the empty candidate $C=\emptyset$, and then enumerates tool combinations up to size $K_{\mathrm{blk}}$, the maximum selected tool count.
Enumeration stops once the maximum number of candidate combinations is reached.
Thus, $\widetilde{\mathcal{C}}_i$ defines a bounded rule-based search space over candidate blocked tool sets.
The feasible candidate set is then
\begin{equation}
\mathcal{C}_i
=
\{C\in\widetilde{\mathcal{C}}_i: C \text{ is feasible}\}.
\end{equation}
In \bench{}, the enumerated combinations for all cases are within the maximum candidate number limit.

\paragraph{Candidate Selection and Tie-Breaking.}
Among feasible candidates, the blocker minimizes the absolute deviation from the target remaining-path count:
\begin{equation}
d_i(C)=|n_i(C)-n^\star|.
\end{equation}
The best candidate pool is
\begin{equation}
\mathcal{C}^{\star}_i
=
\arg\min_{C\in\mathcal{C}_i} d_i(C).
\end{equation}
Let $\sigma_0$ denote the global seed, and define the task-specific seed as $\sigma_i = \sigma_0 + \mathrm{hash}(i)$.
If multiple candidates remain in $\mathcal{C}^{\star}_i$, the final blocked tool set is sampled by seeded tie-breaking:
\begin{equation}
\mathcal{T}^{(i)}_{\mathrm{blk}}
\sim
\mathrm{Uniform}
\left(
\mathcal{C}^{\star}_i;
\sigma_i
\right).
\end{equation}
This seeded tie-breaking makes the selected blocked tool set reproducible for each task.

\paragraph{Instantiating Blocking Perturbations.}
After $\mathcal{T}^{(i)}_{\mathrm{blk}}$ is selected, each selected tool is instantiated with the blocking perturbation types defined in the retrieval-time blocking formalization.
In the main block setting, we use $\mathcal{B}=\{b_{\mathrm{exp}}, b_{\mathrm{imp}}, b_{\mathrm{sem}}\}$, where $b_{\mathrm{exp}}$ denotes explicit failure, $b_{\mathrm{imp}}$ denotes implicit failure, and $b_{\mathrm{sem}}$ denotes semantic misleading.
For every selected tool $\tau \in \mathcal{T}^{(i)}_{\mathrm{blk}}$, the environment constructs one additional tool for each $b\in\mathcal{B}$:
\begin{equation}
\tau
\mapsto
\{\phi_{b_{\mathrm{exp}}}(\tau),
\phi_{b_{\mathrm{imp}}}(\tau),
\phi_{b_{\mathrm{sem}}}(\tau)\}.
\end{equation}
During retrieval, these additional tools replace the original selected tool according to the retrieval-time blocking transformation defined above.

\paragraph{Returned-list Budget under Blocking.}
We use the same per-retrieval cap $\Lambda_{\mathrm{ret}}^{\mathrm{cap}}=30$ in the block setting, which keeps the retrieval context size comparable to the default setting.
Suppose the original typed match returns $K$ executable tools, and $m=N^{(i)}*{\mathrm{blk}}(t)$ of them are selected for blocking.
For each selected tool $\tau$, the environment removes $\tau$ and inserts the three blocking replacements ${\phi*{b_{\mathrm{exp}}}(\tau), \phi_{b_{\mathrm{imp}}}(\tau), \phi_{b_{\mathrm{sem}}}(\tau)}$.
After this replacement step, the returned list contains $K-m+3m=K+2m$ primary tools.
The environment then fills the remaining slots with ordinary noisy tools, up to the cap:
\begin{equation}
\Lambda_{\mathrm{ret}}^{\mathrm{cap}} - (K+2m).
\end{equation}
In this benchmark, $K+2m \leq \Lambda_{\mathrm{ret}}^{\mathrm{cap}}$, so this quantity is always non-negative.
Thus, blocking replacements are always retained, and the cap only controls how many ordinary noisy distractors are appended.

\paragraph{Unresolved Cases.}
If no feasible candidate exists, the blocker marks the task as unresolved and creates no replacement tools.
In that case,
\begin{equation}
\mathcal{T}^{(i)}_{\mathrm{blk}}=\emptyset,
\end{equation}
and retrieval proceeds unchanged.
Under the setting used in our experiments, all task instances are resolved successfully.



\subsubsection{Practical Relevance and Significance of Blocking}
\label{app:blocker-significance}

The blocking setting is designed to approximate realistic failure modes in large-scale tool ecosystems, where agents rarely interact with a perfectly curated and reliable set of tools. In practical deployments, retrieved tools may be unavailable, deprecated, stale, misconfigured, or only superficially related to the agent's current need. Such failures are especially challenging in retrieval-mediated tool use because the agent must decide not only which tool to call, but also whether the retrieved tool is trustworthy and whether its observation should be incorporated into the plan. Our blocking mechanism therefore evaluates an agent's ability to detect unreliable tool access, avoid misleading evidence, and recover by searching for alternative solutions.

\paragraph{Explicit failure blocks.}
Explicit failure blocks simulate tools that appear relevant at retrieval time but fail once invoked.
This situation is common in real systems when an API endpoint is deprecated, an external service is temporarily unavailable, authentication has expired, or a backend schema has changed. 
For example, in a retail workflow, an agent may retrieve a tool that appears to check whether an item is available at a store, but the call returns an error because the inventory service is unavailable. 
Although such failures are relatively easy to recognize after execution, they still require adaptive planning: the agent must avoid repeatedly calling the failed tool, reinterpret the missing information as a planning constraint, and search for another route to the final answer.

\paragraph{Implicit failure blocks.}
Implicit failure blocks model a subtle class of tool failures in which the selected tool appears appropriate according to its description, but the returned observation is semantically invalid for the task. 
Unlike explicit failures, the tool does not raise an error or signal incompatibility; instead, it produces an output that may be syntactically well-formed but is either irrelevant to the user’s request or counterfactual with respect to basic domain knowledge. 
For example, a tool invoked to retrieve an exchange rate may return a negative value, or a weather tool may report a temperature below absolute zero. 
In other cases, the tool may return information that is structurally valid but unrelated or unusable for the intended domain, despite the tool description suggesting that it should be helpful. 
This setting tests whether agents can recognize that a tool observation is not reliable merely because it comes from a seemingly relevant tool, and whether they can reject outputs that are unhelpful, nonsensical, or inconsistent with commonsense constraints rather than incorporating them into subsequent reasoning.

\paragraph{Semantic misleading blocks.}
Semantic misleading blocks represent retrieval noise in which the returned tool is semantically close to the desired tool but supports a different function. This reflects a common problem in large tool repositories: many tools share similar names, overlapping descriptions, or related domain vocabulary, yet differ in their precise semantics. For example, an agent searching for a tool to obtain the estimated delivery date of an order may instead retrieve a tool for estimating the pickup date of an in-store return. Unlike explicit and implicit failure blocks, semantic misleading blocks can often be detected before execution through careful inspection of the tool description and input-output schema. However, they still increase the difficulty of tool selection, particularly under long-horizon planning, where the agent must keep track of many intermediate sub-goals and avoid following superficially relevant but functionally incorrect tools.


Together, these three blocking types cover complementary sources of unreliability. 
Explicit failures test recovery from visible tool unavailability; implicit failures test robustness to silent misinformation; and semantic misleading blocks test fine-grained tool understanding under retrieval noise. 
By evaluating agents under all three conditions, \bench{} measures not only whether an agent can solve a task when the correct tools are available, but also whether it can maintain reliable planning behavior when the tool ecosystem itself becomes partially unreliable.



\section{Additional Experiment Results}
\label{app:additional-experiment-results}

\subsection{Evaluation Robustness}
\label{app:evaluation-robustness}


\begin{table*}[htbp]
\small
\resizebox{\linewidth}{!}{%
\begin{tabular}{lccccccc}
\toprule
\textbf{Models} & \makecell{\textbf{Radius} \\ \textbf{(Accuracy)}} (\%) & \makecell{\textbf{Radius} \\ \textbf{(EGT Prec.)}} (\%) & \makecell{\textbf{Radius} \\ \textbf{(Avg. Turns)}} & \makecell{\textbf{Radius} \\ \textbf{(Mean EDT)}} & \makecell{\textbf{Radius} \\ \textbf{(S/C Ratio)}}  & \makecell{\textbf{Radius} \\ \textbf{(ITCR)}} (\%) & \makecell{\textbf{Radius} \\ \textbf{(UIRR)}} (\%) \\
\midrule
\textit{Qwen3-8B}                 & 0.00 & 3.93 & 3.55 & 0.75 & 0.05 & 1.58 & 0.11 \\
\textit{Qwen3-14B}                & 1.07 & 3.81 & 3.50 & 0.61 & 0.01 & 0.86 & 0.67 \\
\textit{Qwen3-32B}                & 1.68 & 3.12 & 0.71 & 0.69 & 0.18 & 2.03 & 2.18 \\
\textit{Llama-3.1-8B-Instruct}    & 0.00 & 5.42 & 2.36 & 0.80 & 0.26 & 4.02 & 1.87 \\
\textit{Llama-3.3-70B-Instruct}   & 4.28 & 3.58 & 1.58 & 0.75 & 0.22 & 3.03 & 1.11 \\
\textit{DeepSeek-V4-Flash}        & 5.23 & 3.20 & 2.64 & 0.95 & 0.23 & 1.75 & 1.20 \\
\textit{Gemini-3.1-Pro}           & 4.59 & 2.27 & 1.01 & 0.78 & 0.11 & 0.56 & 0.31 \\
\textit{Gemini-3.5-Flash}         & 5.31 & 2.90 & 4.26 & 0.82 & 1.26 & 1.35 & 0.00 \\
\textit{GPT-5.4-Mini}             & 1.84 & 5.35 & 0.60 & 0.87 & 0.11 & 4.29 & 1.98 \\
\textit{GPT-5.4}                  & 5.38 & 3.39 & 0.74 & 0.69 & 0.18 & 1.55 & 0.95 \\
\hline
\textbf{\textit{Average}}         & 2.94 & 3.70 & 2.10 & 0.77 & 0.26 & 2.10 & 1.04 \\
\bottomrule
\end{tabular}%
}
\caption{95\% Confidence Intervals for all metrics. We employ non-parametric bootstrap~\cite{justus2024bootstrap} resampling with 10,000 iterations over the evaluated queries ($N = 327$) to estimate statistical uncertainty. The reported \textbf{Radius} values denote the half-width of the 95\% confidence interval (i.e., the $\pm$ value) for Accuracy, Executed Ground-Truth Datatype Precision (EGT Prec.), Average Turns (Avg. Turns), Mean Explored Datatype (Mean EDT), Search-to-Call Ratio (S/C Ratio), invalid Tool Call Rate (ITCR), and Untrusted Input Rejection Rate (UIRR).}
\label{tab:confidence-interval-retail}
\end{table*}

\paragraph{Theoretical Bounds.} To quantify statistical uncertainty in the default-setting results, we estimate 95\% confidence intervals by non-parametric bootstrap~\cite{justus2024bootstrap,marcon,wang2025diversity,wang2025prospect,zong2025critical} resampling 10,000 times over the evaluated queries.
As shown in Table~\ref{tab:confidence-interval-retail}, the resulting intervals are generally narrow across models. 
Averaged over the evaluated models, the radius is $2.94$ percentage points for Accuracy, $3.70$ percentage points for EGT Prec., $2.10$ turns for Avg. Turns, $0.77$ datatypes for Mean EDT, $0.26$ for S/C Ratio, $2.10$ percentage points for ITCR, and $1.04$ percentage points for UIRR. 
These intervals are small relative to many of the performance gaps in Table~\ref{tab:main_unblocked_results_grounded}, suggesting that the main trends are unlikely to be artifacts of test-set sampling noise.

\begin{figure}[t]
    \centering
    \includegraphics[width=1\linewidth]{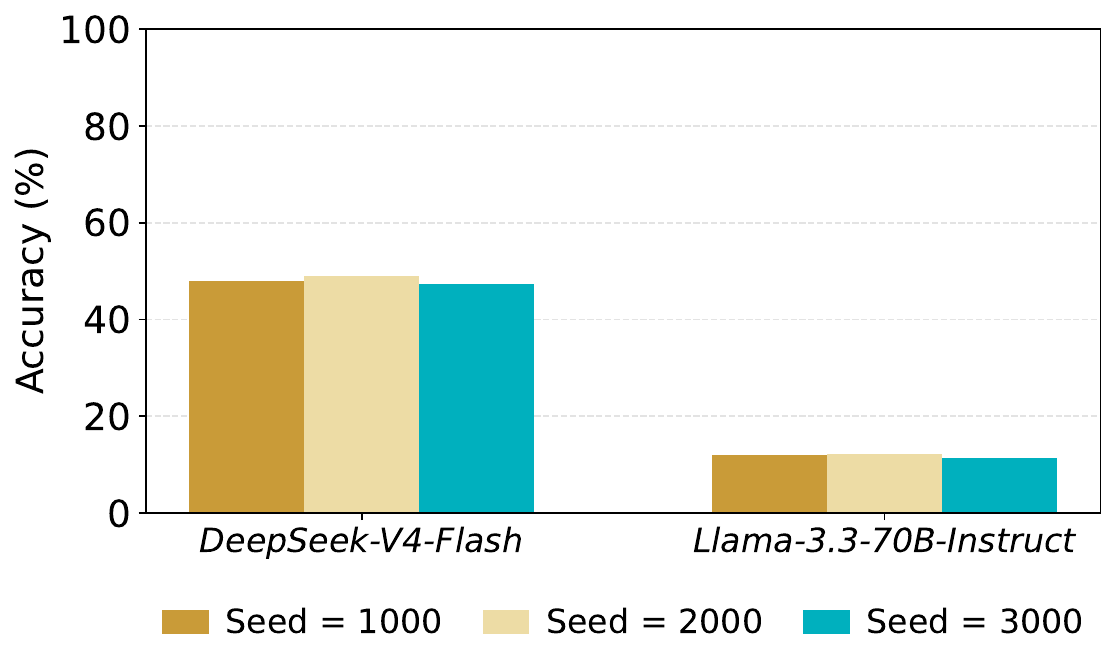}
    \vspace{-0.3in}
    \caption{Accuracy (\%) of \textit{DeepSeek-V4-Flash} and \textit{Llama-3.3-70B-Instruct} across different seeds. Results demonstrate low variations with different seeds, with variations not exceeding 3\% across seeds.}
    \label{fig:seed_acc}
    \vspace{-0.2in}
\end{figure}
\paragraph{Empirical Results.} To further empirically examine robustness to seed variation, we reran the blocked evaluation of \textit{DeepSeek-V4-Flash} and \textit{Llama3.3-70B-Instruct} with seeds 1000, 2000, and 3000 while keeping the model and all other configuration choices fixed. 
As shown in Figure~\ref{fig:seed_acc}, the results are highly consistent across seeds. 
These small fluctuations suggest that the block setting conclusions for this model are not sensitive to the particular seeded choice of blocker configuration.

\subsection{Retriever Robustness}
\label{app:retriever-robustness}


To verify that the main benchmark results are not primarily limited by retriever coverage, we construct a single-step tool retrieval evaluation. 
For each released executable tool $\tau$, we create a one-step task whose initial datatype set is $\mathcal{D}_{\mathrm{in}}(\tau)$ and whose target datatype set is $\mathcal{D}_{\mathrm{out}}(\tau)$.
The task therefore has a valid solution consisting of exactly one tool call after the correct tool is retrieved. 
We instantiate these tasks using the same query construction pipeline as the main benchmark. 
Concrete input values are sampled from the backend, the resulting instance is checked to be executable, and $M_{\mathrm{gen}}$ verbalizes the structured task into a natural-language query. 
The gold answer is obtained by executing the corresponding tool on the instantiated input values.

We run this evaluation on \textit{Qwen3-8B}, a relatively weak model in the main benchmark. 
This setting provides a conservative sanity check because the same model performs poorly on full multi-step tasks but can still be tested on whether it can identify and invoke the correct tool in a one-step setting. 
The ideal trajectory for each single-step query contains three turns, corresponding to one retrieval request, one tool call, and one final answer.
As in the main benchmark, retrieval does not return only the executable tool.
For a single-step task constructed from $\tau$, the returned candidate list may also include noisy variants paired with $\tau$, denoted as $\mathcal{N}(\tau)$.
Thus, the model must select the executable tool from a mixed set of relevant-looking candidates.

As shown in Table~\ref{tab:single-step-retriever-results}, \textit{Qwen3-8B} achieves $85.95\%$ accuracy over $185$ single-step queries, with the ground-truth tool included in $92.43\%$ of the returned retrieval lists.
Its average number of turns is $3.18$, and its search-to-call ratio is $0.95$, both close to the ideal single-step trajectory of one retrieval request, one tool call, and one final answer.
The invalid tool-call rate is $0.00\%$.
These results indicate that the retriever and runtime protocol provide reliable access to the correct tools in single-step settings. 
Thus, the much lower performance of the same model on the full benchmark is better explained by multi-step planning, state tracking, and execution decisions than by failures of the retriever.


\begin{table}[t]
\centering
\small
\resizebox{0.6\columnwidth}{!}{
\begin{tabular}{lc}
\toprule
\textbf{Metric} & \textbf{\textit{Qwen3-8B}} \\
\midrule
Accuracy (\%) & 85.95 \\
GT Tool Retrieved (\%) & 92.43 \\
Avg. Turns & 3.18 \\
S/C Ratio & 0.95 \\
ITCR (\%) & 0.00 \\
EGT Prec. (\%) & 96.45 \\
\bottomrule
\end{tabular}
}
\caption{Single-step tool retrieval evaluation on \textit{Qwen3-8B}. 
Each query is constructed from one tool by using its input datatypes as the initial information and its output datatype as the target. 
The ideal trajectory contains three turns, corresponding to retrieval, tool call, and final answer.}
\label{tab:single-step-retriever-results}
\end{table}

\section{Case Study}
\label{app:case-study}

\subsection{Error Case Study}

We present representative raw trajectories for the mechanisms analyzed in Appendix~\ref{sec:error-analysis}. 
Figure~\ref{fig:case_irrecoverable_drift_wrong_tool_value} shows \textit{Irrecoverable Drift}: the model obtains useful intermediate evidence, then takes a non-progress step and turns an off-path tool value into the final answer. 
Figure~\ref{fig:case_irrecoverable_drift_search_exhaustion} shows a drifted trajectory that ends through \emph{Search Exhaustion}: after partial progress, the model continues retrieving and calling tools but never grounds the target answer. 
Figure~\ref{fig:case_interaction_breakdown} shows \textit{Format Error}, where repeated invalid tool calls terminate the run before the agent can stabilize on a valid solution path. 
Figure~\ref{fig:case_silent_failure_output_accepted} shows how a blocked alternative can contaminate the trajectory when a silent failure returns a plausible target-typed value. 
Together, these cases show the same separation used in the error analysis: where the trajectory breaks, how corrupted tool outputs affect later decisions, and how the model terminates after the trajectory is no longer grounded.

\subsection{Data Case Study}
\label{app:data-case-study}

To make the tool-use path structure concrete, we provide one representative retail task and several tool definitions from an available solution path. Figure~\ref{fig:case_data_query_profile} shows the natural-language query and one valid path for \texttt{retail\_task\_0001}. Figure~\ref{fig:case_data_tools} shows representative tool definitions from the same task family. The agent-visible schema consists of the function name, description, parameters, and strictness constraints; datatype annotations such as input datatype and output datatype are internal benchmark metadata used for retrieval, dependency validation, progress-call annotation, and evaluation. This distinction is important for the error analysis: models must infer useful tool transitions from natural-language tool schemas, while our diagnostics measure progress against feasible solution paths.

These examples also show why the benchmark is not a single-hop lookup task. The representative path begins from a fulfillment-side clue, moves through order and return records, and then enters the payment subflow before reaching the requested account field. A failure can therefore occur before the model obtains useful evidence, after it drifts away from a valid solution path, or after a corrupted executable-looking output is treated as grounded evidence. The data case study grounds the aggregate results by showing the structure that makes progress, drift, recovery, blocked-branch contamination, and format error measurable.


\section{Justifications and Design Choices}
\label{app:justifications}


\paragraph{Retriever.} 


\bench{} uses a simplified retriever to expose the tool space through controlled natural-language queries. 
This design abstracts away some complexity of real-world retrieval systems, since the underlying matching is grounded in the predefined datatype structure rather than noisy documentation, imperfect ranking, or dynamically changing tool catalogs. 
As a result, tool retrieval in \bench{} can be easier and more stable than in deployed tool ecosystems. 
However, this simplification is intentional.
Our goal is not to benchmark retrieval systems themselves, but to isolate agents' ability to explore, plan, and recover under partial tool observability. 
By grounding retrieval in datatypes, we can systematically control which tools are discoverable, which paths remain feasible, and how blocking events affect the solution space. 
This enables reproducible evaluation, fair model comparison, and fine-grained analysis of exploration and re-planning failures, while leaving more realistic retriever noise as an important direction for future extensions.

\paragraph{Task Complexity and Diversity.} 
\bench{} is designed to contain queries that are both complex and diverse. 
The \textbf{\textit{complexity}} of each query is controlled by the underlying state graph rather than by surface-level linguistic difficulty.
We only retain tasks whose shortest valid solution requires at least five distinct tool calls, ensuring that each query requires non-trivial multi-step reasoning and cannot be solved by a single direct lookup. 
Empirically, these tasks also require long interactions in the easiest default setting, with agents taking around 25 turns on average in the easiest setting, which further shows that the benchmark evaluates long-horizon planning rather than isolated tool invocation.
The \textbf{\textit{diversity}} of \bench{} comes from its datatype-grounded construction process. 
Instead of manually writing queries or relying on paraphrase variation, we first define a broad set of domain-specific datatypes that represent distinct kinds of retail information. 
Tools are then generated by pairing different input and output datatype sets. 
As a result, each query is grounded in a different transformation path over these datatypes. 
This makes the query distribution diverse at the semantic and structural levels.

\section{Discussion}
\label{app:discussion}

\subsection{Beyond Simple Search Problem}
\label{app:why-is-not-a-search-problem}

The enforced-exploration results presented in Section~\ref{sec:enforce-analysis} raise a natural question: whether the block setting can be reduced to a simple search problem, and whether the performance drop can be resolved by allocating more test-time interaction.
We argue that this is not the case. 
The main difficulty does not only come from insufficient search over an explicit state space. 
Instead, agents must infer latent intermediate goals, translate them into natural-language retrieval requests, interpret imperfect tool feedback, and revise their plans under partial observability.

\paragraph{Not a rule-based graph search problem.}
Although our benchmark construction relies on an underlying typed state graph (introduced in Appendix~\ref{app:state-graph}), this graph is not exposed to the agent during evaluation. 
The graph is defined over internal datatypes, which are used to construct tools, compute valid solution paths, and verify whether the target information has been reached. 
However, the agent never observes these datatypes or the graph structure. 
It can only interact with the environment by issuing \textit{\textbf{natural-language retrieval requests}} and interpreting the returned tool descriptions. 
The environment then maps these requests to the internal datatype space. 
Therefore, a rule-based graph search algorithm could solve the task only in an oracle setting where the typed graph is directly available. 
This differs from the actual evaluation setting, where the agent must recover useful tool paths from partial and natural-language observations.

\paragraph{Not solved by additional test-time exploration.}
We further examine whether the block-setting degradation is mainly caused by a limited interaction budget.
Inspired by test-time scaling methods that force models to continue reasoning before termination~\citep{s1}, we design an enforced-exploration analysis for our multi-turn setting.
Specifically, whenever an agent attempts to terminate with an incorrect answer, we insert an additional user message asking it to keep exploring the tool space and environment.
This intervention can be triggered multiple times within the same trajectory, up to the enforced budget used in the experiment.
It is used only for analysis, not as a deployable method, because it relies on ground-truth evaluation to determine whether the attempted final answer is incorrect.
The results show that additional interaction alone brings only limited gains.
Even with repeated continuation prompts, most models improve by less than 5 percentage points and remain far below their no-block accuracy.
This suggests that the main bottleneck is not simply the number of available interaction steps, but the agent's ability to diagnose failed or misleading tool paths, abandon an invalid plan,and construct a new recovery plan.
Thus, retrieval-time blocking evaluates adaptive re-planning under partial observability, rather than simple search over a known state space.

\subsection{Evaluation Fairness}
\label{app:evaluation-fairness}

To ensure fair comparison across models, \bench{} applies the same blocking configuration to every model within each blocking setting. 
Specifically, for a given evaluation setting, the set of tools selected for blocking is pre-computed for every task instance and then kept fixed across all evaluated models. 
Thus, all models are evaluated on the same benchmark database, with the same blocked tools, the same replacement tools, and the same perturbation type under each setting.

This design ensures that the induced changes to the solution space are identical across models. 
Since the blocked tools are the same for every model, the solution paths removed by blocking are also the same. 
Conversely, the remaining feasible paths available after blocking are shared across models as well. 
Therefore, any performance differences observed under a blocking setting can be attributed to the models’ ability to detect failures, avoid misleading tools, and adaptively recover through the remaining valid paths, rather than to differences in the blocking conditions they face.

\subsection{Future Work}
\label{app:potential-improvement-methods}

\paragraph{Diversity-aware tool exploration.}
To address the insufficient-exploration problem that our benchmark pinpoints, future agents could be trained to retrieve tools with explicit diversity objectives. 
Instead of repeatedly searching around the same local tool neighborhood, the agent should generate complementary retrieval queries under different views: forward from currently known evidence, backward from the desired target, and bridge queries that connect known and missing information. 
This direction is closely related to prior work on tool retrieval and large-scale API selection, which improves tool discovery through better retrieval, generation, or iterative feedback \citep{APIBench,ToolRet,xu2024enhancing,ToolGen}. 
In \bench{}, such methods could be optimized using exploration-oriented signals, such as increasing the breadth of useful explored datatypes while avoiding redundant searches.

\paragraph{Failure-aware tool verification.}
To mitigate the silent-failure problem highlighted by our blocking analysis, agents need mechanisms for validating whether a tool response is trustworthy before incorporating it into later steps. 
A concrete method is to add a verification phase after each suspicious tool call: the agent checks whether the output has a valid type, whether it is semantically plausible, whether it contradicts previous evidence, and whether an independent alternative tool can confirm it. 
This direction connects to prior work on imperfect, opaque, or evolving tools, where agents must learn tool behavior through interaction or adapt to changing APIs \citep{OpaqueToolsBench,ToolSandbox, acikgoz2026toolr0}. 
In \bench{}, this method is especially useful for implicit failure blocks, where the tool does not return an explicit error but instead produces a counterfactual value that can mislead later planning.

\paragraph{Backtracking-based recovery planning.}
To address the brittle re-planning problem exposed when direct paths are blocked, agents could be equipped with explicit backtracking and recovery policies. 
When a tool fails, returns an implausible result, or becomes inconsistent with the current plan, the agent should mark the corresponding path as unreliable, roll back to the most recent valid state, and search for an alternative path from either the current evidence or the final target. This is related to search-based and hierarchical planning methods, which decompose long-horizon tasks into smaller decision points and revise plans when execution diverges from expectation \citep{AnyTool,verma2025leap,koh2026tree}. In \bench{}, such a method would directly target the longest-path setting, where agents must recover through less direct chains of intermediate tools rather than simply extending the same failed trajectory.

\paragraph{Training with blocker-aware trajectories.}
To improve robustness under unreliable tool access, agents can be trained on trajectories that explicitly contain retrieval misses, explicit errors, implicit errors, and semantic distractions. Rather than only learning successful tool-call demonstrations, the model should learn recovery behaviors: recognizing a blocked path, avoiding repeated calls to failed tools, discarding suspicious observations, and searching for alternative tools. Reinforcement learning is a natural fit here, since \bench{} provides executable feedback and can define rewards for final accuracy, relevant execution, low invalid-call rate, and successful recovery after blockers. This is aligned with recent work showing that tool-use agents can be improved through interaction-based learning and reward-driven tool use \citep{schick2023toolformer,ToolRL}. Compared with simple test-time scaling, blocker-aware training directly teaches the missing adaptive behavior rather than merely giving the model more turns.

\subsection{Significance and Generalization}
\label{app:significance-generalization}

\paragraph{On the Significance of \bench{}.}

\bench{} introduces two key novelties for evaluating real-world tool-using agents. First, it explicitly evaluates agents' ability to explore implicit sub-goals in large-scale tool ecosystems. Unlike settings where the required tools or intermediate steps are given in advance, \bench{} requires agents to infer what information is missing, retrieve relevant tools, and construct a valid multi-step solution path. This capability is increasingly important in practical agent scenarios, where external capabilities such as online skills, plugins, and MCP servers are often only partially visible and must be discovered through retrieval before they can be used.

A central contribution of \bench{} is that it does not measure this ability only through final accuracy. 
Instead, it provides \textbf{\textit{dedicated metrics for both exploration and exploitation}}. 
Exploration-oriented metrics capture the breadth and diversity of intermediate information uncovered by the agent, while exploitation-oriented metrics assess whether the agent can use the discovered tools along task-relevant paths.
These metrics help disentangle which capabilities are associated with final task success, revealing whether strong performance depends more on discovering intermediate information, efficiently navigating the tool space, or accurately executing over the explored tools.

Second, \bench{} introduces \textit{\textbf{blocking mechanism}} to evaluate whether agents can adapt when originally useful tools become unavailable or unreliable. 
This reflects realistic large-scale tool environments: an online skill may be removed, an API may return an error, an MCP server may expose a misleadingly similar tool, or a retrieved capability may silently produce an unreliable result. 
In such cases, a robust agent should not simply follow the first plausible path it finds; it should detect the disruption, revise its plan, and recover through alternative tool-use paths.

Together, implicit sub-goal exploration and dynamic blocking make \bench{} a broadly applicable benchmark for studying adaptive planning in large, partially observable tool ecosystems. 
These two components capture challenges that are central to future agent deployments, where success depends not only on calling tools correctly, but also on discovering useful capabilities, exploiting them coherently, and recovering when the environment changes.

\paragraph{On the Generalization of \bench{}.}

Although \bench{} is instantiated in the retail domain, its core design is \textit{\textbf{domain-general}}. 
The benchmark can be adapted to other domains by replacing the domain-specific datatypes, tool library, and backend database while preserving the same exploration protocol and dynamic blocking mechanism. 

In particular, the \textit{\textbf{exploration setting}} applies to any task where a user provides initial information and specifies a desired outcome, but leaves the intermediate sub-goals and tool-use path implicit. 
This input-to-output structure appears in many multi-step tool-use domains, including travel planning, enterprise workflow automation, customer support, healthcare administration, finance, and software engineering. 
In such settings, agents must determine which intermediate information is needed and which tools can produce it, making implicit sub-goal exploration a domain-general challenge rather than a retail-specific one.

The \textit{\textbf{blocking mechanism}} is also broadly applicable. 
Across tool-use environments, relevant tools may become unavailable, return errors, expose outdated information, or be replaced by superficially similar but functionally different alternatives. 
Therefore, the ability to detect disrupted tool-use paths and recover through alternative paths is not specific to retail, but is a general requirement for robust multi-turn tool-use planning. 
For these reasons, \bench{} provides a general framework for studying exploration, exploitation, and adaptive re-planning in large-scale tool ecosystems, and its findings are expected to be informative for a broad range of multi-step tool-use settings.

\section{Human Annotation}
\label{app:human-annotation}

To validate the quality of the constructed tool library and datatype inventory, we conducted a human annotation study with five annotators, all of whom had relevant research experience. 
Each annotator evaluated 10 tools and 5 datatypes, resulting in 50 annotated tools and 25 annotated datatypes in total. 
Following the standard Likert-scale convention~\citep{likert1932technique}, annotators rated each item from 1 to 5 based on its reasonableness and realism (screenshots provided in Figure~\ref{fig:annotation_1} and~\ref{fig:annotation_2}). The annotated tools received an average score of 4.32, and the annotated datatypes received an average score of 4.56. These results indicate that both the generated tools and datatypes are of high quality and align well with realistic retail-domain scenarios.

\begin{figure*}[htbp]
\begin{tcolorbox}[
    colback=black!5!white,
    colframe=black,
    title=Error Case: Irrecoverable Drift with Wrong Tool Value,
    fonttitle=\bfseries,
    colbacktitle=black!20!white,
    coltitle=black,
    boxrule=1.5pt,
    arc=5pt,
    boxsep=5pt,
    left=12pt,
    right=12pt,
    top=12pt,
    bottom=12pt
]
\begin{lstlisting}[basicstyle=\ttfamily\footnotesize,breaklines=true,columns=fullflexible,keepspaces=true]
[Model / setting]
DeepSeek-V4-Flash, Default

[Raw query]
Finance is reconciling a commerce-side transaction and one linked payment field is still missing.
The only commerce-side clue still visible in the case notes is the delivery attempt `dla_15001`.
They want the processor-side value that corresponds to the same case.
Can you identify and return the corresponding payment method ID?

[Observed trajectory excerpt]
3. Get_Outbound_Shipment_ID_From_Attempt_Record_ID_V2("dla_15001")
   -> shipment_id = "shp_14001"

5. Get_Placed_Order_ID_From_Shipment_Reference_Quick("shp_14001")
   -> order_id = "ord_7001"

6. retrieve_tools(by_io_info = {"output": "payment method id", "input": ["order"]})
   -> tool_count = 0

11. Get_Order_Draft_From_Live_Order_ID_Instant("ord_7001")
    -> order_draft_id = "odrft_4001"

15. Get_Default_Method_ID_From_Order_Draft_Reference_Turbo("odrft_4001")
    -> default_payment_method_id = "pm_9001"

16. final_answer
    -> pm_9001

[Expected vs. observed]
expected_answer = "pm_9001"
target_datatype = "payment_method_id"
observed_datatype = "default_payment_method_id"
target_datatype_reached = false
failure_reason_detail = "target_datatype_not_reached"
\end{lstlisting}
\end{tcolorbox}
\caption{A representative default-setting failure of \emph{Irrecoverable Drift} with a wrong tool-value ending. The model obtains useful intermediate evidence, then follows a non-progress branch to \textit{default\_payment\_method\_id} and returns that off-path value instead of grounding the requested \textit{payment\_method\_id}.}
\label{fig:case_irrecoverable_drift_wrong_tool_value}
\end{figure*}

\begin{figure*}[htbp]
\begin{tcolorbox}[
    colback=black!5!white,
    colframe=black,
    title=Error Case: Irrecoverable Drift with Search Exhaustion,
    fonttitle=\bfseries,
    colbacktitle=black!20!white,
    coltitle=black,
    boxrule=1.5pt,
    arc=5pt,
    boxsep=5pt,
    left=12pt,
    right=12pt,
    top=12pt,
    bottom=12pt
]
\begin{lstlisting}[basicstyle=\ttfamily\footnotesize,breaklines=true,columns=fullflexible,keepspaces=true]
[Model / setting]
Gemini-3.5-Flash, Default

[Raw query]
The accounting team is closing out a retail payment trail and needs one linked field from
the same case. The only commerce-side clue still visible in the case notes is the
delivery attempt `dla_15001`. The missing payment detail is blocking the audit trail
for this purchase. Can you identify and return the corresponding authorization code?

[Observed trajectory excerpt]
1. retrieve_tools(by_inputs = ["delivery attempt"])
   -> tool_count = 30

2. Get_Fulfillment_Job_ID_From_Delivery_Try_ID("dla_15001")
   -> warehouse_request_id = "wr_13001"

4. Get_Live_Order_ID_From_Warehouse_Request_Max("wr_13001")
   -> order_id = "ord_7001"

6. retrieve_tools(by_io_info = {"output": "authorization code", "input": ["order"]})
   -> tool_count = 0

...

80. retrieve_tools(by_inputs = ["payment intent"])
    -> tool_count = 30

98. retrieve_tools(by_inputs = ["order", "invoice"])
    -> tool_count = 0

99. retrieve_tools(by_inputs = ["order", "receipt"])
    -> tool_count = 0

100. retrieve_tools(by_inputs = ["order", "billing"])
     -> tool_count = 0

[Terminal outcome]
final_answer = null
failure_reason_detail = "exceeded_max_steps"

[Expected vs. observed]
expected_answer = "AUTH-AX91"
target_datatype = "auth_code"
target_datatype_reached = false
\end{lstlisting}
\end{tcolorbox}
\caption{A representative default-setting failure where partial progress turns into drift and then \emph{Search Exhaustion}. The model obtains useful early datatypes, but then spends almost the entire remaining budget on retrieval without grounding \textit{auth\_code}.}
\label{fig:case_irrecoverable_drift_search_exhaustion}
\end{figure*}

\begin{figure*}[htbp]
\begin{tcolorbox}[
    colback=black!5!white,
    colframe=black,
    title=Error Case: Silent Failure Output Accepted,
    fonttitle=\bfseries,
    colbacktitle=black!20!white,
    coltitle=black,
    boxrule=1.5pt,
    arc=5pt,
    boxsep=5pt,
    left=12pt,
    right=12pt,
    top=12pt,
    bottom=12pt
]
\begin{lstlisting}[basicstyle=\ttfamily\footnotesize,breaklines=true,columns=fullflexible,keepspaces=true]
[Model / setting]
Gemini-3.5-Flash, Blocker

[Raw query]
A payments analyst is tracing the downstream charge record for a retail purchase.
The only concrete reference left in the ticket is the order ID `ord_7001`.
The missing payment detail is blocking the audit trail for this purchase.
Can you identify and return the corresponding funding account reference?

[Observed trajectory excerpt]
...
20. Get_Return_Request_From_Order_Enterprise("ord_7001")
    -> return_request_id = "rrq_16001"

22. Get_Finance_Refund_ID_From_Review_Ticket_ID_Nano("raud_17001")
    -> refund_id = "ref_18001"

24. Get_Auth_ID_From_Payment_Record_ID_Core("pay_12001")
    -> auth_id = "auth_11001"

26. Get_Payment_Intent_From_Auth_Reference_Quick("auth_11001")
    -> payment_intent_id = "-2147483648"
    -> tool_type = "blocker_misleading"

27. Get_Checkout_Intent_ID_From_Charge_ID("pay_12001")
    -> payment_intent_id = "pi_10001"

29. Get_Account_From_Payment_Intent_V3_Preview("pi_10001")
    -> output_datatype = "account_number"
    -> output_value = "acct_tok_5212"
    -> tool_type = "noisy_misleading"

30. final_answer
    -> acct_tok_5212

[Expected vs. observed]
expected_answer = "acct_tok_4242"
target_datatype_reached = false
failure_reason_detail = "target_datatype_not_reached"
\end{lstlisting}
\end{tcolorbox}
\caption{A representative blocked-setting failure in which a corrupted executable-looking output is treated as grounded evidence. The model recovers a valid \textit{payment\_intent\_id}, but then accepts a noisy account-number value and returns it as the final answer.}
\label{fig:case_silent_failure_output_accepted}
\end{figure*}

\begin{figure*}[htbp]
\begin{tcolorbox}[
    colback=black!5!white,
    colframe=black,
    title=Error Case: Format Error,
    fonttitle=\bfseries,
    colbacktitle=black!20!white,
    coltitle=black,
    boxrule=1.5pt,
    arc=5pt,
    boxsep=5pt,
    left=12pt,
    right=12pt,
    top=12pt,
    bottom=12pt
]
\begin{lstlisting}[basicstyle=\ttfamily\footnotesize,breaklines=true,columns=fullflexible,keepspaces=true]
[Model / setting]
GPT-5.4, Default

[Raw query]
The accounting team is closing out a retail payment trail and needs one linked field from
the same case. The only commerce-side clue still visible in the case notes is the draft
item `odritm_5001`. They need the finance-side answer before the reconciliation ticket
can be closed. Can you identify and return the corresponding payment method type?

[Observed trajectory excerpt]
5. Get_Order_Item_Reference_From_Cart_Item_Reference_Fast("odritm_5001")
   -> order_item_id = "orditm_8001"

8. Get_Order_Reference_From_Placed_Order_Item_ID_Enterprise("orditm_8001")
   -> order_id = "ord_7001"

15. Get_Quote_Record_ID_From_Order_Reference_Fast("ord_7001")
    -> pricing_snapshot_id = "psnap_6001"

19. Get_Account_ID_From_Draft_Line_ID("odritm_5001")
    -> user_id = "usr_1001"

21. Get_Saved_Method_ID_From_User_Record_ID_Max("usr_1001")
    -> default_payment_method_id = "pm_9001"

22. Get_Payment_Type_From_Payment_Method_Reference("pm_9001")
    -> rejected: missing required datatype ["payment_method_id"]

24. Get_Payment_Type_From_Payment_Method_Reference("pm_9001")
    -> rejected: missing required datatype ["payment_method_id"]

25. Get_Account_ID_From_Saved_Payment_Method_ID("pm_9001")
    -> rejected: missing required datatype ["payment_method_id"]

[Terminal outcome]
final_answer = null
failure_reason_detail = "exceeded_max_tool_call_errors"

[Expected vs. observed]
expected_answer = "credit_card"
target_datatype_reached = false
\end{lstlisting}
\end{tcolorbox}
\caption{A representative \emph{Format Error}. The model makes partial progress, but then repeatedly calls tools with values whose held datatype does not satisfy the required input datatype, producing invalid tool calls until the run terminates.}
\label{fig:case_interaction_breakdown}
\end{figure*}

\begin{figure*}[htbp]
\begin{tcolorbox}[
    colback=black!5!white,
    colframe=black,
    title=Error Case: Premature Answering,
    fonttitle=\bfseries,
    colbacktitle=black!20!white,
    coltitle=black,
    boxrule=1.5pt,
    arc=5pt,
    boxsep=5pt,
    left=12pt,
    right=12pt,
    top=12pt,
    bottom=12pt
]
\begin{lstlisting}[basicstyle=\ttfamily\footnotesize,breaklines=true,columns=fullflexible,keepspaces=true]
[Model / setting]
Llama-3.3-70B-Instruct, Default

[Raw query]
Finance is reconciling a commerce-side transaction and one linked payment field is still 
missing.
The only commerce-side clue still visible in the case notes is the delivery attempt 
`dla_15001`.
They need the finance-side answer before the reconciliation ticket can be closed.
What authorization status is recorded for that case right now?

[Observed trajectory excerpt]
1. Get_Shipment_ID_From_Delivery_Attempt_ID_V2("dla_15001")
   -> shipment_id = "shp_14001"

2. Get_Order_ID_From_Shipment_ID_Quick("shp_14001")
   -> order_id = "ord_7001"

3. Get_Order_Status_From_Order_ID_Nano("ord_7001")
   -> order_status = "refunded"

4. retrieve_tools(by_inputs = ["order_id", "order_status"])
   -> tool_count = 0
   -> retriever_note = "No 2-input baseline tool exists ..."

5. final_answer
   -> {"answer": "refunded"}

[Expected vs. observed]
expected_answer = "approved"
target_datatype_reached = false
failure_reason_detail = "target_datatype_not_reached"
\end{lstlisting}
\end{tcolorbox}
\caption{A representative default-setting failure of \emph{Premature Answering}. The model follows a valid prefix of the path, but after observing \textit{order\_status = refunded}, it answers with a semantically adjacent status instead of continuing toward \textit{auth\_status}.}
\label{fig:case_default_premature_answer}
\end{figure*}

\begin{figure*}[htbp]
\begin{tcolorbox}[
    colback=black!5!white,
    colframe=black,
    title=Error Case: Unresolved Execution,
    fonttitle=\bfseries,
    colbacktitle=black!20!white,
    coltitle=black,
    boxrule=1.5pt,
    arc=5pt,
    boxsep=5pt,
    left=12pt,
    right=12pt,
    top=12pt,
    bottom=12pt
]
\begin{lstlisting}[basicstyle=\ttfamily\footnotesize,breaklines=true,columns=fullflexible,keepspaces=true]
[Model / setting]
DeepSeek-V4-Flash, Default

[Raw query]
A payments analyst is tracing the downstream charge record for a retail purchase.
The team can confirm the order draft `odrft_4001`, and the same case also points to
the purchased variant `var_3001`. They want the processor-side value that
corresponds to the same case. Please trace the linked records and return the
payment intent ID.

[Observed trajectory excerpt]
4. retrieve_tools(by_inputs = ["order_draft_id", "variant_id"])
   -> tool_count = 1

7. Get_Order_Draft_Item_ID_From_Order_Draft_ID_And_Variant_ID("odrft_4001", "var_3001")
   -> order_draft_item_id = "odritm_5001"

9. Get_Order_Item_ID_From_Order_Draft_Item_ID_Fast("odritm_5001")
   -> order_item_id = "orditm_8001"

10. Get_Order_ID_From_Order_Item_ID_Enterprise("orditm_8001")
    -> order_id = "ord_7001"

...

24. Get_Pricing_Snapshot_ID_From_Order_ID_Fast("ord_7001")
    -> pricing_snapshot_id = "psnap_6001"

25. retrieve_tools(by_inputs = ["payment_intent_id"])
    -> tool_count = 10

28. Get_Return_Request_ID_From_Order_ID_Enterprise("ord_7001")
    -> return_request_id = "rrq_16001"

29. Get_Return_Audit_ID_From_Return_Request_ID_Edge("rrq_16001")
    -> return_audit_id = "raud_17001"

30. Get_Refund_ID_From_Return_Audit_ID_Nano("raud_17001")
    -> refund_id = "ref_18001"

[Terminal outcome]
final_answer = null
failure_reason_detail = "exceeded_max_steps"

[Expected vs. observed]
expected_answer = "pi_10001"
target_datatype = "payment_intent_id"
target_datatype_reached = false
\end{lstlisting}
\end{tcolorbox}
\caption{A representative default-setting failure of \emph{Unresolved Execution}. The model follows a correct early prefix of the path and continues executing valid actions, but it spends too much of the budget on additional retrieval and never commits to a final answer.}
\label{fig:case_default_unresolved_execution}
\end{figure*}

\begin{figure*}[htbp]
\begin{tcolorbox}[
    colback=black!5!white,
    colframe=black,
    title=Error Case: Noisy Response Adoption,
    fonttitle=\bfseries,
    colbacktitle=black!20!white,
    coltitle=black,
    boxrule=1.5pt,
    arc=5pt,
    boxsep=5pt,
    left=12pt,
    right=12pt,
    top=12pt,
    bottom=12pt
]
\begin{lstlisting}[basicstyle=\ttfamily\footnotesize,breaklines=true,columns=fullflexible,keepspaces=true]
[Model / setting]
GPT-5.4, Blocked

[Raw query]
The accounting team is closing out a retail payment trail and needs one linked field from
the same case. The only concrete reference left in the ticket is the delivery attempt
`dla_15001`. They need the finance-side answer before the reconciliation ticket can be
closed. What is the exact funding account reference associated with that purchase?

[Observed trajectory excerpt]
...
26. Get_Auth_ID_From_Refund_ID_Prime("ref_18001")
    -> auth_id = "auth_11001"

27. Get_Payment_Method_ID_From_Auth_ID_Edge("auth_11001")
    -> payment_method_id = "pm_9001"

28. Get_Account_Number_From_Payment_Method_ID("pm_9001")
    -> output_value = "error: endpoint temporarily unavailable."

29. Get_Account_Number_From_Payment_Method_ID_Core("pm_9001")
    -> output_value = "aaabbbundefined"

30. final_answer
    -> aaabbbundefined

[Expected vs. observed]
expected_answer = "acct_tok_4242"
target_datatype_reached = true
failure_reason_detail = "final_answer_wrong"
\end{lstlisting}
\end{tcolorbox}
\caption{A representative blocked failure of \emph{Noisy Response Adoption}. The model does reach the target datatype \textit{account\_number}, but then accepts a corrupted target-typed value and returns it as the final answer.}
\label{fig:case_blocked_wrong_after_target}
\end{figure*}

\begin{figure*}[htbp]
\begin{tcolorbox}[
    colback=black!5!white,
    colframe=black,
    title=Error Case: Format Error,
    fonttitle=\bfseries,
    colbacktitle=black!20!white,
    coltitle=black,
    boxrule=1.5pt,
    arc=5pt,
    boxsep=5pt,
    left=12pt,
    right=12pt,
    top=12pt,
    bottom=12pt
]
\begin{lstlisting}[basicstyle=\ttfamily\footnotesize,breaklines=true,columns=fullflexible,keepspaces=true]
[Model / setting]
Qwen3-32B, Blocked

[Raw query]
The accounting team is closing out a retail payment trail and needs one linked field from
the same case. The only concrete reference left in the ticket is the delivery attempt
`dla_15001`. They need the finance-side answer before the reconciliation ticket can be
closed. What is the exact funding account reference associated with that purchase?

[Observed trajectory excerpt]
1. Get_Shipment_ID_From_Delivery_Attempt_ID("dla_15001")
   -> shipment_id = "error"

4. Get_Warehouse_Request_ID_From_Delivery_Attempt_ID("dla_15001")
   -> warehouse_request_id = "wr_13001"

6. Get_Order_ID_From_Warehouse_Request_ID_Max("wr_13001")
   -> order_id = "ord_7001"

12. Get_Default_Payment_Method_ID_From_User_ID_Max("usr_1001")
    -> default_payment_method_id = "pm_9001"

15. Get_Account_Number_From_Payment_Method_ID("pm_9001")
    -> rejected: missing required datatype ["payment_method_id"]

18. Get_Default_Payment_Method_ID_From_Payment_Method_ID_Prime("pm_9001")
    -> rejected: missing required datatype ["payment_method_id"]

19. Get_Account_Number_From_Payment_Method_ID("pm_9001")
    -> rejected again: missing required datatype ["payment_method_id"]

[Terminal outcome]
final_answer = null
failure_reason_detail = "exceeded_max_tool_call_errors"
\end{lstlisting}
\end{tcolorbox}
\caption{A representative blocked failure of \emph{Format Error}. The model makes local progress, but then repeatedly treats \textit{default\_payment\_method\_id} as if it were \textit{payment\_method\_id}, causing an invalid-call loop and eventual termination without a grounded answer.}
\label{fig:case_blocked_collapse}
\end{figure*}

\begin{figure*}[htbp]
\begin{tcolorbox}[
    colback=black!5!white,
    colframe=black,
    title=Data Case: Representative Query and Task Profile,
    fonttitle=\bfseries,
    colbacktitle=black!20!white,
    coltitle=black,
    boxrule=1.5pt,
    arc=5pt,
    boxsep=5pt,
    left=12pt,
    right=12pt,
    top=12pt,
    bottom=12pt
]
\begin{lstlisting}[basicstyle=\ttfamily\footnotesize,breaklines=true,columns=fullflexible,keepspaces=true]
[Natural-language query]
The accounting team is closing out a retail payment trail and needs one linked field from
the same case. The only concrete reference left in the ticket is the delivery attempt
`dla_15001`. They need the finance-side answer before the reconciliation ticket can be
closed. What is the exact funding account reference associated with that purchase?

[Underlying task record: retail_task_0001]
{
  "input_datatypes": ["delivery_attempt_id"],
  "num_inputs": 1,
  "target_datatype": "account_number",
  "steps": 8,
  "one_available_path": [
    "get_outbound_shipment_id_from_attempt_record_id_v2",
    "get_placed_order_id_from_shipment_reference_quick",
    "get_return_request_from_order_enterprise",
    "get_internal_return_review_id_from_return_ticket_id_edge",
    "get_finance_refund_id_from_review_ticket_id_nano",
    "get_auth_id_from_finance_refund_id_prime",
    "get_payment_setup_id_from_authorization_record_id_core",
    "get_account_from_payment_intent_v3"
  ]
}
\end{lstlisting}
\end{tcolorbox}
\caption{A representative retail query and its task profile. The same task is expressed both as a natural-language user query (shown to agents) and as a typed planning problem with explicit input datatypes, target datatype, and one available solution path (used only for benchmark construction and evaluation).}
\label{fig:case_data_query_profile}
\end{figure*}

\clearpage
\onecolumn

\begin{center}
\begin{tcolorbox}[
    breakable,
    colback=black!5!white,
    colframe=black,
    title=Data Case: Representative Tool Definitions,
    fonttitle=\bfseries,
    colbacktitle=black!20!white,
    coltitle=black,
    boxrule=1.5pt,
    arc=5pt,
    boxsep=5pt,
    left=12pt,
    right=12pt,
    top=12pt,
    bottom=12pt
]
\begin{lstlisting}[basicstyle=\ttfamily\footnotesize,breaklines=true,columns=fullflexible,keepspaces=true]
[Tool 1]
### agent_visible
{
  "type": "function",
  "name": "get_outbound_shipment_id_from_attempt_record_id_v2",
  "description": "Given `delivery attempt id`, this returns the shipment that this
  specific delivery attempt belongs to. Use it when you need the broader parcel context
  behind one attempt event.",
  "parameters": {
    "type": "object",
    "properties": {
      "delivery_attempt_id": {
        "type": "string",
        "description": "Unique ID of a specific delivery attempt for a shipment. One
        shipment can have multiple delivery attempts, so this is narrower than shipment_id."
      }
    },
    "required": ["delivery_attempt_id"],
    "additionalProperties": false
  },
  "strict": true
}

### internal_metadata
{
  "input_datatypes": ["delivery_attempt_id"],
  "output_datatype": "shipment_id"
}

[Tool 2]
### agent_visible
{
  "type": "function",
  "name": "get_return_request_from_order_enterprise",
  "description": "Given `completed order record id`, this returns the customer-facing
  return request created for that order, if one exists. Use it when you need to move from
  the purchase record into the return workflow.",
  "parameters": {
    "type": "object",
    "properties": {
      "completed_order_record_id": {
        "type": "string",
        "description": "Unique ID of the final placed order. Unlike order_draft_id, this
        refers to the real order after checkout."
      }
    },
    "required": ["completed_order_record_id"],
    "additionalProperties": false
  },
  "strict": true
}

### internal_metadata
{
  "input_datatypes": ["order_id"],
  "output_datatype": "return_request_id"
}

[Tool 3]
### agent_visible
{
  "type": "function",
  "name": "get_account_from_payment_intent_v3",
  "description": "Given `payment initiation id`, this returns the recognizable account
  or card number tied to that payment attempt, usually in masked form. Use it when you need
  a human-readable identifier for the instrument that was used.",
  "parameters": {
    "type": "object",
    "properties": {
      "payment_initiation_id": {
        "type": "string",
        "description": "Unique ID of the payment intent for an order. This represents
        the stage where the system is preparing to start a payment, not the final payment
        result."
      }
    },
    "required": ["payment_initiation_id"],
    "additionalProperties": false
  },
  "strict": true
}

### internal_metadata
{
  "input_datatypes": ["payment_intent_id"],
  "output_datatype": "account_number"
}
\end{lstlisting}


\end{tcolorbox}
\end{center}

\refstepcounter{figure}\label{fig:case_data_tools}
\smallskip
\noindent\parbox{\textwidth}{\raggedright
\textbf{Figure~\thefigure:} Representative tool definitions from the same task family. The \texttt{agent\_visible} section shows the function schema exposed to the model, while \texttt{internal\_metadata} is used only by the benchmark implementation. Together, these tools span fulfillment, after-sales, and payment datatypes, illustrating why successful planning often requires cross-subflow composition rather than single-hop lookup.
}

\clearpage
\onecolumn

\begin{center}
\begin{tcolorbox}[
    breakable,
    colback=black!5!white,
    colframe=black,
    title=Inference Prompt,
    fonttitle=\bfseries,
    colbacktitle=black!20!white,
    coltitle=black,
    boxrule=1.5pt,
    arc=5pt,
    boxsep=5pt,
    left=12pt,
    right=12pt,
    top=12pt,
    bottom=12pt
]
\begin{lstlisting}[basicstyle=\ttfamily\footnotesize,breaklines=true,columns=fullflexible,keepspaces=true]

[System message]

[System Prompt]

You are an agent that solves user queries by planning over a tool ecosystem under a strict step budget. Your goal is to obtain the required information via tool interactions and produce a valid final answer.

---

[Environment]

- Information is only accessible via tools.
- Tools take inputs and produce outputs, and can be chained.
- The tool graph may be incomplete: direct or intuitive paths may not exist, but a valid solution path is guaranteed.

Domain context:
You are working in the retail domain.

This is a high-level overview of how records are modeled in this benchmark so you can plan tool search and tool use more effectively.

It is only background knowledge. It does not replace the actual tool definitions you retrieve.

At a high level, this retail setup contains these kinds of records:

- customer accounts
- products and product variants
- in-progress carts or draft orders
- finalized orders and order line items
- saved payment methods, payment attempts, authorizations, and completed payments
- warehouse or fulfillment requests, shipments, and delivery attempts
- return requests, return reviews, and refunds

The main business flow usually looks like this:

- a customer account leads to a draft order
- a draft order contains draft line items
- a draft order can produce a pricing result or quote
- that quote can turn into a finalized order
- a finalized order contains finalized order items

Payment usually follows a separate but connected flow:

- a saved payment method is used to start a payment attempt
- that payment attempt may go through an authorization step
- if successful, it becomes a completed payment

Fulfillment usually follows this flow:

- an order can create one or more warehouse or fulfillment requests
- each warehouse request produces one shipment in this benchmark setup
- a shipment can have one or more delivery attempts

After-sales usually follows this flow:

- a customer may create a return request for an order
- that return request may go through an internal review step
- if approved, it may lead to a refund

Some important modeling distinctions:

- draft-stage records are different from finalized order records
- a product is different from a specific product variant or SKU
- a payment attempt is different from an authorization result, and both are different from the final payment record
- a return request is different from the internal review of that request, and both are different from the final refund record
- a fulfillment request is different from a shipment, and a shipment is different from one delivery attempt

Useful planning intuition:

- many tasks start from one identifier and ask for information stored on a related record
- if you know which stage of the business process you are in, it is usually easier to search for the right tools
- if a task starts from cart or quote information, think in terms of the pre-checkout flow
- if a task starts from payment information, think in terms of the payment flow or after-sales flow
- if a task starts from shipping information, think in terms of the fulfillment flow
- if a task starts from return or refund information, think in terms of the after-sales flow

Business rules in this benchmark:

- returns are whole-order only
- each order uses exactly one payment account or payment method
- a single warehouse request is not split into multiple shipments

Use this overview as a guide for planning, but rely on retrieved tool definitions to determine what each tool actually does.

When planning, prefer transitions that follow this business flow instead of jumping to loosely related datatypes.
Stay within the retail domain unless the query explicitly requires another domain.
domain='retail'.

---

[Actions]

At each step, take EXACTLY ONE action:

1. <retrieve_tools> - discover tools
(already satisfied at query start in all-tools mode; use it only if you want the full inventory repeated)
2. <tool_call> - call a discovered tool
3. <final_answer> - return the final answer

Use exactly one tag per response.

---

[Action Usage]

- retrieve_tools
Use when you need tools to obtain required information.

<retrieve_tools>
{
  "meta_tool": "..."
}
</retrieve_tools>


- tool_call
Use when:
- The tool was retrieved in this query
- All required inputs are available

<tool_call>
{
  "tool_name": "...",
  "arguments": { ... }
}
</tool_call>


- final_answer
Use ONLY when the required information has been correctly obtained.

<final_answer>
...
</final_answer>

---

[Planning]

At each step, decide based on:

1. What information you currently have
2. What information is required for the final answer
3. What step will bring you closer to obtaining it
4. Which tool best helps produce that information

---

[Execution Strategy]

- Prefer making progress via tool calls once relevant tools are identified.
- Do not over-retrieve when actionable tools are available.
- If progress stalls, revise the plan and explore alternative compositions, instead of declaring the task unsolvable or guessing the answer.

---

[Tool Selection]

- Tool suffixes (v2, lite, pro, turbo, etc.) are not reliable signals of quality.
- The transformation defined by the tool's inputs, outputs, and description is meaningful.

If one tool does not lead to progress, consider alternative variants.

---

[Correctness and Validation]

Before using <final_answer>:

- The required information MUST come from a successful tool call in this query
- Similar-looking fields are NOT sufficient

Strictly:

- Do NOT guess or fabricate missing values
- If the required information was not produced by a tool, you are NOT done

---

[Budget]

You have at most 100 steps.

- Each action costs 1 step
- Avoid unproductive loops
- Balance retrieval and execution

---

[Rules]

You MUST:
- Retrieve before calling tools
- Only call tools retrieved in previous turns
- Ensure inputs are satisfied before calling

You MUST NOT:
- Mix multiple actions in one response
- Output outside the required tags
- Produce plausible but unverified outputs

---

[Mindset]

You are a constrained planner:

- Each step should reduce the gap between what you have and what is required
- When tools are available, act instead of over-searching
- If a path fails, adapt - solutions may require indirect reasoning

---

Begin.

[Initial user message template]

[query_text]

\end{lstlisting}
\end{tcolorbox}
\captionof{figure}{Full inference prompt used in our experiments.
}
\label{fig:inference-prompt}
\end{center}

\begin{figure*}[htbp]
\begin{tcolorbox}[
    colback=black!5!white,
    colframe=black,
    title=Enforce Exploration Prompt,
    fonttitle=\bfseries,
    colbacktitle=black!20!white,
    coltitle=black,
    boxrule=1.5pt,
    arc=5pt,
    boxsep=5pt,
    left=12pt,
    right=12pt,
    top=12pt,
    bottom=12pt
]
\begin{lstlisting}[basicstyle=\ttfamily\footnotesize,breaklines=true,columns=fullflexible,keepspaces=true]
Environment feedback:
The final answer you just gave does not appear to be correct, and you have not obtained the target information yet. Please continue exploring if needed before giving another final answer.
\end{lstlisting}
\end{tcolorbox}
\caption{Additional feedback prompt injected in the blocked+enforced setting when the agent outputs an incorrect final answer before reaching the target datatype. 
}
\label{fig:enforced-prompt}
\end{figure*}
\begin{figure*}[t]
  \centering
  \includegraphics[width=1\linewidth]{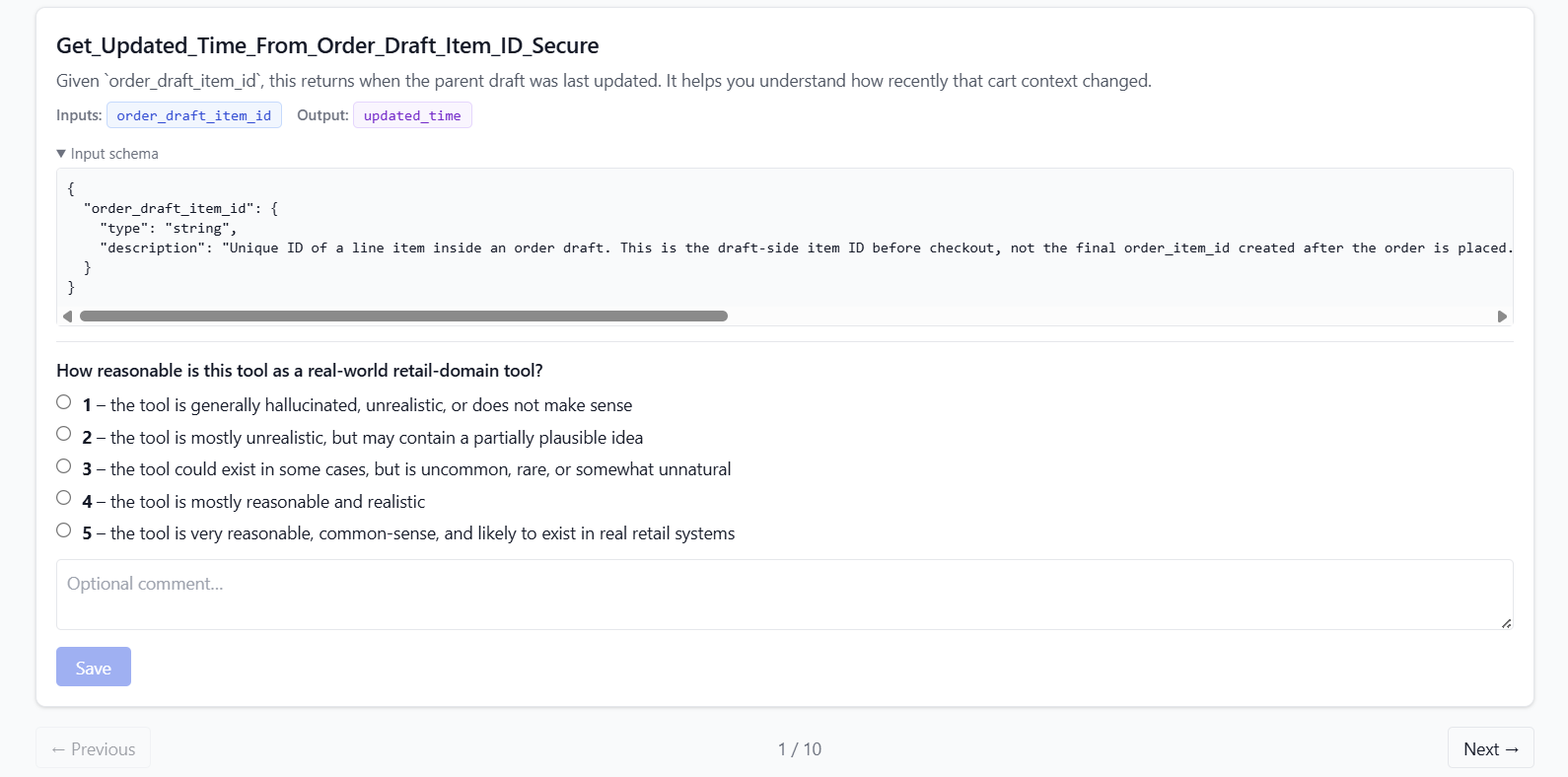}
  \caption{Annotation interface for evaluating tool-document quality. Annotators are shown a tool document and asked to rate how reasonable and usable the tool is on a 1--5 Likert scale.}
  \label{fig:annotation_1}
\end{figure*}
\begin{figure*}[t]
  \centering
  \includegraphics[width=1\linewidth]{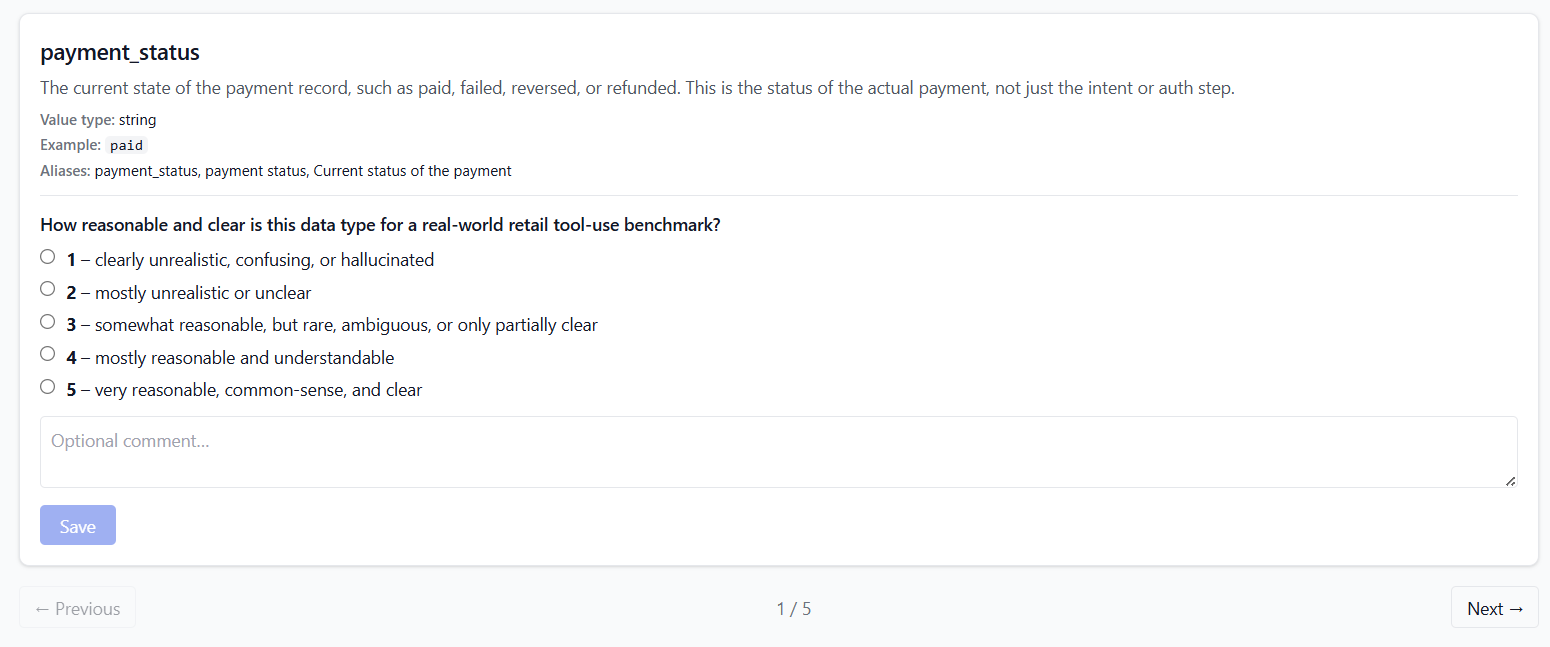}
  \caption{Annotation interface for evaluating datatype quality. Annotators are shown detailed definition of a datatype and asked to rate how reasonable the datatype is on a 1--5 Likert scale.}
  \label{fig:annotation_2}
\end{figure*}


\begin{thebibliography}{81}
\providecommand{\natexlab}[1]{#1}

\bibitem[{Acikgoz et~al.(2026)Acikgoz, Qian, H{\"u}botter, Ji, Hakkani-T{\"u}r, and Tur}]{acikgoz2026toolr0}
Emre~Can Acikgoz, Cheng Qian, Jonas H{\"u}botter, Heng Ji, Dilek Hakkani-T{\"u}r, and Gokhan Tur. 2026.
\newblock Tool-r0: Self-evolving llm agents for tool-learning from zero data.
\newblock \emph{arXiv preprint arXiv:2602.21320}.

\bibitem[{Basu et~al.(2024)Basu, Abdelaziz, Kate, Agarwal, Crouse, Rizk, Bradford, Munawar, Kumaravel, Goyal, Wang, Lastras, and Kapanipathi}]{NESTful}
Kinjal Basu, Ibrahim Abdelaziz, Kiran Kate, Mayank Agarwal, Maxwell Crouse, Yara Rizk, Kelsey Bradford, Asim Munawar, Sadhana Kumaravel, Saurabh Goyal, Xin Wang, Luis~A. Lastras, and Pavan Kapanipathi. 2024.
\newblock \href {https://arxiv.org/abs/2409.03797} {{NESTFUL}: A benchmark for evaluating {LLM}s on nested sequences of {API} calls}.
\newblock \emph{Preprint}, arXiv:2409.03797.

\bibitem[{Chen et~al.(2025{\natexlab{a}})Chen, Hao, Liu, Huang, Zeng, Yu, Li, Huang, Liu, Xinzhi, and Liu}]{ACEBench}
Chen Chen, Xinlong Hao, Weiwen Liu, Xu~Huang, Xingshan Zeng, Shuai Yu, Dexun Li, Yuefeng Huang, Xiangcheng Liu, Wang Xinzhi, and Wu~Liu. 2025{\natexlab{a}}.
\newblock \href {https://doi.org/10.18653/v1/2025.findings-emnlp.697} {{ACEB}ench: A comprehensive evaluation of {LLM} tool usage}.
\newblock In \emph{Findings of the Association for Computational Linguistics: EMNLP 2025}, pages 12970--12998, Suzhou, China. Association for Computational Linguistics.

\bibitem[{Chen et~al.(2025{\natexlab{b}})Chen, Zhang, Cong, Guo, Wu, Lin, Feng, and Wang}]{ToolQA-D}
Guoxin Chen, Zhong Zhang, Xin Cong, Fangda Guo, Yesai Wu, Yankai Lin, Wenzheng Feng, and Yasheng Wang. 2025{\natexlab{b}}.
\newblock \href {https://arxiv.org/abs/2410.06617} {Learning evolving tools for large language models}.
\newblock \emph{Preprint}, arXiv:2410.06617.

\bibitem[{Chen et~al.(2025{\natexlab{c}})Chen, Liang, and Wang}]{chen2025smurfs}
Junzhi Chen, Juhao Liang, and Benyou Wang. 2025{\natexlab{c}}.
\newblock Smurfs: Multi-agent system using context-efficient dfsdt for tool planning.
\newblock In \emph{Proceedings of the 2025 Conference of the Nations of the Americas Chapter of the Association for Computational Linguistics: Human Language Technologies (Volume 1: Long Papers)}, pages 3281--3298.

\bibitem[{Chen et~al.(2024)Chen, Su, Zuo, Yang, Yuan, Chan, Yu, Lu, Hung, Qian et~al.}]{chen2024agentverse}
Weize Chen, Yusheng Su, Jingwei Zuo, Cheng Yang, Chenfei Yuan, Chi-Min Chan, Heyang Yu, Yaxi Lu, Yi-Hsin Hung, Chen Qian, and 1 others. 2024.
\newblock Agentverse: Facilitating multi-agent collaboration and exploring emergent behaviors.
\newblock In \emph{International Conference on Learning Representations}, volume 2024, pages 20094--20136.

\bibitem[{{DeepSeek-AI}(2026)}]{DeepSeek-V4}
{DeepSeek-AI}. 2026.
\newblock {DeepSeek-V4: Towards Highly Efficient Million-Token Context Intelligence}.
\newblock \url{https://huggingface.co/deepseek-ai/DeepSeek-V4-Pro/blob/main/DeepSeek_V4.pdf}.

\bibitem[{Du et~al.(2024)Du, Wei, and Zhang}]{AnyTool}
Yu~Du, Fangyun Wei, and Hongyang Zhang. 2024.
\newblock Anytool: Self-reflective, hierarchical agents for large-scale api calls.
\newblock \emph{arXiv preprint arXiv:2402.04253}.

\bibitem[{Elder et~al.(2025)Elder, Murthi, Kang, Naik, Kate, Basu, and Contractor}]{LiveAPIBench}
Benjamin Elder, Anupama Murthi, Jungkoo Kang, Ankita~Rajaram Naik, Kiran Kate, Kinjal Basu, and Danish Contractor. 2025.
\newblock \href {https://arxiv.org/abs/2506.11266} {Live {API}-bench: 2500+ live {API}s for testing multi-step tool calling}.
\newblock \emph{Preprint}, arXiv:2506.11266.

\bibitem[{Erdogan et~al.(2025)Erdogan, Lee, Kim, Moon, Furuta, Anumanchipalli, Keutzer, and Gholami}]{erdogan2025planact}
Lutfi~Eren Erdogan, Nicholas Lee, Sehoon Kim, Suhong Moon, Hiroki Furuta, Gopala Anumanchipalli, Kurt Keutzer, and Amir Gholami. 2025.
\newblock \href {https://arxiv.org/abs/2503.09572} {Plan-and-act: Improving planning of agents for long-horizon tasks}.
\newblock \emph{Preprint}, arXiv:2503.09572.

\bibitem[{{Google}(2026)}]{Gemini-3.1-Pro}
{Google}. 2026.
\newblock Gemini 3.1 pro: A smarter model for your most complex tasks.
\newblock \url{https://blog.google/innovation-and-ai/models-and-research/gemini-models/gemini-3-1-pro/}.

\bibitem[{Grattafiori et~al.(2024)Grattafiori, Dubey, Jauhri, Pandey, Kadian, Al-Dahle, Letman, Mathur, Schelten, Vaughan, Yang, Fan, Goyal, Hartshorn, Yang, Mitra, Sravankumar, Korenev, Hinsvark, Rao, Zhang, Rodriguez, Gregerson, Spataru, Roziere, Biron, Tang, Chern, Caucheteux, Nayak, Bi, Marra, McConnell, Keller, Touret, Wu, Wong, Ferrer, Nikolaidis, Allonsius, Song, Pintz, Livshits, Wyatt, Esiobu, Choudhary, Mahajan, Garcia-Olano, Perino, Hupkes, Lakomkin, AlBadawy, Lobanova, Dinan, Smith, Radenovic, Guzmán, Zhang, Synnaeve, Lee, Anderson, Thattai, Nail, Mialon, Pang, Cucurell, Nguyen, Korevaar, Xu, Touvron, Zarov, Ibarra, Kloumann, Misra, Evtimov, Zhang, Copet, Lee, Geffert, Vranes, Park, Mahadeokar, Shah, van~der Linde, Billock, Hong, Lee, Fu, Chi, Huang, Liu, Wang, Yu, Bitton, Spisak, Park, Rocca, Johnstun, Saxe, Jia, Alwala, Prasad, Upasani, Plawiak, Li, Heafield, Stone, El-Arini, Iyer, Malik, Chiu, Bhalla, Lakhotia, Rantala-Yeary, van~der Maaten, Chen, Tan, Jenkins, Martin, Madaan, Malo, Blecher,
  Landzaat, de~Oliveira, Muzzi, Pasupuleti, Singh, Paluri, Kardas, Tsimpoukelli, Oldham, Rita, Pavlova, Kambadur, Lewis, Si, Singh, Hassan, Goyal, Torabi, Bashlykov, Bogoychev, Chatterji, Zhang, Duchenne, Çelebi, Alrassy, Zhang, Li, Vasic, Weng, Bhargava, Dubal, Krishnan, Koura, Xu, He, Dong, Srinivasan, Ganapathy, Calderer, Cabral, Stojnic, Raileanu, Maheswari, Girdhar, Patel, Sauvestre, Polidoro, Sumbaly, Taylor, Silva, Hou, Wang, Hosseini, Chennabasappa, Singh, Bell, Kim, Edunov, Nie, Narang, Raparthy, Shen, Wan, Bhosale, Zhang, Vandenhende, Batra, Whitman, Sootla, Collot, Gururangan, Borodinsky, Herman, Fowler, Sheasha, Georgiou, Scialom, Speckbacher, Mihaylov, Xiao, Karn, Goswami, Gupta, Ramanathan, Kerkez, Gonguet, Do, Vogeti, Albiero, Petrovic, Chu, Xiong, Fu, Meers, Martinet, Wang, Wang, Tan, Xia, Xie, Jia, Wang, Goldschlag, Gaur, Babaei, Wen, Song, Zhang, Li, Mao, Coudert, Yan, Chen, Papakipos, Singh, Srivastava, Jain, Kelsey, Shajnfeld, Gangidi, Victoria, Goldstand, Menon, Sharma, Boesenberg,
  Baevski, Feinstein, Kallet, Sangani, Teo, Yunus, Lupu, Alvarado, Caples, Gu, Ho, Poulton, Ryan, Ramchandani, Dong, Franco, Goyal, Saraf, Chowdhury, Gabriel, Bharambe, Eisenman, Yazdan, James, Maurer, Leonhardi, Huang, Loyd, Paola, Paranjape, Liu, Wu, Ni, Hancock, Wasti, Spence, Stojkovic, Gamido, Montalvo, Parker, Burton, Mejia, Liu, Wang, Kim, Zhou, Hu, Chu, Cai, Tindal, Feichtenhofer, Gao, Civin, Beaty, Kreymer, Li, Adkins, Xu, Testuggine, David, Parikh, Liskovich, Foss, Wang, Le, Holland, Dowling, Jamil, Montgomery, Presani, Hahn, Wood, Le, Brinkman, Arcaute, Dunbar, Smothers, Sun, Kreuk, Tian, Kokkinos, Ozgenel, Caggioni, Kanayet, Seide, Florez, Schwarz, Badeer, Swee, Halpern, Herman, Sizov, Guangyi, Zhang, Lakshminarayanan, Inan, Shojanazeri, Zou, Wang, Zha, Habeeb, Rudolph, Suk, Aspegren, Goldman, Zhan, Damlaj, Molybog, Tufanov, Leontiadis, Veliche, Gat, Weissman, Geboski, Kohli, Lam, Asher, Gaya, Marcus, Tang, Chan, Zhen, Reizenstein, Teboul, Zhong, Jin, Yang, Cummings, Carvill, Shepard, McPhie,
  Torres, Ginsburg, Wang, Wu, U, Saxena, Khandelwal, Zand, Matosich, Veeraraghavan, Michelena, Li, Jagadeesh, Huang, Chawla, Huang, Chen, Garg, A, Silva, Bell, Zhang, Guo, Yu, Moshkovich, Wehrstedt, Khabsa, Avalani, Bhatt, Mankus, Hasson, Lennie, Reso, Groshev, Naumov, Lathi, Keneally, Liu, Seltzer, Valko, Restrepo, Patel, Vyatskov, Samvelyan, Clark, Macey, Wang, Hermoso, Metanat, Rastegari, Bansal, Santhanam, Parks, White, Bawa, Singhal, Egebo, Usunier, Mehta, Laptev, Dong, Cheng, Chernoguz, Hart, Salpekar, Kalinli, Kent, Parekh, Saab, Balaji, Rittner, Bontrager, Roux, Dollar, Zvyagina, Ratanchandani, Yuvraj, Liang, Alao, Rodriguez, Ayub, Murthy, Nayani, Mitra, Parthasarathy, Li, Hogan, Battey, Wang, Howes, Rinott, Mehta, Siby, Bondu, Datta, Chugh, Hunt, Dhillon, Sidorov, Pan, Mahajan, Verma, Yamamoto, Ramaswamy, Lindsay, Lindsay, Feng, Lin, Zha, Patil, Shankar, Zhang, Zhang, Wang, Agarwal, Sajuyigbe, Chintala, Max, Chen, Kehoe, Satterfield, Govindaprasad, Gupta, Deng, Cho, Virk, Subramanian, Choudhury,
  Goldman, Remez, Glaser, Best, Koehler, Robinson, Li, Zhang, Matthews, Chou, Shaked, Vontimitta, Ajayi, Montanez, Mohan, Kumar, Mangla, Ionescu, Poenaru, Mihailescu, Ivanov, Li, Wang, Jiang, Bouaziz, Constable, Tang, Wu, Wang, Wu, Gao, Kleinman, Chen, Hu, Jia, Qi, Li, Zhang, Zhang, Adi, Nam, Yu, Wang, Zhao, Hao, Qian, Li, He, Rait, DeVito, Rosnbrick, Wen, Yang, Zhao, and Ma}]{Llama-3}
Aaron Grattafiori, Abhimanyu Dubey, Abhinav Jauhri, Abhinav Pandey, Abhishek Kadian, Ahmad Al-Dahle, Aiesha Letman, Akhil Mathur, Alan Schelten, Alex Vaughan, Amy Yang, Angela Fan, Anirudh Goyal, Anthony Hartshorn, Aobo Yang, Archi Mitra, Archie Sravankumar, Artem Korenev, Arthur Hinsvark, and 542 others. 2024.
\newblock \href {https://arxiv.org/abs/2407.21783} {The llama 3 herd of models}.
\newblock \emph{Preprint}, arXiv:2407.21783.

\bibitem[{Guo et~al.(2025)Guo, Liu, Fan, He, Li, Li, Wang, and Fung}]{guo2025mathematical}
Dadi Guo, Jiayu Liu, Zhiyuan Fan, Zhitao He, Haoran Li, Yuxin Li, Yumeng Wang, and Yi~R Fung. 2025.
\newblock Mathematical proof as a litmus test: Revealing failure modes of advanced large reasoning models.
\newblock \emph{arXiv preprint arXiv:2506.17114}.

\bibitem[{Guo et~al.(2026)Guo, Xie, Liu, Liu, Fan, Ren, Shao, Zhou, Liu, and Fung}]{Code2Math}
Dadi Guo, Yuejin Xie, Qingyu Liu, Jiayu Liu, Zhiyuan Fan, Qihan Ren, Shuai Shao, Tianyi Zhou, Dongrui Liu, and Yi~R Fung. 2026.
\newblock Code2math: Can your code agent effectively evolve math problems through exploration?
\newblock \emph{arXiv preprint arXiv:2603.03202}.

\bibitem[{Hallinan et~al.(2026)Hallinan, Venkatesh, Ren, Karimireddy, Paranjape, Zhang, and Hessel}]{OpaqueToolsBench}
Skyler Hallinan, Thejas Venkatesh, Xiang Ren, Sai~Praneeth Karimireddy, Ashwin Paranjape, Yuhao Zhang, and Jack Hessel. 2026.
\newblock \href {https://arxiv.org/abs/2602.15197} {Opaquetoolsbench: Learning nuances of tool behavior through interaction}.
\newblock \emph{Preprint}, arXiv:2602.15197.

\bibitem[{Hong et~al.(2024)Hong, Zhuge, Chen, Zheng, Cheng, Wang, Zhang, Yau, Lin, Zhou et~al.}]{hong2024metagpt}
Sirui Hong, Mingchen Zhuge, Jonathan Chen, Xiawu Zheng, Yuheng Cheng, Jinlin Wang, Ceyao Zhang, Steven Yau, Zijuan Lin, Liyang Zhou, and 1 others. 2024.
\newblock Metagpt: Meta programming for a multi-agent collaborative framework.
\newblock In \emph{International Conference on Learning Representations}, volume 2024, pages 23247--23275.

\bibitem[{Hou et~al.(2025)Hou, Zhao, Wang, and Wang}]{MCP}
Xinyi Hou, Yanjie Zhao, Shenao Wang, and Haoyu Wang. 2025.
\newblock \href {https://arxiv.org/abs/2503.23278} {Model context protocol (mcp): Landscape, security threats, and future research directions}.
\newblock \emph{Preprint}, arXiv:2503.23278.

\bibitem[{Huang et~al.(2024)Huang, Shi, Li, Fan, Wu, Zhang, Liu, Zhou, Wan, Gong, and Sun}]{huang2024metatoolbenchmarklargelanguage}
Yue Huang, Jiawen Shi, Yuan Li, Chenrui Fan, Siyuan Wu, Qihui Zhang, Yixin Liu, Pan Zhou, Yao Wan, Neil~Zhenqiang Gong, and Lichao Sun. 2024.
\newblock \href {https://arxiv.org/abs/2310.03128} {Metatool benchmark for large language models: Deciding whether to use tools and which to use}.
\newblock \emph{Preprint}, arXiv:2310.03128.

\bibitem[{Justus et~al.(2024)Justus, Rodrigues, and Sousa}]{justus2024bootstrap}
Vin{\'\i}cius~Litvinoff Justus, Vitor~Batista Rodrigues, and Alex Rodrigo dos~Santos Sousa. 2024.
\newblock Bootstrap confidence intervals: A comparative simulation study.
\newblock \emph{arXiv preprint arXiv:2404.12967}.

\bibitem[{Karpas et~al.(2022)Karpas, Abend, Belinkov, Lenz, Lieber, Ratner, Shoham, Bata, Levine, Leyton-Brown et~al.}]{karpas2022mrkl}
Ehud Karpas, Omri Abend, Yonatan Belinkov, Barak Lenz, Opher Lieber, Nir Ratner, Yoav Shoham, Hofit Bata, Yoav Levine, Kevin Leyton-Brown, and 1 others. 2022.
\newblock Mrkl systems: A modular, neuro-symbolic architecture that combines large language models, external knowledge sources and discrete reasoning.
\newblock \emph{arXiv preprint arXiv:2205.00445}.

\bibitem[{Kim et~al.(2026)Kim, Ren, Hao, Sun, Wang, Ma, Ye, Han, Yin, Ji, Shen, Fan, Yao, and Guo}]{kim2026perfectapiscomprehensiveevaluation}
Doyoung Kim, Zhiwei Ren, Jie Hao, Zhongkai Sun, Lichao Wang, Xiyao Ma, Zack Ye, Xu~Han, Jun Yin, Heng Ji, Wei Shen, Xing Fan, Benjamin Yao, and Chenlei Guo. 2026.
\newblock \href {https://arxiv.org/abs/2601.00268} {Beyond perfect apis: A comprehensive evaluation of llm agents under real-world api complexity}.
\newblock \emph{Preprint}, arXiv:2601.00268.

\bibitem[{Koh et~al.(2026{\natexlab{a}})Koh, McAleer, Fried, and Salakhutdinov}]{koh2025treesearch}
Jing~Yu Koh, Stephen McAleer, Daniel Fried, and Ruslan Salakhutdinov. 2026{\natexlab{a}}.
\newblock \href {https://arxiv.org/abs/2407.01476} {Tree search for language model agents}.
\newblock \emph{Preprint}, arXiv:2407.01476.

\bibitem[{Koh et~al.(2026{\natexlab{b}})Koh, McAleer, Fried, and Salakhutdinov}]{koh2026tree}
Jing~Yu Koh, Stephen McAleer, Daniel Fried, and Ruslan Salakhutdinov. 2026{\natexlab{b}}.
\newblock \href {https://arxiv.org/abs/2407.01476} {Tree search for language model agents}.
\newblock \emph{Preprint}, arXiv:2407.01476.

\bibitem[{Kwon et~al.(2023)Kwon, Li, Zhuang, Sheng, Zheng, Yu, Gonzalez, Zhang, and Stoica}]{vllm}
Woosuk Kwon, Zhuohan Li, Siyuan Zhuang, Ying Sheng, Lianmin Zheng, Cody~Hao Yu, Joseph~E. Gonzalez, Hao Zhang, and Ion Stoica. 2023.
\newblock Efficient memory management for large language model serving with pagedattention.
\newblock In \emph{Proceedings of the ACM SIGOPS 29th Symposium on Operating Systems Principles}.

\bibitem[{Li et~al.(2023{\natexlab{a}})Li, Hammoud, Itani, Khizbullin, and Ghanem}]{li2023camel}
Guohao Li, Hasan Hammoud, Hani Itani, Dmitrii Khizbullin, and Bernard Ghanem. 2023{\natexlab{a}}.
\newblock Camel: Communicative agents for" mind" exploration of large language model society.
\newblock \emph{Advances in neural information processing systems}, 36:51991--52008.

\bibitem[{Li et~al.(2026)Li, Zhao, Zhao, Zeng, Wu, Wang, Ge, Cao, Huang, Liu, Liu, Su, Guo, Zhou, Zhang, Michelini, Wang, Yue, Zhou, Neubig, and He}]{Tool-Decathlon}
Junlong Li, Wenshuo Zhao, Jian Zhao, Weihao Zeng, Haoze Wu, Xiaochen Wang, Rui Ge, Yuxuan Cao, Yuzhen Huang, Wei Liu, Junteng Liu, Zhaochen Su, Yiyang Guo, Fan Zhou, Lueyang Zhang, Juan Michelini, Xingyao Wang, Xiang Yue, Shuyan Zhou, and 2 others. 2026.
\newblock \href {https://arxiv.org/abs/2510.25726} {The tool decathlon: Benchmarking language agents for diverse, realistic, and long-horizon task execution}.
\newblock \emph{Preprint}, arXiv:2510.25726.

\bibitem[{Li et~al.(2023{\natexlab{b}})Li, Zhao, Yu, Song, Li, Yu, Li, Huang, and Li}]{API-Bank}
Minghao Li, Yingxiu Zhao, Bowen Yu, Feifan Song, Hangyu Li, Haiyang Yu, Zhoujun Li, Fei Huang, and Yongbin Li. 2023{\natexlab{b}}.
\newblock \href {https://doi.org/10.18653/v1/2023.emnlp-main.187} {{API}-bank: A comprehensive benchmark for tool-augmented {LLM}s}.
\newblock In \emph{Proceedings of the 2023 Conference on Empirical Methods in Natural Language Processing}, pages 3102--3116, Singapore. Association for Computational Linguistics.

\bibitem[{Likert(1932)}]{likert1932technique}
Rensis Likert. 1932.
\newblock \href {https://www.scirp.org/reference/ReferencesPapers?ReferenceID=534541} {A technique for the measurement of attitudes}.
\newblock \emph{Archives of Psychology}, (140):5--55.

\bibitem[{Ling et~al.(2025)Ling, Wang, Fan, Lam, and Hu}]{ling2025elhplan}
Shaobin Ling, Yun Wang, Chenyou Fan, Tin~Lun Lam, and Junjie Hu. 2025.
\newblock Elhplan: Efficient long-horizon task planning for multi-agent collaboration.
\newblock \emph{arXiv preprint arXiv:2509.24230}.

\bibitem[{Liu et~al.(2025{\natexlab{a}})Liu, Qian, Su, Zong, Huang, He, and Fung}]{CostBench}
Jiayu Liu, Cheng Qian, Zhaochen Su, Qing Zong, Shijue Huang, Bingxiang He, and Yi~R Fung. 2025{\natexlab{a}}.
\newblock Costbench: Evaluating multi-turn cost-optimal planning and adaptation in dynamic environments for llm tool-use agents.
\newblock \emph{arXiv preprint arXiv:2511.02734}.

\bibitem[{Liu et~al.(2026{\natexlab{a}})Liu, Qian, Wang, Li, Liu, Wang, Kim, Wang, Chen, Fung et~al.}]{AdaPlanBench}
Jiayu Liu, Cheng Qian, Zhenhailong Wang, Bingxuan Li, Jiateng Liu, Heng Wang, Jeonghwan Kim, Yumeng Wang, Xiusi Chen, Yi~R Fung, and 1 others. 2026{\natexlab{a}}.
\newblock Adaplanbench: Evaluating adaptive planning in large language model agents under world and user constraints.
\newblock \emph{arXiv preprint arXiv:2606.05622}.

\bibitem[{Liu et~al.(2024)Liu, Tang, Wang, Xu, Shi, Wang, and Song}]{liu-etal-2024-gprooft}
Jiayu Liu, Junhao Tang, Hanwen Wang, Baixuan Xu, Haochen Shi, Weiqi Wang, and Yangqiu Song. 2024.
\newblock \href {https://doi.org/10.18653/v1/2024.fever-1.14} {{GP}roof{T}: A multi-dimension multi-round fact checking framework based on claim fact extraction}.
\newblock In \emph{Proceedings of the Seventh Fact Extraction and VERification Workshop (FEVER)}, pages 118--129, Miami, Florida, USA. Association for Computational Linguistics.

\bibitem[{Liu et~al.(2026{\natexlab{b}})Liu, Wang, Zong, Zeng, Zheng, Shi, Guo, Xu, Li, and Song}]{NAACL}
Jiayu Liu, Rui Wang, Qing Zong, Qingcheng Zeng, Tianshi Zheng, Haochen Shi, Dadi Guo, Baixuan Xu, Chunyang Li, and Yangqiu Song. 2026{\natexlab{b}}.
\newblock Naacl: Noise-aware verbal confidence calibration for llms in rag systems.
\newblock \emph{arXiv preprint arXiv:2601.11004}.

\bibitem[{Liu et~al.(2025{\natexlab{b}})Liu, Zong, Wang, and Song}]{marcon}
Jiayu Liu, Qing Zong, Weiqi Wang, and Yangqiu Song. 2025{\natexlab{b}}.
\newblock \href {https://doi.org/10.18653/v1/2025.acl-short.18} {Revisiting epistemic markers in confidence estimation: Can markers accurately reflect large language models' uncertainty?}
\newblock In \emph{Proceedings of the 63rd Annual Meeting of the Association for Computational Linguistics (Volume 2: Short Papers)}, pages 206--221, Vienna, Austria. Association for Computational Linguistics.

\bibitem[{Liu et~al.(2023)Liu, Yu, Zhang, Xu, Lei, Lai, Gu, Ding, Men, Yang, Zhang, Deng, Zeng, Du, Zhang, Shen, Zhang, Su, Sun, Huang, Dong, and Tang}]{AgentBench}
Xiao Liu, Hao Yu, Hanchen Zhang, Yifan Xu, Xuanyu Lei, Hanyu Lai, Yu~Gu, Hangliang Ding, Kaiwen Men, Kejuan Yang, Shudan Zhang, Xiang Deng, Aohan Zeng, Zhengxiao Du, Chenhui Zhang, Sheng Shen, Tianjun Zhang, Yu~Su, Huan Sun, and 3 others. 2023.
\newblock \href {https://arxiv.org/abs/2308.03688} {{AgentBench}: Evaluating {LLM}s as agents}.
\newblock \emph{Preprint}, arXiv:2308.03688.

\bibitem[{Lu et~al.(2025)Lu, Holleis, Zhang, Aumayer, Nan, Bai, Ma, Ma, Li, Yin, Wang, and Pang}]{ToolSandbox}
Jiarui Lu, Thomas Holleis, Yizhe Zhang, Bernhard Aumayer, Feng Nan, Haoping Bai, Shuang Ma, Shen Ma, Mengyu Li, Guoli Yin, Zirui Wang, and Ruoming Pang. 2025.
\newblock \href {https://doi.org/10.18653/v1/2025.findings-naacl.65} {{T}ool{S}andbox: A stateful, conversational, interactive evaluation benchmark for {LLM} tool use capabilities}.
\newblock In \emph{Findings of the Association for Computational Linguistics: NAACL 2025}, pages 1160--1183, Albuquerque, New Mexico. Association for Computational Linguistics.

\bibitem[{Luo et~al.(2025)Luo, Shen, Yang, Zhao, Jwalapuram, Saha, Sahoo, Savarese, Xiong, and Li}]{MCP-Universe}
Ziyang Luo, Zhiqi Shen, Wenzhuo Yang, Zirui Zhao, Prathyusha Jwalapuram, Amrita Saha, Doyen Sahoo, Silvio Savarese, Caiming Xiong, and Junnan Li. 2025.
\newblock \href {https://arxiv.org/abs/2508.14704} {Mcp-universe: Benchmarking large language models with real-world model context protocol servers}.
\newblock \emph{Preprint}, arXiv:2508.14704.

\bibitem[{Mo et~al.(2026)Mo, Zhong, Chen, Yuan, Chen, Lu, Lin, He, Han, and Sun}]{LiveMCPBench}
Guozhao Mo, Wenliang Zhong, Jiawei Chen, Qianhao Yuan, Xuanang Chen, Yaojie Lu, Hongyu Lin, Ben He, Xianpei Han, and Le~Sun. 2026.
\newblock \href {https://arxiv.org/abs/2508.01780} {Livemcpbench: Can agents navigate an ocean of mcp tools?}
\newblock \emph{Preprint}, arXiv:2508.01780.

\bibitem[{Muennighoff et~al.(2025)Muennighoff, Yang, Shi, Li, Fei-Fei, Hajishirzi, Zettlemoyer, Liang, Cand{\`e}s, and Hashimoto}]{s1}
Niklas Muennighoff, Zitong Yang, Weijia Shi, Xiang~Lisa Li, Li~Fei-Fei, Hannaneh Hajishirzi, Luke Zettlemoyer, Percy Liang, Emmanuel Cand{\`e}s, and Tatsunori Hashimoto. 2025.
\newblock \href {https://doi.org/10.18653/v1/2025.emnlp-main.1025} {s1: Simple test-time scaling}.
\newblock In \emph{Proceedings of the 2025 Conference on Empirical Methods in Natural Language Processing}, pages 20275--20321, Suzhou, China. Association for Computational Linguistics.

\bibitem[{Newell and Simon(1972)}]{newell1972human}
A.~Newell and H.A. Simon. 1972.
\newblock \href {https://books.google.com/books?id=h03uAAAAMAAJ} {\emph{Human Problem Solving}}.
\newblock ACS symposium series. Prentice-Hall.

\bibitem[{{OpenAI}(2026)}]{GPT-5.4}
{OpenAI}. 2026.
\newblock Introducing gpt-5.4.
\newblock \url{https://openai.com/index/introducing-gpt-5-4/}.

\bibitem[{Patil et~al.(2025)Patil, Mao, Yan, Ji, Suresh, Stoica, and Gonzalez}]{BFCL-v4}
Shishir~G Patil, Huanzhi Mao, Fanjia Yan, Charlie Cheng-Jie Ji, Vishnu Suresh, Ion Stoica, and Joseph~E. Gonzalez. 2025.
\newblock \href {https://openreview.net/forum?id=2GmDdhBdDk} {The berkeley function calling leaderboard ({BFCL}): From tool use to agentic evaluation of large language models}.
\newblock In \emph{Forty-second International Conference on Machine Learning}.

\bibitem[{Patil et~al.(2023)Patil, Zhang, Wang, and Gonzalez}]{APIBench}
Shishir~G. Patil, Tianjun Zhang, Xin Wang, and Joseph~E. Gonzalez. 2023.
\newblock \href {https://arxiv.org/abs/2305.15334} {Gorilla: Large language model connected with massive apis}.
\newblock \emph{Preprint}, arXiv:2305.15334.

\bibitem[{Qian et~al.(2026{\natexlab{a}})Qian, Acikgoz, He, Wang, Chen, Hakkani-Tur, Tur, and Ji}]{ToolRL}
Cheng Qian, Emre~Can Acikgoz, Qi~He, Hongru Wang, Xiusi Chen, Dilek Hakkani-Tur, Gokhan Tur, and Heng Ji. 2026{\natexlab{a}}.
\newblock Toolrl: Reward is all tool learning needs.
\newblock \emph{Advances in Neural Information Processing Systems}, 38:105523--105553.

\bibitem[{Qian et~al.(2026{\natexlab{b}})Qian, Ha, Liu, He, Kim, Liu, Li, Tiwari, Dalal, Wang et~al.}]{CreativityBench}
Cheng Qian, Hyeonjeong Ha, Jiayu Liu, Bingxiang He, Jeonghwan Kim, Jiateng Liu, Bingxuan Li, Aditi Tiwari, Dwip Dalal, Zhenhailong Wang, and 1 others. 2026{\natexlab{b}}.
\newblock Creativitybench: Evaluating agent creative reasoning via affordance-based tool repurposing.
\newblock \emph{arXiv preprint arXiv:2605.02910}.

\bibitem[{Qian et~al.(2025{\natexlab{a}})Qian, Han, Luo, He, Chen, Zhang, Du, Yao, Yang, Zhang, Li, and Ji}]{EscapeBench}
Cheng Qian, Peixuan Han, Qinyu Luo, Bingxiang He, Xiusi Chen, Yuji Zhang, Hongyi Du, Jiarui Yao, Xiaocheng Yang, Denghui Zhang, Yunzhu Li, and Heng Ji. 2025{\natexlab{a}}.
\newblock \href {https://doi.org/10.18653/v1/2025.acl-long.39} {{E}scape{B}ench: Towards advancing creative intelligence of language model agents}.
\newblock In \emph{Proceedings of the 63rd Annual Meeting of the Association for Computational Linguistics (Volume 1: Long Papers)}, pages 798--820, Vienna, Austria. Association for Computational Linguistics.

\bibitem[{Qian et~al.(2025{\natexlab{b}})Qian, Liu, Kokane, Prabhakar, Qiu, Chen, Liu, Ji, Yao, Heinecke, Savarese, Xiong, and Wang}]{xRouter}
Cheng Qian, Zuxin Liu, Shirley Kokane, Akshara Prabhakar, Jielin Qiu, Haolin Chen, Zhiwei Liu, Heng Ji, Weiran Yao, Shelby Heinecke, Silvio Savarese, Caiming Xiong, and Huan Wang. 2025{\natexlab{b}}.
\newblock \href {https://arxiv.org/abs/2510.08439} {xrouter: Training cost-aware llms orchestration system via reinforcement learning}.
\newblock \emph{Preprint}, arXiv:2510.08439.

\bibitem[{Qian et~al.(2025{\natexlab{c}})Qian, Liu, Prabhakar, Liu, Zhang, Chen, Ji, Yao, Heinecke, Savarese, Xiong, and Wang}]{UserBench}
Cheng Qian, Zuxin Liu, Akshara Prabhakar, Zhiwei Liu, Jianguo Zhang, Haolin Chen, Heng Ji, Weiran Yao, Shelby Heinecke, Silvio Savarese, Caiming Xiong, and Huan Wang. 2025{\natexlab{c}}.
\newblock \href {https://arxiv.org/abs/2507.22034} {Userbench: An interactive gym environment for user-centric agents}.
\newblock \emph{Preprint}, arXiv:2507.22034.

\bibitem[{Qin et~al.(2023{\natexlab{a}})Qin, Liang, Ye, Zhu, Yan, Lu, Lin, Cong, Tang, Qian, Zhao, Hong, Tian, Xie, Zhou, Gerstein, Li, Liu, and Sun}]{ToolBench}
Yujia Qin, Shihao Liang, Yining Ye, Kunlun Zhu, Lan Yan, Yaxi Lu, Yankai Lin, Xin Cong, Xiangru Tang, Bill Qian, Sihan Zhao, Lauren Hong, Runchu Tian, Ruobing Xie, Jie Zhou, Mark Gerstein, Dahai Li, Zhiyuan Liu, and Maosong Sun. 2023{\natexlab{a}}.
\newblock \href {https://arxiv.org/abs/2307.16789} {Toolbench: Facilitating large language models to master 16000+ real-world apis}.
\newblock \emph{Preprint}, arXiv:2307.16789.

\bibitem[{Qin et~al.(2023{\natexlab{b}})Qin, Liang, Ye, Zhu, Yan, Lu, Lin, Cong, Tang, Qian, Zhao, Hong, Tian, Xie, Zhou, Gerstein, Li, Liu, and Sun}]{ToolLLM}
Yujia Qin, Shihao Liang, Yining Ye, Kunlun Zhu, Lan Yan, Yaxi Lu, Yankai Lin, Xin Cong, Xiangru Tang, Bill Qian, Sihan Zhao, Lauren Hong, Runchu Tian, Ruobing Xie, Jie Zhou, Mark Gerstein, Dahai Li, Zhiyuan Liu, and Maosong Sun. 2023{\natexlab{b}}.
\newblock \href {https://arxiv.org/abs/2307.16789} {{ToolLLM}: Facilitating large language models to master 16000+ real-world {API}s}.
\newblock \emph{Preprint}, arXiv:2307.16789.

\bibitem[{Qu et~al.(2024)Qu, Dai, Wei, Cai, Wang, Yin, Xu, and Wen}]{COLT}
Changle Qu, Sunhao Dai, Xiaochi Wei, Hengyi Cai, Shuaiqiang Wang, Dawei Yin, Jun Xu, and Ji-Rong Wen. 2024.
\newblock \href {https://doi.org/10.1145/3627673.3679847} {Towards completeness-oriented tool retrieval for large language models}.
\newblock In \emph{Proceedings of the 33rd ACM International Conference on Information and Knowledge Management}, CIKM ’24, page 1930–1940. ACM.

\bibitem[{Ren et~al.(2024)Ren, Ichter, and Majumdar}]{BackwardPlanningLLM}
Allen~Z. Ren, Brian Ichter, and Anirudha Majumdar. 2024.
\newblock \href {https://arxiv.org/abs/2411.01790} {Thinking forward and backward: Effective backward planning with large language models}.
\newblock \emph{Preprint}, arXiv:2411.01790.

\bibitem[{Schick et~al.(2023)Schick, Dwivedi-Yu, Dess{\`\i}, Raileanu, Lomeli, Hambro, Zettlemoyer, Cancedda, and Scialom}]{schick2023toolformer}
Timo Schick, Jane Dwivedi-Yu, Roberto Dess{\`\i}, Roberta Raileanu, Maria Lomeli, Eric Hambro, Luke Zettlemoyer, Nicola Cancedda, and Thomas Scialom. 2023.
\newblock Toolformer: Language models can teach themselves to use tools.
\newblock \emph{Advances in neural information processing systems}, 36:68539--68551.

\bibitem[{Shen et~al.(2023)Shen, Song, Tan, Zhang, Ren, Yuan, Lu, Li, and Zhuang}]{shen2023taskbench}
Yongliang Shen, Kaitao Song, Xu~Tan, Wenqi Zhang, Kan Ren, Siyu Yuan, Weiming Lu, Dongsheng Li, and Yueting Zhuang. 2023.
\newblock \href {https://arxiv.org/abs/2311.18760} {Taskbench: Benchmarking large language models for task automation}.
\newblock \emph{Preprint}, arXiv:2311.18760.

\bibitem[{Shi et~al.(2025)Shi, Wang, Yan, Ren, Wang, Yin, and Ren}]{ToolRet}
Zhengliang Shi, Yuhan Wang, Lingyong Yan, Pengjie Ren, Shuaiqiang Wang, Dawei Yin, and Zhaochun Ren. 2025.
\newblock \href {https://doi.org/10.18653/v1/2025.findings-acl.1258} {Retrieval models aren{'}t tool-savvy: Benchmarking tool retrieval for large language models}.
\newblock In \emph{Findings of the Association for Computational Linguistics: ACL 2025}, pages 24497--24524, Vienna, Austria. Association for Computational Linguistics.

\bibitem[{Song et~al.(2023)Song, Xiong, Zhu, Wu, Qian, Song, Huang, Li, Wang, Yao, Tian, and Li}]{RestGPT}
Yifan Song, Weimin Xiong, Dawei Zhu, Wenhao Wu, Han Qian, Mingbo Song, Hailiang Huang, Cheng Li, Ke~Wang, Rong Yao, Ye~Tian, and Sujian Li. 2023.
\newblock \href {https://arxiv.org/abs/2306.06624} {Restgpt: Connecting large language models with real-world restful apis}.
\newblock \emph{Preprint}, arXiv:2306.06624.

\bibitem[{Su et~al.(2025)Su, Diao, Lu, Liu, Xu, Dong, Fu, Belcak, Ye, Yin, Dong, Bakhturina, Yu, Choi, Kautz, and Molchanov}]{ToolOrchestra}
Hongjin Su, Shizhe Diao, Ximing Lu, Mingjie Liu, Jiacheng Xu, Xin Dong, Yonggan Fu, Peter Belcak, Hanrong Ye, Hongxu Yin, Yi~Dong, Evelina Bakhturina, Tao Yu, Yejin Choi, Jan Kautz, and Pavlo Molchanov. 2025.
\newblock \href {https://arxiv.org/abs/2511.21689} {Toolorchestra: Elevating intelligence via efficient model and tool orchestration}.
\newblock \emph{Preprint}, arXiv:2511.21689.

\bibitem[{Sullivan et~al.(2025)Sullivan, Hartmann, and Koller}]{RandomWorld}
Michael Sullivan, Mareike Hartmann, and Alexander Koller. 2025.
\newblock \href {https://arxiv.org/abs/2506.11045} {Procedural environment generation for tool-use agents}.
\newblock \emph{Preprint}, arXiv:2506.11045.

\bibitem[{Trivedi et~al.(2024)Trivedi, Khot, Hartmann, Manku, Dong, Li, Gupta, Sabharwal, and Balasubramanian}]{AppWorld}
Harsh Trivedi, Tushar Khot, Mareike Hartmann, Ruskin Manku, Vinty Dong, Edward Li, Shashank Gupta, Ashish Sabharwal, and Niranjan Balasubramanian. 2024.
\newblock \href {https://doi.org/10.18653/v1/2024.acl-long.850} {{A}pp{W}orld: A controllable world of apps and people for benchmarking interactive coding agents}.
\newblock In \emph{Proceedings of the 62nd Annual Meeting of the Association for Computational Linguistics (Volume 1: Long Papers)}, pages 16022--16076, Bangkok, Thailand. Association for Computational Linguistics.

\bibitem[{Turney(2024)}]{Pearson}
Shaun Turney. 2024.
\newblock \href {https://www.scribbr.com/statistics/pearson-correlation-coefficient/} {Pearson correlation coefficient (r) | guide \& examples}.
\newblock \textit{Scribbr}.

\bibitem[{Verma and Bharadwaj(2025)}]{verma2025leap}
Nikhil Verma and Manasa Bharadwaj. 2025.
\newblock \href {https://doi.org/10.18653/v1/2025.acl-industry.64} {{LEAP} {\&} {LEAN}: Look-ahead planning and agile navigation for {LLM} agents}.
\newblock In \emph{Proceedings of the 63rd Annual Meeting of the Association for Computational Linguistics (Volume 6: Industry Track)}, pages 896--933, Vienna, Austria. Association for Computational Linguistics.

\bibitem[{Wang et~al.(2023)Wang, Xu, Lan, Hu, Lan, Lee, and Lim}]{PlanAndSolve}
Lei Wang, Wanyu Xu, Yihuai Lan, Zhiqiang Hu, Yunshi Lan, Roy Ka-Wei Lee, and Ee-Peng Lim. 2023.
\newblock \href {https://arxiv.org/abs/2305.04091} {Plan-and-solve prompting: Improving zero-shot chain-of-thought reasoning by large language models}.
\newblock \emph{Preprint}, arXiv:2305.04091.

\bibitem[{Wang et~al.(2025{\natexlab{a}})Wang, Han, Ji, Wang, Baldwin, and Li}]{ToolGen}
Renxi Wang, Xudong Han, Lei Ji, Shu Wang, Timothy Baldwin, and Haonan Li. 2025{\natexlab{a}}.
\newblock \href {https://arxiv.org/abs/2410.03439} {Toolgen: Unified tool retrieval and calling via generation}.
\newblock \emph{Preprint}, arXiv:2410.03439.

\bibitem[{Wang et~al.(2025{\natexlab{b}})Wang, Lin, Liu, Zong, Zheng, Wang, and Song}]{wang2025prospect}
Rui Wang, Qihan Lin, Jiayu Liu, Qing Zong, Tianshi Zheng, Weiqi Wang, and Yangqiu Song. 2025{\natexlab{b}}.
\newblock Prospect theory fails for llms: Revealing instability of decision-making under epistemic uncertainty.
\newblock \emph{arXiv preprint arXiv:2508.08992}.

\bibitem[{Wang et~al.(2026)Wang, Chen, Wang, Wu, Fang, Cai, Gu, Su, Zhang, Wang, Cai, and Chua}]{AgentNoiseBench}
Ruipeng Wang, Yuxin Chen, Yukai Wang, Chang Wu, Junfeng Fang, Xiaodong Cai, Qi~Gu, Hui Su, An~Zhang, Xiang Wang, Xunliang Cai, and Tat-Seng Chua. 2026.
\newblock \href {https://arxiv.org/abs/2602.11348} {Agentnoisebench: Benchmarking robustness of tool-using llm agents under noisy condition}.
\newblock \emph{Preprint}, arXiv:2602.11348.

\bibitem[{Wang et~al.(2025{\natexlab{c}})Wang, Fan, Liu, Huang, and Fung}]{wang2025diversity}
Yumeng Wang, Zhiyuan Fan, Jiayu Liu, Jen-tse Huang, and Yi~R Fung. 2025{\natexlab{c}}.
\newblock Diversity-enhanced reasoning for subjective questions.
\newblock \emph{arXiv preprint arXiv:2507.20187}.

\bibitem[{Wang et~al.(2025{\natexlab{d}})Wang, Chang, Patel, Biju, Wu, Liu, Ding, Rezazadeh, Shah, Bao, and Siow}]{MCP-Bench}
Zhenting Wang, Qi~Chang, Hemani Patel, Shashank Biju, Cheng-En Wu, Quan Liu, Aolin Ding, Alireza Rezazadeh, Ankit Shah, Yujia Bao, and Eugene Siow. 2025{\natexlab{d}}.
\newblock \href {https://arxiv.org/abs/2508.20453} {Mcp-bench: Benchmarking tool-using llm agents with complex real-world tasks via mcp servers}.
\newblock \emph{Preprint}, arXiv:2508.20453.

\bibitem[{Wu et~al.(2024)Wu, Bansal, Zhang, Wu, Li, Zhu, Jiang, Zhang, Zhang, Liu et~al.}]{wu2024autogen}
Qingyun Wu, Gagan Bansal, Jieyu Zhang, Yiran Wu, Beibin Li, Erkang Zhu, Li~Jiang, Xiaoyun Zhang, Shaokun Zhang, Jiale Liu, and 1 others. 2024.
\newblock Autogen: Enabling next-gen llm applications via multi-agent conversations.
\newblock In \emph{First conference on language modeling}.

\bibitem[{Xi et~al.(2026)Xi, Liang, Liu, Zhang, Peng, Nan, Nayim, Zhang, Mundada, Qin, Huang, and Zhou}]{ToolGym}
Ziqiao Xi, Shuang Liang, Qi~Liu, Jiaqing Zhang, Letian Peng, Fang Nan, Meshal Nayim, Tianhui Zhang, Rishika Mundada, Lianhui Qin, Biwei Huang, and Kun Zhou. 2026.
\newblock \href {https://arxiv.org/abs/2601.06328} {C-world: A computer use agent environment creator}.
\newblock \emph{Preprint}, arXiv:2601.06328.

\bibitem[{Xu et~al.(2024{\natexlab{a}})Xu, Li, Xia, and Li}]{Tool-Retrieval-is-not-reliable-1}
Qiancheng Xu, Yongqi Li, Heming Xia, and Wenjie Li. 2024{\natexlab{a}}.
\newblock \href {https://arxiv.org/abs/2406.17465} {Enhancing tool retrieval with iterative feedback from large language models}.
\newblock \emph{Preprint}, arXiv:2406.17465.

\bibitem[{Xu et~al.(2024{\natexlab{b}})Xu, Li, Xia, and Li}]{xu2024enhancing}
Qiancheng Xu, Yongqi Li, Heming Xia, and Wenjie Li. 2024{\natexlab{b}}.
\newblock \href {https://arxiv.org/abs/2406.17465} {Enhancing tool retrieval with iterative feedback from large language models}.
\newblock \emph{Preprint}, arXiv:2406.17465.

\bibitem[{Yang et~al.(2025{\natexlab{a}})Yang, Li, Yang, Zhang, Hui, Zheng, Yu, Gao, Huang, Lv, Zheng, Liu, Zhou, Huang, Hu, Ge, Wei, Lin, Tang, Yang, Tu, Zhang, Yang, Yang, Zhou, Zhou, Lin, Dang, Bao, Yang, Yu, Deng, Li, Xue, Li, Zhang, Wang, Zhu, Men, Gao, Liu, Luo, Li, Tang, Yin, Ren, Wang, Zhang, Ren, Fan, Su, Zhang, Zhang, Wan, Liu, Wang, Cui, Zhang, Zhou, and Qiu}]{Qwen3}
An~Yang, Anfeng Li, Baosong Yang, Beichen Zhang, Binyuan Hui, Bo~Zheng, Bowen Yu, Chang Gao, Chengen Huang, Chenxu Lv, Chujie Zheng, Dayiheng Liu, Fan Zhou, Fei Huang, Feng Hu, Hao Ge, Haoran Wei, Huan Lin, Jialong Tang, and 41 others. 2025{\natexlab{a}}.
\newblock \href {https://arxiv.org/abs/2505.09388} {Qwen3 technical report}.
\newblock \emph{Preprint}, arXiv:2505.09388.

\bibitem[{Yang et~al.(2024)Yang, Jimenez, Wettig, Lieret, Yao, Narasimhan, and Press}]{SWEAgent}
John Yang, Carlos~E. Jimenez, Alexander Wettig, Kilian Lieret, Shunyu Yao, Karthik Narasimhan, and Ofir Press. 2024.
\newblock \href {https://arxiv.org/abs/2405.15793} {{SWE}-agent: Agent-computer interfaces enable automated software engineering}.
\newblock \emph{Preprint}, arXiv:2405.15793.

\bibitem[{Yang et~al.(2025{\natexlab{b}})Yang, Chen, Zhang, Yuan, Chen, Richardson, Xiao, and Yang}]{SELFGOAL}
Ruihan Yang, Jiangjie Chen, Yikai Zhang, Siyu Yuan, Aili Chen, Kyle Richardson, Yanghua Xiao, and Deqing Yang. 2025{\natexlab{b}}.
\newblock \href {https://doi.org/10.18653/v1/2025.naacl-long.36} {{SELFGOAL}: Your language agents already know how to achieve high-level goals}.
\newblock In \emph{Proceedings of the 2025 Conference of the Nations of the Americas Chapter of the Association for Computational Linguistics: Human Language Technologies (Volume 1: Long Papers)}, pages 799--819, Albuquerque, New Mexico. Association for Computational Linguistics.

\bibitem[{Yao et~al.(2024)Yao, Shinn, Razavi, and Narasimhan}]{yao2024taubench}
Shunyu Yao, Noah Shinn, Pedram Razavi, and Karthik Narasimhan. 2024.
\newblock \href {https://arxiv.org/abs/2406.12045} {$\tau$-bench: A benchmark for tool-agent-user interaction in real-world domains}.
\newblock \emph{Preprint}, arXiv:2406.12045.

\bibitem[{Yao et~al.(2023)Yao, Zhao, Yu, Du, Shafran, Narasimhan, and Cao}]{ReAct}
Shunyu Yao, Jeffrey Zhao, Dian Yu, Nan Du, Izhak Shafran, Karthik Narasimhan, and Yuan Cao. 2023.
\newblock \href {https://arxiv.org/abs/2210.03629} {React: Synergizing reasoning and acting in language models}.
\newblock \emph{Preprint}, arXiv:2210.03629.

\bibitem[{Ye et~al.(2026)Ye, Zhang, Jia, and Hu}]{TRUSTDESC}
Hengkai Ye, Zhechang Zhang, Jinyuan Jia, and Hong Hu. 2026.
\newblock \href {https://arxiv.org/abs/2604.07536} {Trustdesc: Preventing tool poisoning in llm applications via trusted description generation}.
\newblock \emph{Preprint}, arXiv:2604.07536.

\bibitem[{Ye et~al.(2025)Ye, Du, Yao, Lin, Xu, Chen, Wang, Zhu, Xi, Yuan, Gui, Zhang, Huang, and Chen}]{ToolHop}
Junjie Ye, Zhengyin Du, Xuesong Yao, Weijian Lin, Yufei Xu, Zehui Chen, Zaiyuan Wang, Sining Zhu, Zhiheng Xi, Siyu Yuan, Tao Gui, Qi~Zhang, Xuanjing Huang, and Jiecao Chen. 2025.
\newblock \href {https://doi.org/10.18653/v1/2025.acl-long.150} {{T}ool{H}op: A query-driven benchmark for evaluating large language models in multi-hop tool use}.
\newblock In \emph{Proceedings of the 63rd Annual Meeting of the Association for Computational Linguistics (Volume 1: Long Papers)}, pages 2995--3021, Vienna, Austria. Association for Computational Linguistics.

\bibitem[{Zhou et~al.(2023)Zhou, Xu, Zhu, Zhou, Lo, Sridhar, Cheng, Ou, Bisk, Fried, Alon, and Neubig}]{WebArena}
Shuyan Zhou, Frank~F. Xu, Hao Zhu, Xuhui Zhou, Robert Lo, Abishek Sridhar, Xianyi Cheng, Tianyue Ou, Yonatan Bisk, Daniel Fried, Uri Alon, and Graham Neubig. 2023.
\newblock \href {https://arxiv.org/abs/2307.13854} {{WebArena}: A realistic web environment for building autonomous agents}.
\newblock \emph{Preprint}, arXiv:2307.13854.

\bibitem[{Zong et~al.(2025)Zong, Liu, Zheng, Li, Xu, Shi, Wang, Wang, Chan, and Song}]{zong2025critical}
Qing Zong, Jiayu Liu, Tianshi Zheng, Chunyang Li, Baixuan Xu, Haochen Shi, Weiqi Wang, Zhaowei Wang, Chunkit Chan, and Yangqiu Song. 2025.
\newblock Critical: Can critique help llm uncertainty or confidence calibration?
\newblock \emph{arXiv preprint arXiv:2510.24505}.

\bibitem[{Zou et~al.(2025)Zou, Yang, Qi, Chen, Ai, Shen, He, and Wang}]{AutoTool}
Jiaru Zou, Ling Yang, Yunzhe Qi, Sirui Chen, Mengting Ai, Ke~Shen, Jingrui He, and Mengdi Wang. 2025.
\newblock \href {https://arxiv.org/abs/2512.13278} {Autotool: Dynamic tool selection and integration for agentic reasoning}.
\newblock \emph{Preprint}, arXiv:2512.13278.

\end{thebibliography}
\end{document}